\documentclass{article}
\usepackage[T1]{fontenc}
\usepackage{microtype}
\usepackage{hyperref}
\usepackage{nano_preprint}
\usepackage{xurl}
\usepackage{amsmath,amssymb,amsthm}
\DeclareMathSizes{7.5}{7}{5}{5}
\usepackage{booktabs,tabularx,array}
\usepackage{graphicx}
\usepackage{xcolor}
\usepackage{tikz}
\usetikzlibrary{arrows.meta,positioning}
\newcommand{\nano}[1]{\texttt{NANO#1}}

\newcommand{\no}{\mathrm{N.O.}}
\newcommand{\na}{\mathrm{N.A.}}
\newcommand{\E}{\mathbb{E}}
\newcommand{\R}{\mathbb{R}}
\newcommand{\ind}{\mathbf{1}}
\newcommand{\normal}{\mathcal{N}}

\newcolumntype{Y}{>{\raggedright\arraybackslash}X}

\theoremstyle{plain}
\newtheorem{theorem}{Theorem}

\edef\NanoTopFloatLimit{\number\value{topnumber}}
\AddToHookNext{shipout/after}{\setcounter{topnumber}{\NanoTopFloatLimit}}

\icmltitlerunning{NANO: Unifying Missingness and Input Uncertainty in Tabular Learning}
\hypersetup{
  pdftitle={\nano{}: Unifying Missingness and Input Uncertainty in Tabular Learning},
  pdfauthor={Kuangdai Leng, Maria Tsekhmistrenko, Matteo Bagagli, Simon Jeffery, Tarje Nissen-Meyer},
  pdfsubject={Tabular learning with applicability and confidence},
  pdfkeywords={tabular learning, missing data, structural absence, input uncertainty}
}

\begin{document}

\twocolumn[
  \icmltitle{NANO: Unifying Missingness and Input Uncertainty in Tabular Learning}
  \begin{icmlauthorlist}
    \icmlauthor{Kuangdai Leng}{erp}
    \icmlauthor{Maria Tsekhmistrenko}{erp}
    \icmlauthor{Matteo Bagagli}{erp}
    \icmlauthor{Simon Jeffery}{erp,harper}
    \icmlauthor{Tarje Nissen-Meyer}{erp,exeter}
  \end{icmlauthorlist}
  \icmlaffiliation{erp}{Earth Rover Program, London, United Kingdom}
  \icmlaffiliation{harper}{Centre for Crop and Environmental Science, Harper Adams University, Newport, Shropshire, United Kingdom}
  \icmlaffiliation{exeter}{Department of Mathematics and Statistics \& Center for Environmental Intelligence, University of Exeter, Exeter, United Kingdom}
  \icmlcorrespondingauthor{Tarje Nissen-Meyer}{tarje@earthroverprogram.org}
  \vskip 0.3in
]
\begin{NoHyper}
\printAffiliationsAndNotice{}
\end{NoHyper}

\begin{abstract}
Missing values are common in tabular data, yet current approaches to handling them often overlook two dimensions. First, an observed value may be uncertain, while a \emph{Not-Observed} (N.O.) value may still have a reasonable estimate. Inputs should therefore span a continuous spectrum from no evidence to a fully trusted observation. Second, \emph{Not-Applicable} (N.A.) means that no meaningful value exists for a record, yet this semantic state often lacks explicit representation. To address both dimensions, we introduce \nano{}, a unified plug-in for input encoding that fits most token-based architectures. Guided by Bayesian evidence weighting, \nano{} adjusts the influence of a supplied value according to confidence, with all zero-confidence estimates mapped to the same N.O. representation. \nano{} accepts, but does not require, per-example confidence information during training or inference. For N.A., a separate representation enables models to encode inapplicability and its schema-defined reasons into context. Experiments across eleven datasets and nine backbones show that mixed-confidence training enables a single \nano{} model to generalize across inference confidence levels, remaining competitive without estimates and improving steadily as estimate reliability and confidence increase. The effectiveness of \nano{}’s handling of N.A. is further demonstrated on three augmented datasets. 
\end{abstract}

\section{Introduction}
\label{sec:introduction}

Tabular learning builds predictive models from tables, where rows describe records, and columns contain numerical or categorical attributes \citep{borisov2024survey}. A common task for self-supervised representation learning is \emph{masked feature modeling}: a model learns to recover deliberately hidden entries from the remaining features \citep{yoon2020vime}. By varying which features are masked during training, the same model learns to predict different subsets of features from the remaining ones, enabling any-to-any prediction.

Real-world tables are often incomplete before any training mask is applied. Some gaps arise randomly, but many reflect systematic differences in data collection. In scientific datasets, the variables recorded can change with observation periods, geographic coverage, and experimental protocols. For example, LUCAS-based soil data have uneven property coverage across survey campaigns \citep{leng2026lucas}, while FLUXNET measurements have uneven availability across sites and observation periods \citep{pastorello2020fluxnet}. Such variation creates structured missingness when records are combined into a shared table, and these patterns can carry statistical information about the sampled populations \citep{little2019statistical,mitra2023structured}. The model's input layer should therefore preserve the statistical information and semantic meaning of missing values.

To learn from incomplete data, existing approaches include imputing missing values \citep{vanbuuren2011mice,yoon2018gain}, encoding missing entries with learned embeddings \citep[e.g.,][]{somepalli2021saint}, adapting models to incomplete inputs \citep{caruso2026naim}, and representing missing values probabilistically \citep{smieja2018processing}. These directions are reviewed in Section~\ref{sec:related}. However, two aspects of missingness remain insufficiently represented. First, the usual binary distinction between observed and missing values cannot capture common situations encountered in practice: an observed value is not necessarily reliable, while an unobserved value may still have a reasonable estimate. These intermediate states call for a continuous manner to express how much evidence an input provides. Second, an entry may be missing because the field does not apply to the record. This absence has a specific meaning and can convey structural information about the record \citep{chechik2008absent,comar2013missing}. Treating such entries simply as unknown values obscures this semantic distinction.

\begin{figure}[t]
  \centering
  \includegraphics[width=\columnwidth]{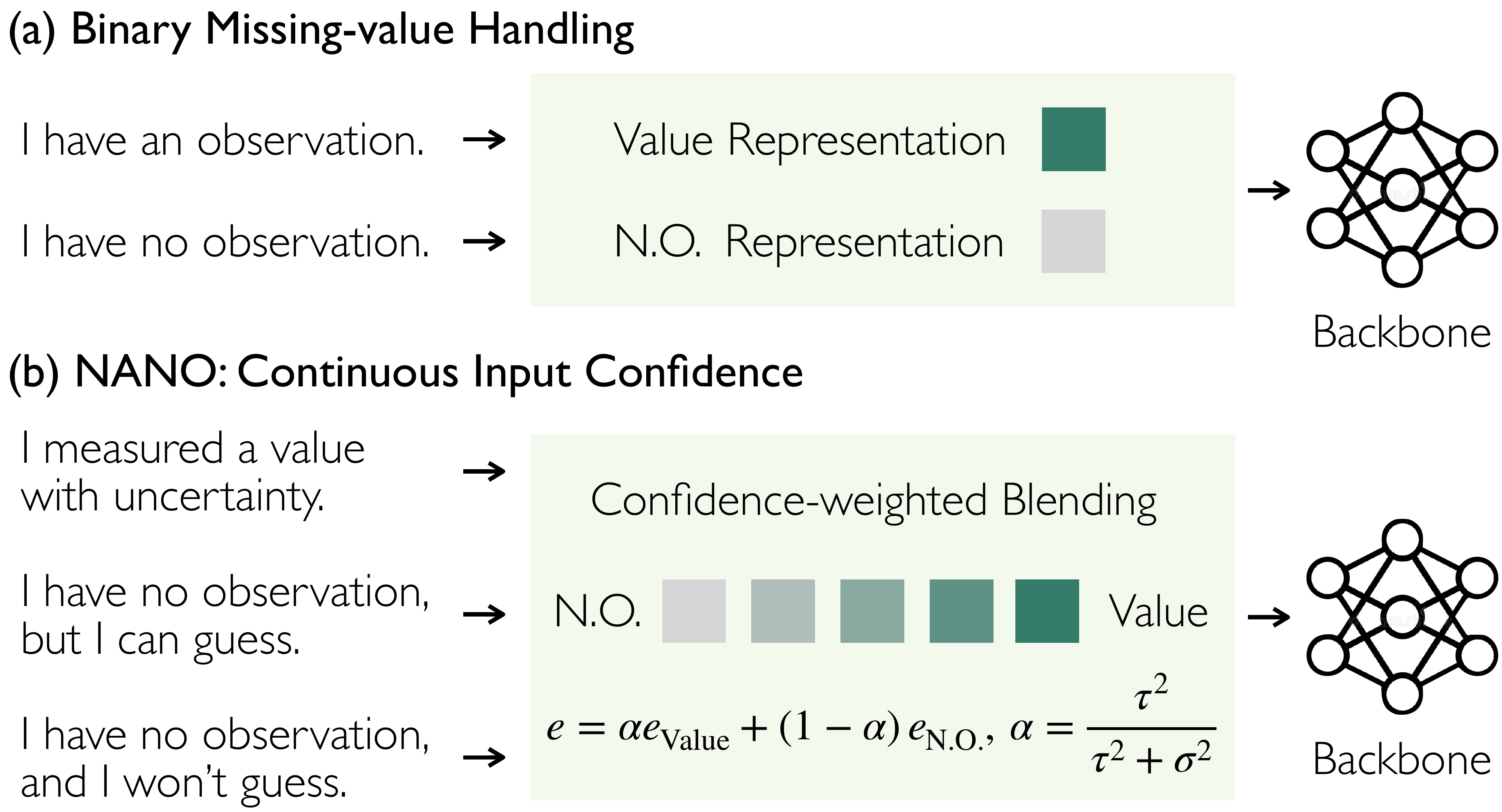}
  \caption{From binary missing-value handling to continuous input confidence. \nano{} uses a confidence weight $\alpha$ to blend the value embedding $e_{\mathrm{Value}}$ and the N.O. embedding $e_{\mathrm{N.O.}}$. In a Gaussian reference model, $\alpha$ is determined by $\tau^2$, the prior variance of the latent representation before considering the supplied value, and $\sigma^2$, the supplied value's error variance in the same latent space. Bayesian weighting gives greater influence to the supplied value when its error is smaller or prior uncertainty is larger. Given fixed $\tau^2>0$, a wild guess ($\sigma^2\to\infty$) gives $\alpha\to0$, recovering the N.O. limit; a fully trusted observation ($\sigma^2\to0$) gives $\alpha\to1$, recovering the value limit.}
  \label{fig:no-concept}
\end{figure}

For \emph{Not-Observed} (N.O.) entries, a meaningful value exists but has not been observed or recorded. Our recipe is to treat N.O. as a limiting case of input uncertainty. This treatment rests on two insights. First, \emph{a cell with no observation is semantically equivalent to it being filled with a random guess with zero confidence}; their input encodings should therefore be identical regardless of the guessed value. Second, \emph{an uncertain observation and a credible estimate can carry equivalent evidence about the underlying value}, so both belong on the same continuous scale. Guided by Bayesian evidence weighting, \nano{} constructs this scale by blending value and N.O. representations according to confidence. The resulting continuum spans absent evidence, observations or estimates under varying confidence, up to a fully trusted observation, as illustrated in Figure~\ref{fig:no-concept}.

For \emph{Not-Applicable} (N.A.) entries, no meaningful value exists for the record, but such absence is itself informative. For example, ``time since previous contact'' has no value for a customer who has never been contacted; this absence describes the customer's history and can help interpret other columns. Therefore, N.A. deserves its own representation. More significantly, applicability rules can often be inferred from the table schema (the definitions and relationships of its columns), without requiring record-wise annotations. These schema-defined relationships motivate preserving known N.A. reason information in input encoding.

We bring these insights together in \nano{}, a unified framework for tabular input encoding. Our main contributions are summarized below:

\textbf{Connecting missingness and input uncertainty.}\quad We semantically connect N.O. with the infinite-uncertainty limit of input evidence, which naturally leads to a continuous input encoding from N.O. to fully trusted values. The confidence weighting rule admits a general second-moment justification, and the resulting blend yields the exact Bayesian posterior mean under Gaussian priors and noise.

\textbf{A plug-in encoder with flexible input confidence.}\quad \nano{} modulates input embeddings without changing their dimensionality, preserving compatibility with existing token-based architectures. Mixed-confidence training enables a single model to generalize across confidence levels at inference, supporting the extension from missing-value prediction to confidence-conditioned refinement of estimates.

\textbf{N.A. Semantics and Schema-defined Reasons.}\quad We give N.A. a distinct representation and enrich it with information from schema-defined reason columns. This allows known inapplicability and its reasons to contribute structural context without requiring reason annotations for each record.

\section{Related Work}
\label{sec:related}

\textbf{Missingness and Input Representation.}\quad
Missingness is often structured, reflecting systematic differences in which variables are collected across records \citep{mitra2023structured}. \citet{lemorvan2020neumiss} show that even a linear complete-data relationship can require different prediction rules for different observed feature subsets. Beyond this statistical role, missing entries can differ in meaning: a value may be unknown, or the field may not apply to the record \citep{chechik2008absent,comar2013missing}. \nano{} introduces an inductive bias to help models use both statistical and semantic information carried by missing values.

\textbf{Not-Observed Inputs.}\quad
Approaches to N.O. inputs make different choices about what a predictor should receive from an incomplete record. Imputation reconstructs unavailable values from relationships among features, allowing backbones to operate on completed inputs. These relationships can be estimated through feature-wise conditional models, as in MICE \citep{vanbuuren2011mice} and missForest \citep{stekhoven2012missforest}, or learned through an adversarial generator, as in GAIN \citep{yoon2018gain}. ReMasker \citep{du2024remasker} reconstructs missing entries from available context, using mask tokens to identify the positions to be recovered. Probabilistic imputers such as the extended Gaussian copula (EGC) \citep{zhao2022probabilistic} describe a range of plausible values for missing entries, allowing uncertainty about these values to be retained through multiple imputation. Non-imputation approaches handle missingness directly within the predictor. Not Another Imputation Method (NAIM) \citep{caruso2026naim} excludes missing features from attention context. SAINT \citep[e.g.,][]{somepalli2021saint} uses learned feature-specific missing-state embeddings to represent unknown values explicitly, allowing the model to learn how missing states relate to observed context. Expected Activation (EAct) \citep{smieja2018processing} averages first-layer responses over a conditional distribution of possible values, allowing uncertainty about missing entries to influence later representations.

The similarities between \nano{} and these approaches are twofold. First, like approaches using missing-state tokens \citep[e.g.,][]{somepalli2021saint}, \nano{} uses a column-specific N.O. embedding shared across records, representing a cell with neither an observation nor an informative estimate. This choice facilitates blending N.O. and value representations. Second, the use of prior information connects \nano{} to probabilistic treatments such as EGC \citep{zhao2022probabilistic} and EAct \citep{smieja2018processing}. In the present formulation, \nano{} uses column-specific prior references, while the backbone learns dependencies among columns. Despite these similarities, the key difference lies in \nano{}'s input interface: an unobserved cell can carry an estimate together with confidence in that estimate. This setting enables a transition from \emph{predicting a missing value} to \emph{refining an estimate}, with ordinary missing-value prediction recovered at zero confidence. A single trained model can therefore predict without estimates or refine estimates at different confidence levels.

\textbf{Uncertainty in Inputs.}\quad
An available input need not be an exact measurement, as addressed in methods for learning from uncertain covariates. \citet{shivaswamy2006second} construct classifiers and regressors from input means and covariances, including uncertainty retained after imputing missing components. \citet{mchutchon2011gaussian} model measurement noise in the covariates and propagate test input uncertainty into Gaussian process predictions. \citet{damianou2016variational} present a related view: \emph{missing inputs can be treated as a special or limiting case of uncertain-input modeling} through a variational Gaussian process. However, they do not explicitly formalize N.O. as the common limit of arbitrary estimates with vanishing reliability. \nano{} makes this equivalence an explicit theoretical condition and constructs a continuous encoding in which any estimate converges to the same N.O. representation as confidence falls to zero.

\textbf{Not-Applicable Inputs.}\quad
The distinction between unknown and inapplicable values has already motivated explicit learning methods. \citet{chechik2008absent} classify each instance in the subspace of its existing features, while \citet{comar2013missing} develop tree and kernel classifiers for inapplicable continuous-valued features. Survey research also uses known applicability rules to preserve skip patterns during imputation \citep{zhang2024skip}. Such rules reflect an important property of tabular data: applicability mechanisms can often be specified at the table-schema level. Within a token-based framework, \nano{} uses these relationships to define reason columns, whose available values enrich the column-specific N.A. representations. This data-motivated design explicitly connects the N.A. state with available information about why it occurs.

\textbf{Tabular Backbones.}\quad
Many tabular architectures focus on feature interactions \citep[e.g.,][]{gorishniy2021revisiting,somepalli2021saint}. Methods for incomplete inputs include NAIM \citep{caruso2026naim}, which uses attention masking to exclude missing features from the context; ReMasker \citep{du2024remasker}, which excludes missing entries from the encoder and introduces mask tokens in the decoder; GRAPE \citep{you2020handling}, which infers missing cell values by passing messages along observed cell edges; and VAEAC \citep{ivanov2019vaeac}, which models missing values conditioned on arbitrary observed subsets. \nano{} complements these learning mechanisms by producing one embedding per cell, which the backbone receives at the corresponding column position.

\section{Method}
\label{sec:method}

\nano{} extends ordinary value embeddings to express uncertain evidence and inapplicability while preserving the input format expected by the backbone. We begin with ordinary value encodings, then develop a continuous formulation connecting N.O. and input uncertainty, followed by a separate representation for N.A. and its schema-defined reasons.

\subsection{Value Encodings}
\label{sec:contract}

Let $i\in\{1,\ldots,N\}$ index records and $j\in\{1,\ldots,M\}$ index columns or features. Write $x_{ij}$ for the value of cell $(i,j)$. A numerical column takes values in $\R^{D_j}$, where $D_j=1$ for a scalar; a categorical column has $K_j$ categories. Each column has a value encoder $e_j$ that maps its values to vector embeddings of the same dimension $D$:
\begin{equation}
 u_{ij}=e_j(x_{ij})=
 \begin{cases}
 f_j(x_{ij}),&\text{numerical},\\
 E_j[x_{ij}],&\text{categorical},
 \end{cases}
 \label{eq:value-encoding}
\end{equation}
where $f_j:\R^{D_j}\to\R^D$ is a numerical encoder such as an MLP, and $E_j\in\R^{K_j\times D}$ is a learned embedding table. Adding a learned column identity $c_j\in\R^D$ gives the final token embedding $t_{ij}=c_j+u_{ij}$, which is passed to the backbone (e.g., to its first self-attention layer). A vector-valued numerical cell occupies one token.

\subsection{Not-Observed Inputs}
\label{sec:no}

\subsubsection{A Common Scale for Evidence}
Let $v_{ij}$ be an estimate of $x_{ij}$ and $\alpha_{ij}\in[0,1]$ its confidence. Write $z_{ij}=e_j(v_{ij})$ for its value embedding, and let $b_j^\no{}\in\R^D$ represent the N.O. state for column $j$. We define
\begin{equation}
 h_j(v_{ij},\alpha_{ij})=\alpha_{ij} e_j(v_{ij})+(1-\alpha_{ij})b_j^\no{}.
 \label{eq:gate}
\end{equation}
At zero confidence, every estimate gives the same encoding as the N.O. state; at full confidence ($\alpha_{ij}=1$), the ordinary value encoding is recovered. Between them, confidence moves the encoding continuously along the line segment.

Now we establish a general reference for choosing $\alpha_{ij}$ from a prior and an evidence model. Let $u\in\R^D$ be the true token to be estimated, $\mu=\E[u]$ its prior mean, and $z\in\R^D$ a nonconstant evidence token. Assume the decomposition
\begin{equation}
 u=\mu+\theta+\xi,\qquad z=\mu+\theta+\varepsilon,
 \label{eq:shared-signal}
\end{equation}
where $\theta$ is the variation shared by the true token and the evidence, and $\xi$ and $\varepsilon$ are their respective residual components. We assume that $\theta$, $\xi$, and $\varepsilon$ all have zero means and finite second moments, and are pairwise uncorrelated.

\begin{theorem}[Evidence-weighted blending]
\label{thm:evidence-blending}
Under the shared-signal assumptions above, consider the following scalar blend that estimates $u$ from the evidence $z$ and the prior reference $\mu$:
\begin{equation}
 \hat u(\lambda)=\lambda z+(1-\lambda)\mu,\qquad\lambda\in\R.
 \label{eq:reference-blend}
\end{equation}
The optimal weight that minimizes the mean squared error $\E\|u-\hat u(\lambda)\|^2$ is unique and equals
\begin{equation}
 \lambda^*=\frac{\E\|\theta\|^2}
 {\E\|\theta\|^2+\E\|\varepsilon\|^2}\in[0,1].
 \label{eq:general-weight}
\end{equation}
\end{theorem}

The proof is given in Appendix~\ref{app:second-moment}. Note that $\xi$ contributes the constant $\E\|\xi\|^2$ to the error for every $\lambda$, so it affects the minimum attainable error but not the optimal weight $\lambda^*$.

Under this reference model, evidence receives more weight when it captures more of the true variation and less weight when it contains more noise. When $\E\|\varepsilon\|^2=0$ (i.e., all variation in the evidence is shared with the true token), the blend uses the evidence token $z$ directly as $\lambda^*=1$. With $\E\|\theta\|^2$ held fixed, $\E\|\varepsilon\|^2\to\infty$ (i.e., a noise-dominated guess) gives $\lambda^*\to0$, returning to the prior reference $\mu$.

We now return to the input encoding in Eq.~\eqref{eq:gate}. Applying Theorem~\ref{thm:evidence-blending} to cell $(i,j)$ gives $\alpha_{ij}=\lambda_{ij}^*$ as the optimal scalar weight when confidence is calibrated to the reference model. At the N.O. endpoint, no observation or estimate is available beyond the prior, making its mean $\mu_j$ the optimal constant estimate of $u_{ij}$ under squared error and hence motivating its use as the column's N.O. representation $b_j^\no{}$.

\subsubsection{Numerical Inputs}
\label{sec:no-num}
For numerical cell $(i,j)$, let the true token $u_{ij}$ have prior mean $\mu_j$ and covariance $\tau_j^2 I_D$, where $I_D$ is the identity matrix. Let the centered error $\varepsilon_{ij}$ be uncorrelated with $u_{ij}$ and have covariance $\sigma_{ij}^2 I_D$, with $\tau_j,\sigma_{ij}\ge0$ and $\tau_j^2+\sigma_{ij}^2>0$. Here $\tau_j$ measures prior uncertainty in column $j$, and $\sigma_{ij}$ measures the error of the estimate for cell $(i,j)$ on the same token scale. We further assume that the noise-free evidence recovers the true token exactly: $\xi_{ij}=0$, equivalently $z_{ij}=u_{ij}+\varepsilon_{ij}$. Theorem~\ref{thm:evidence-blending} then gives the calibrated confidence (see proof in Appendix~\ref{app:num-specialization})
\begin{equation}
 \alpha_{ij}=\lambda_{ij}^*=\frac{\tau_j^2}{\tau_j^2+\sigma_{ij}^2}\in[0,1].
 \label{eq:precision}
\end{equation}
The form is intuitive: an estimate receives more weight when its error is smaller or the prior uncertainty is larger. For fixed $\tau_j>0$, $\sigma_{ij}\to\infty$ gives the N.O. limit $\alpha_{ij}\to0$, while $\sigma_{ij}=0$ gives the fully trusted limit $\alpha_{ij}=1$.

For a fixed prior scale $\tau_j>0$, specifying $\alpha_{ij}$ is equivalent to specifying $\sigma_{ij}$ through Eq.~\eqref{eq:precision}, including the endpoint limits. However, their meanings differ: $\sigma_{ij}$ describes the estimate's error scale, whereas $\alpha_{ij}$ expresses confidence relative to the prior. An application can therefore derive confidence from an error model or specify it directly from domain knowledge.

The optimal $\alpha_{ij}$ given by Eq.~\eqref{eq:precision} depends only on second moments. The following theorem further shows that, with a Gaussian prior and error, the same blend with this weight gives the exact Bayesian posterior mean; therefore, the posterior mean in Eq.~\eqref{eq:gaussian-mean} minimizes expected squared error among \emph{all} estimators based on $z_{ij}$, including nonlinear ones. Its proof is given in Appendix~\ref{app:num-specialization}.

\begin{theorem}[Gaussian evidence blending]
\label{thm:gaussian-blending}
Suppose $u_{ij}\sim\normal(\mu_j,\tau_j^2 I_D)$ and $\varepsilon_{ij}\sim\normal(0,\sigma_{ij}^2 I_D)$ are independent, with $\tau_j,\sigma_{ij}\ge0$ and $\tau_j^2+\sigma_{ij}^2>0$, and let $z_{ij}=u_{ij}+\varepsilon_{ij}$. The exact posterior mean is
\begin{equation}
 \E[u_{ij}\mid z_{ij}]
 =\alpha_{ij}z_{ij}+(1-\alpha_{ij})\mu_j,
 \label{eq:gaussian-mean}
\end{equation}
where $\alpha_{ij}$ is given by Eq.~\eqref{eq:precision}.
\end{theorem}

While $\mu_j$ is the optimal constant reference under squared error, \nano{} uses a learnable N.O. embedding and blends it with $z_{ij}$ through Eq.~\eqref{eq:gate}. Learning this reference directly avoids specifying a full prior and averaging the nonlinear encoder's outputs under that prior. Since this learned reference need not equal the mean of encoded values, the Bayesian result motivates the blending rule without guaranteeing an exact posterior interpretation for the learned encoding.

\subsubsection{Categorical Inputs}
\label{sec:no-cat}
For categorical inputs, estimation error is described by a conditional distribution over categories rather than a numerical error scale. Let the categorical value $x_{ij}$ have a known prior $\rho_j$ over its $K_j$ possible categories.

Let $Q_{ij}(\ell\mid k)=\Pr(v_{ij}=\ell\mid x_{ij}=k)$ be a known observation model: it describes how an estimate $v_{ij}$ can differ from the true category. Together, $\rho_j$ and $Q_{ij}$ specify the joint distribution of the true token $u_{ij}=e_j(x_{ij})$ and the evidence token $z_{ij}=e_j(v_{ij})$, with $\mu_j=\E[u_{ij}]$:
\begin{equation}
\mu_j=\sum_{k=1}^{K_j}\rho_j(k)\,e_j(k).
 \label{eq:cat-token}
\end{equation}

\begin{theorem}[Categorical evidence blending]
\label{thm:categorical-blending}
For the reference choice $b_j^\no{}=\mu_j$ in Eq.~\eqref{eq:gate}, if $u_{ij}$ and $z_{ij}$ (nonconstant) satisfy the shared-signal assumptions of Theorem~\ref{thm:evidence-blending}, the unique weight minimizing $\E\|u_{ij}-h_j(v_{ij},\alpha_{ij})\|^2$ is
\begin{equation}
 \alpha_{ij}=\lambda_{ij}^*
 =\frac{\E\langle u_{ij}-\mu_j,\,z_{ij}-\mu_j\rangle}
 {\E\|z_{ij}-\mu_j\|^2}\in[0,1].
 \label{eq:cat-confidence}
\end{equation}
\end{theorem}

This weight is the shared variation normalized by the evidence's total variation. An independent draw from the prior gives exactly zero weight, whereas an error-free estimate gives full weight. This result establishes optimal scalar blending; an exact Bayesian posterior mean requires additional structure in the observation model. The proof and the class-wise expansion of Eq.~\eqref{eq:cat-confidence} are given in Appendix~\ref{app:categorical}.

Representing each uncertain input by its full observation model is data demanding during both training and inference. \nano{} instead uses one sampled candidate category and a confidence score. For a fixed true category $x_{ij}$, let $q_{ij}(k)=Q_{ij}(k\mid x_{ij})$. Under the same reference $b_j^\no{}=\mu_j$, drawing $v_{ij}\sim q_{ij}$ gives
\begin{equation}
 \E_{q_{ij}}[h_j(v_{ij},\alpha_{ij})]
 =\alpha_{ij}\sum_{k=1}^{K_j}q_{ij}(k)e_j(k)+(1-\alpha_{ij})b_j^\no{}.
 \label{eq:cat-distribution-blending}
\end{equation}
The drawn sample approximates blending the full candidate distribution with the N.O. reference: it preserves the expected embedding while introducing sampling variation.

We seek a reference channel in which the specified confidence $\alpha_{ij}$ is the optimal weight in Theorem~\ref{thm:categorical-blending} and the blend around $\mu_j$ is the exact posterior mean of the true token. As proved in Appendix~\ref{app:categorical}, both properties hold for a prior-resampling channel, which retains the true category $x_{ij}$ with probability $\alpha_{ij}$ and otherwise draws independently from $\rho_j$:
\begin{equation}
 q_{ij}(k)=\alpha_{ij}\ind[k=x_{ij}]+(1-\alpha_{ij})\rho_j(k),
 \label{eq:cat-candidate-distribution}
\end{equation}
where the prior distribution $\rho_j$ can be estimated from observed data frequencies or domain knowledge.

In practice, the estimated prior need not match the true categorical distribution, so we use an independently learnable N.O. embedding $b_j^\no{}$, as in numerical encoding, rather than constrain it to $\mu_j$ in Eq.~\eqref{eq:cat-token}.

\subsection{Not-Applicable Inputs}
\label{sec:na}

N.A. means the cell cannot take a meaningful value. \nano{} represents this state with a column-specific learned embedding $b_j^{\na}\in\R^D$, which forms part of the context for predicting applicable cells in the record.

\nano{} further enriches this representation with information about why the field is inapplicable. Applicability often depends on constraints among columns, as defined by the table schema. These constraints are shared across records, so identifying reason columns requires only per-column rather than per-record annotations, while reason values come from existing fields.

Let $\mathcal R_j$ be the schema-defined set of reason columns for column $j$, and let $\mathcal T_{ij}\subseteq\mathcal R_j$ contain those whose values remain observed and applicable after masking. We encode an N.A. cell as
\begin{equation}
 h_{ij}^{\na}=b_j^{\na}+W_jr_{ij},
 \ \ r_{ij}=\frac{\sum_{k\in\mathcal T_{ij}}[c_k+e_k(x_{ik})]}
 {\max(1,|\mathcal T_{ij}|)}.
 \label{eq:reason}
\end{equation}
Here $W_j\in\R^{D\times D}$ is a learned, bias-free linear projection that maps pooled context to a reason representation for column $j$. If no such reason value is available, $r_{ij}=0$ and the encoding reduces to $b_j^{\na}$.

\section{Experiments}
\label{sec:experiments}

We evaluate \nano{} using masked feature modeling. We compare it with four baselines across a shared testbed of eleven datasets and nine backbones.

\subsection{Experimental Setup}
\label{sec:experimental-setup}

\textbf{Datasets.}\quad
Table~\ref{tab:dataset-overview} summarizes eleven datasets covering demographics \citep{becker1996adult}, finance \citep{moro2014telemarketing,yeh2009comparisons}, commerce and product measurements \citep{wickham2016ggplot2,sakar2019prediction,cortez2009modeling}, access control \citep{hamner2013amazon}, healthcare \citep{strack2014impact}, industrial monitoring \citep{scania2016aps}, and environmental science \citep{beringer2016ozflux,leng2026lucas}. The collection spans different sizes, feature types, and rates of native missingness, and includes vector-valued numerical cells. Sources and preprocessing are detailed in Appendix~\ref{app:datasets}. 

\begin{table}[!b]
\centering
\captionsetup{skip=3pt}
\caption{Datasets by size. NUM and CAT represent numerical and categorical features. Native Miss (\%) reports source missingness before masking, without distinguishing N.O. and N.A.}
\label{tab:dataset-overview}
\small
\setlength{\tabcolsep}{2.5pt}
\setlength{\aboverulesep}{0.3ex}
\setlength{\belowrulesep}{0.3ex}
\renewcommand{\arraystretch}{0.9}
\begin{tabular*}{\columnwidth}{@{\extracolsep{\fill}}lrrrrr@{}}
\toprule
& & \multicolumn{2}{c}{Columns} & \multicolumn{2}{c}{Native Miss (\%)}\\
\cmidrule(lr){3-4}\cmidrule(lr){5-6}
Dataset & Rows & NUM & CAT & NUM & CAT\\
\midrule
Wine Quality & 6.5~k & 11 & 1 & 0 & 0\\
Online Shoppers & 12~k & 10 & 8 & 0 & 0\\
Credit Default & 30~k & 14 & 10 & 0 & 0\\
Amazon Employee & 33~k & 0 & 8 & -- & 0\\
Bank Marketing & 41~k & 10 & 11 & 9.6 & 2.8\\
Adult & 49~k & 4 & 9 & 0 & 1.5\\
Diamonds & 54~k & 7 & 3 & $<0.1$ & 0\\
SoilFormer & 65~k & 29 & 5 & 11.2 & 10.6\\
APS Failure & 76~k & 107 & 1 & 12.6 & 0\\
Diabetes & 102~k & 8 & 37 & 0 & 10.5\\
FLUXNET & 157~k & 12 & 0 & 9.9 & --\\
\bottomrule
\end{tabular*}
\end{table}

\textbf{Input Heads: Focus of Comparison.}\quad
Since \nano{} operates at the input interface, we compare it with four alternative input treatments under common backbones. Conventional baselines include \emph{Context Exclusion (CExcl)}, which excludes unavailable context \citep[e.g.,][]{caruso2026naim}, and \emph{Missing-state Embedding (MsEmb)}, which learns a shared embedding per column as a missing-state indicator \citep[e.g.,][]{somepalli2021saint}. Probabilistic baselines include \emph{Expected Activation (EAct)} \citep{smieja2018processing}, which averages first-layer responses over possible values, and \emph{Extended Gaussian Copula (EGC)} \citep{zhao2022probabilistic}, which models feature dependencies for conditional imputation. EGC priors are unavailable for Amazon Employee, whose eight high-cardinality columns expand to 15,283 latent dimensions, and Diabetes, whose prior fit failed to converge. Details about baselines are given in Appendix~\ref{app:input-heads}.

For \nano{}, numerical reference scales $\tau_j^{\mathrm{ref}}$ are estimated by three-fold cross-fitting on the training split. Categorical priors $\rho_j^{\mathrm{ref}}$ are obtained by normalizing the counts of observed training categories without smoothing. Both are estimated from training data alone and shared across all backbones and seeds for each dataset. Given confidence $\alpha_{ij}\in(0,1]$, numerical estimates are simulated by adding Gaussian noise with standard deviation $\sigma_{ij}^{\mathrm{sim}}$ computed from $\tau_j^{\mathrm{ref}}$ using Eq.~\eqref{eq:simulated-numerical-estimate}.
Categorical estimates are sampled from $q_{ij}^{\mathrm{sim}}$, determined by $\rho_j^{\mathrm{ref}}$, the masked category, and confidence $\alpha_{ij}$ using Eq.~\eqref{eq:simulated-categorical-estimate}.

\textbf{Backbones.}\quad
We consider nine tabular backbones: FT-Transformer \citep{gorishniy2021revisiting}, NPT \citep{kossen2021self}, NAIM \citep{caruso2026naim}, SAINT \citep{somepalli2021saint}, ReMasker \citep{du2024remasker}, a PMAE-derived Mixer \citep{kim2025pmae,tolstikhin2021mixer}, TabNet \citep{arik2021tabnet}, VAEAC \citep{ivanov2019vaeac}, and GRAPE \citep{you2020handling}. The first five are Transformers; the remaining four learn feature interactions through MLP mixing, sparse feature selection, conditional latent-variable modeling, and graph message passing, respectively. For architectures with dedicated missing-input mechanisms, including NAIM, ReMasker, GRAPE, and VAEAC, we make their input treatment interchangeable while retaining the backbone, allowing every compared input head to operate within the same architecture (Appendix~\ref{app:backbones}).

\textbf{Loss and Metrics.}\quad
We use cell-specific loss attenuation to account for differences in reconstruction uncertainty \citep{kendall2017uncertainties}. For a scalar-valued numerical cell, the readout predicts a mean $\hat\mu_{ij}$ and log variance $\hat\zeta_{ij}$, giving
\begin{equation}
 \ell_{ij}
 =e^{-\hat\zeta_{ij}}(x_{ij}-\hat\mu_{ij})^2+\hat\zeta_{ij}.
 \label{eq:reconstruction-loss}
\end{equation}
This is twice the Gaussian negative log-likelihood (NLL), with the constant term omitted. Vector-valued numerical cells share one uncertainty scalar across components; categorical cells use a heuristic that similarly attenuates cross-entropy. Modeling output uncertainty is not required by \nano{}'s input interface, but enables more in-depth comparisons of the results. Full loss definitions are given in Appendix~\ref{app:loss-metrics}.

For model evaluation, we use numerical RMSE and categorical error rate (C-ERR), each averaged equally across columns of the corresponding type. We also report NLL for both feature types, alongside other metrics of predictive uncertainty and estimate refinement. Metric definitions and aggregation are given in Appendix~\ref{app:loss-metrics}.

\textbf{Training and Evaluation.}\quad
We independently mask 30\% of applicable observed cells, treating each numerical vector as one token. Masking hides recorded values from the input while retaining them for supervision; native N.O. and N.A. cells remain in the input table and are never targets. Training masks vary by epoch, while validation and test masks are fixed and identical across input heads. We train each configuration for 100 epochs under five random seeds, selecting one checkpoint per run for all test conditions. Shared protocol details are given in Appendix~\ref{app:training}.

\subsection{N.O. Results}
\label{sec:no-results}

We define four confidence levels: \texttt{CL-Zero}, \texttt{CL-Low}, \texttt{CL-Medium}, and \texttt{CL-High} (Table~\ref{tab:no-training-mixtures}). \texttt{CL-Zero} uses the N.O. reference without an estimate; the other levels draw $\alpha_{ij}$ uniformly within their specified ranges. We adopt three mixed-confidence training recipes, \nano{-N25}, \nano{-N40}, and \nano{-N70}, which sample these four levels with the ratios given in Table~\ref{tab:no-training-mixtures}. At inference, each recipe is tested separately at all four levels.

Overall, \emph{we compare seven input heads (four baselines plus three \nano{} recipes) across eleven datasets, nine backbones, and five random seeds, giving 3,465 planned training runs}. Excluding the EGC runs for which prior fitting was unsuccessful, we train 3,375 models in total.

\begin{table}[t]
\centering
\caption{Confidence levels and training recipes for \nano{}.}
\label{tab:no-training-mixtures}
\small
\setlength{\tabcolsep}{0pt}
\begin{tabular*}{\columnwidth}{@{\extracolsep{\fill}}lrrrr@{}}
\toprule
Recipe & \texttt{CL-Zero} & \texttt{CL-Low} & \texttt{CL-Medium} & \texttt{CL-High}\\
Range of $\alpha_{ij}$ & $0$ & $[0.05,0.25]$ & $[0.35,0.65]$ & $[0.75,0.95]$\\
\midrule
\nano{-N25} & 25\% & 25\% & 25\% & 25\%\\
\nano{-N40} & 40\% & 20\% & 20\% & 20\%\\
\nano{-N70} & 70\% & 10\% & 10\% & 10\%\\
\bottomrule
\end{tabular*}
\end{table}

\subsubsection{Cross-head Comparisons}

\begin{table*}[t]
\caption{Point-prediction comparison across input heads. All columns use \texttt{CL-Zero} except the last, which evaluates the same \nano{-N70} models at \texttt{CL-Low}. Means and SDs pool nine backbones and five seeds, retaining all runs except four failed models, all from VAEAC on SoilFormer: one with CExcl and three with EAct (Appendix~\ref{app:loss-metrics}). Bold marks the best result under \texttt{CL-Zero}; bold italic marks where \texttt{CL-Low} achieves the best overall performance. ---: absent feature type; $\dagger$: unsuccessful EGC prior fitting.}
\label{tab:table_abs_failure_filtered}
\centering
\small
\newcommand{\noTableDataFont}{\fontsize{8pt}{9pt}\selectfont}
\renewcommand{\arraystretch}{1.0}
% Generated by results_new/no/src/11_manuscript_table.py.
% Regenerate numeric cells; edit captions and spacing in the parent table.
% Data: results_new/no/metrics/table_abs_failure_filtered/main.csv
% Data SHA256: cc476452d2604cb5d2a3d21e4e0a607509b8d7e0af91ecaf699da99f314f5c15
\begin{tabular*}{\textwidth}{@{\extracolsep{\fill}}lccccc|c@{}}
\toprule
 & \multicolumn{5}{c}{\texttt{CL-Zero} (No Evidence)} & \texttt{CL-Low} \\
\cmidrule(lr){2-6}\cmidrule(l){7-7}
Benchmark & CExcl & MsEmb & EAct & EGC & \nano{-N70} & \nano{-N70} \\
\midrule
\multicolumn{7}{@{}l}{\textit{(a) Numerical features: macro RMSE in training z-score units $\downarrow$}} \\
\addlinespace[1pt]
\noTableDataFont Adult & \noTableDataFont $\boldsymbol{0.938}\,{\scriptstyle\pm\,0.006}$ & \noTableDataFont $\boldsymbol{0.938}\,{\scriptstyle\pm\,0.007}$ & \noTableDataFont $0.939\,{\scriptstyle\pm\,0.007}$ & \noTableDataFont $\boldsymbol{0.938}\,{\scriptstyle\pm\,0.006}$ & \noTableDataFont $0.941\,{\scriptstyle\pm\,0.011}$ & \noTableDataFont $\boldsymbol{\mathit{0.835}}\,{\scriptstyle\pm\,0.060}$ \\
\noTableDataFont Amazon Employee & \noTableDataFont --- & \noTableDataFont --- & \noTableDataFont --- & \noTableDataFont --- & \noTableDataFont --- & \noTableDataFont --- \\
\noTableDataFont APS Failure & \noTableDataFont $0.695\,{\scriptstyle\pm\,0.059}$ & \noTableDataFont $0.682\,{\scriptstyle\pm\,0.054}$ & \noTableDataFont $0.676\,{\scriptstyle\pm\,0.049}$ & \noTableDataFont $\boldsymbol{0.658}\,{\scriptstyle\pm\,0.045}$ & \noTableDataFont $0.686\,{\scriptstyle\pm\,0.061}$ & \noTableDataFont $0.668\,{\scriptstyle\pm\,0.075}$ \\
\noTableDataFont Bank Marketing & \noTableDataFont $0.500\,{\scriptstyle\pm\,0.018}$ & \noTableDataFont $0.503\,{\scriptstyle\pm\,0.024}$ & \noTableDataFont $0.520\,{\scriptstyle\pm\,0.016}$ & \noTableDataFont $0.509\,{\scriptstyle\pm\,0.028}$ & \noTableDataFont $\boldsymbol{0.494}\,{\scriptstyle\pm\,0.023}$ & \noTableDataFont $\boldsymbol{\mathit{0.442}}\,{\scriptstyle\pm\,0.036}$ \\
\noTableDataFont Credit Default & \noTableDataFont $\boldsymbol{0.637}\,{\scriptstyle\pm\,0.036}$ & \noTableDataFont $\boldsymbol{0.637}\,{\scriptstyle\pm\,0.038}$ & \noTableDataFont $\boldsymbol{0.637}\,{\scriptstyle\pm\,0.036}$ & \noTableDataFont $\boldsymbol{0.637}\,{\scriptstyle\pm\,0.030}$ & \noTableDataFont $0.640\,{\scriptstyle\pm\,0.038}$ & \noTableDataFont $\boldsymbol{\mathit{0.609}}\,{\scriptstyle\pm\,0.051}$ \\
\noTableDataFont Diabetes & \noTableDataFont $\boldsymbol{0.846}\,{\scriptstyle\pm\,0.013}$ & \noTableDataFont $\boldsymbol{0.846}\,{\scriptstyle\pm\,0.012}$ & \noTableDataFont $\boldsymbol{0.846}\,{\scriptstyle\pm\,0.008}$ & \noTableDataFont $\dagger$ & \noTableDataFont $\boldsymbol{0.846}\,{\scriptstyle\pm\,0.020}$ & \noTableDataFont $\boldsymbol{\mathit{0.768}}\,{\scriptstyle\pm\,0.035}$ \\
\noTableDataFont Diamonds & \noTableDataFont $\boldsymbol{0.268}\,{\scriptstyle\pm\,0.022}$ & \noTableDataFont $\boldsymbol{0.268}\,{\scriptstyle\pm\,0.018}$ & \noTableDataFont $0.280\,{\scriptstyle\pm\,0.023}$ & \noTableDataFont $\boldsymbol{0.268}\,{\scriptstyle\pm\,0.036}$ & \noTableDataFont $0.274\,{\scriptstyle\pm\,0.022}$ & \noTableDataFont $\boldsymbol{\mathit{0.238}}\,{\scriptstyle\pm\,0.021}$ \\
\noTableDataFont FLUXNET & \noTableDataFont $0.483\,{\scriptstyle\pm\,0.020}$ & \noTableDataFont $\boldsymbol{0.480}\,{\scriptstyle\pm\,0.014}$ & \noTableDataFont $0.572\,{\scriptstyle\pm\,0.421}$ & \noTableDataFont $0.486\,{\scriptstyle\pm\,0.019}$ & \noTableDataFont $0.487\,{\scriptstyle\pm\,0.017}$ & \noTableDataFont $\boldsymbol{\mathit{0.428}}\,{\scriptstyle\pm\,0.020}$ \\
\noTableDataFont Online Shoppers & \noTableDataFont $0.698\,{\scriptstyle\pm\,0.036}$ & \noTableDataFont $0.696\,{\scriptstyle\pm\,0.040}$ & \noTableDataFont $0.713\,{\scriptstyle\pm\,0.030}$ & \noTableDataFont $\boldsymbol{0.684}\,{\scriptstyle\pm\,0.027}$ & \noTableDataFont $0.703\,{\scriptstyle\pm\,0.039}$ & \noTableDataFont $\boldsymbol{\mathit{0.640}}\,{\scriptstyle\pm\,0.064}$ \\
\noTableDataFont SoilFormer & \noTableDataFont $0.500\,{\scriptstyle\pm\,0.070}$ & \noTableDataFont $0.532\,{\scriptstyle\pm\,0.215}$ & \noTableDataFont $0.520\,{\scriptstyle\pm\,0.130}$ & \noTableDataFont $\boldsymbol{0.481}\,{\scriptstyle\pm\,0.033}$ & \noTableDataFont $0.497\,{\scriptstyle\pm\,0.045}$ & \noTableDataFont $\boldsymbol{\mathit{0.465}}\,{\scriptstyle\pm\,0.050}$ \\
\noTableDataFont Wine Quality & \noTableDataFont $0.682\,{\scriptstyle\pm\,0.043}$ & \noTableDataFont $0.682\,{\scriptstyle\pm\,0.049}$ & \noTableDataFont $0.666\,{\scriptstyle\pm\,0.037}$ & \noTableDataFont $\boldsymbol{0.663}\,{\scriptstyle\pm\,0.029}$ & \noTableDataFont $0.680\,{\scriptstyle\pm\,0.038}$ & \noTableDataFont $\boldsymbol{\mathit{0.634}}\,{\scriptstyle\pm\,0.051}$ \\
\midrule
\multicolumn{7}{@{}l}{\textit{(b) Categorical features: macro C-ERR (\%) $\downarrow$}} \\
\addlinespace[1pt]
\noTableDataFont Adult & \noTableDataFont $\boldsymbol{30.04}\,{\scriptstyle\pm\,0.50}$ & \noTableDataFont $30.06\,{\scriptstyle\pm\,0.61}$ & \noTableDataFont $30.10\,{\scriptstyle\pm\,0.76}$ & \noTableDataFont $30.12\,{\scriptstyle\pm\,0.67}$ & \noTableDataFont $30.34\,{\scriptstyle\pm\,0.78}$ & \noTableDataFont $\boldsymbol{\mathit{29.38}}\,{\scriptstyle\pm\,1.03}$ \\
\noTableDataFont Amazon Employee & \noTableDataFont $39.64\,{\scriptstyle\pm\,7.60}$ & \noTableDataFont $39.90\,{\scriptstyle\pm\,8.07}$ & \noTableDataFont $\boldsymbol{34.53}\,{\scriptstyle\pm\,5.73}$ & \noTableDataFont $\dagger$ & \noTableDataFont $40.49\,{\scriptstyle\pm\,8.04}$ & \noTableDataFont $39.74\,{\scriptstyle\pm\,8.40}$ \\
\noTableDataFont APS Failure & \noTableDataFont $0.77\,{\scriptstyle\pm\,0.10}$ & \noTableDataFont $\boldsymbol{0.75}\,{\scriptstyle\pm\,0.10}$ & \noTableDataFont $0.78\,{\scriptstyle\pm\,0.11}$ & \noTableDataFont $0.76\,{\scriptstyle\pm\,0.12}$ & \noTableDataFont $\boldsymbol{0.75}\,{\scriptstyle\pm\,0.09}$ & \noTableDataFont $\boldsymbol{\mathit{0.70}}\,{\scriptstyle\pm\,0.07}$ \\
\noTableDataFont Bank Marketing & \noTableDataFont $29.03\,{\scriptstyle\pm\,1.71}$ & \noTableDataFont $29.09\,{\scriptstyle\pm\,1.81}$ & \noTableDataFont $29.31\,{\scriptstyle\pm\,1.50}$ & \noTableDataFont $\boldsymbol{28.81}\,{\scriptstyle\pm\,1.33}$ & \noTableDataFont $29.14\,{\scriptstyle\pm\,1.72}$ & \noTableDataFont $\boldsymbol{\mathit{27.36}}\,{\scriptstyle\pm\,2.42}$ \\
\noTableDataFont Credit Default & \noTableDataFont $22.34\,{\scriptstyle\pm\,2.26}$ & \noTableDataFont $22.48\,{\scriptstyle\pm\,2.40}$ & \noTableDataFont $22.68\,{\scriptstyle\pm\,1.25}$ & \noTableDataFont $\boldsymbol{21.77}\,{\scriptstyle\pm\,1.42}$ & \noTableDataFont $22.63\,{\scriptstyle\pm\,2.22}$ & \noTableDataFont $22.16\,{\scriptstyle\pm\,2.43}$ \\
\noTableDataFont Diabetes & \noTableDataFont $22.21\,{\scriptstyle\pm\,0.40}$ & \noTableDataFont $22.23\,{\scriptstyle\pm\,0.54}$ & \noTableDataFont $\boldsymbol{22.16}\,{\scriptstyle\pm\,0.69}$ & \noTableDataFont $\dagger$ & \noTableDataFont $22.36\,{\scriptstyle\pm\,0.62}$ & \noTableDataFont $\boldsymbol{\mathit{21.66}}\,{\scriptstyle\pm\,0.82}$ \\
\noTableDataFont Diamonds & \noTableDataFont $51.95\,{\scriptstyle\pm\,3.63}$ & \noTableDataFont $52.21\,{\scriptstyle\pm\,3.26}$ & \noTableDataFont $53.20\,{\scriptstyle\pm\,3.34}$ & \noTableDataFont $\boldsymbol{51.53}\,{\scriptstyle\pm\,2.41}$ & \noTableDataFont $53.41\,{\scriptstyle\pm\,3.65}$ & \noTableDataFont $\boldsymbol{\mathit{51.11}}\,{\scriptstyle\pm\,4.39}$ \\
\noTableDataFont FLUXNET & \noTableDataFont --- & \noTableDataFont --- & \noTableDataFont --- & \noTableDataFont --- & \noTableDataFont --- & \noTableDataFont --- \\
\noTableDataFont Online Shoppers & \noTableDataFont $36.63\,{\scriptstyle\pm\,2.32}$ & \noTableDataFont $36.56\,{\scriptstyle\pm\,2.48}$ & \noTableDataFont $36.40\,{\scriptstyle\pm\,1.82}$ & \noTableDataFont $\boldsymbol{36.22}\,{\scriptstyle\pm\,2.11}$ & \noTableDataFont $36.76\,{\scriptstyle\pm\,2.26}$ & \noTableDataFont $\boldsymbol{\mathit{35.69}}\,{\scriptstyle\pm\,2.78}$ \\
\noTableDataFont SoilFormer & \noTableDataFont $\boldsymbol{29.38}\,{\scriptstyle\pm\,1.87}$ & \noTableDataFont $29.55\,{\scriptstyle\pm\,2.04}$ & \noTableDataFont $31.85\,{\scriptstyle\pm\,1.85}$ & \noTableDataFont $29.64\,{\scriptstyle\pm\,2.11}$ & \noTableDataFont $29.73\,{\scriptstyle\pm\,1.72}$ & \noTableDataFont $\boldsymbol{\mathit{28.64}}\,{\scriptstyle\pm\,2.13}$ \\
\noTableDataFont Wine Quality & \noTableDataFont $1.34\,{\scriptstyle\pm\,0.47}$ & \noTableDataFont $1.33\,{\scriptstyle\pm\,0.47}$ & \noTableDataFont $1.36\,{\scriptstyle\pm\,0.40}$ & \noTableDataFont $\boldsymbol{0.81}\,{\scriptstyle\pm\,0.21}$ & \noTableDataFont $1.43\,{\scriptstyle\pm\,0.53}$ & \noTableDataFont $1.04\,{\scriptstyle\pm\,0.50}$ \\
\bottomrule
\end{tabular*}

\end{table*}

We first compare the four baselines with \nano{-N70}, which uses no estimates for 70\% of masked cells during training. Table~\ref{tab:table_abs_failure_filtered} reports RMSE and C-ERR aggregated over backbones and seeds. Using \texttt{CL-Zero} or no estimates, \nano{-N70} remains competitive with the other heads. EGC often achieves lower errors where its prior is available, although its conditional imputation can add significant iterative computation during training (Appendix~\ref{app:input-heads}). Switching to \texttt{CL-Low} substantially improves \nano{-N70}'s prediction, which achieves the best performance on most datasets. Recall that \texttt{CL-Low} assumes $\alpha_{ij}\in[0.05, 0.25]$. Under the Gaussian reference model, the midpoint $\alpha_{ij}=0.15$ corresponds to $\sigma_{ij}/\tau_j=\sqrt{(1-0.15)/0.15}\approx2.38$: the estimate's error can exceed twice the prior reference scale, a modest requirement that rough estimates can meet in many applications. These results support two intended behaviors: \emph{mixed-confidence training preserves competitive prediction without estimates, while even low-confidence estimates can improve predictions without retraining}.

Appendix~\ref{app:cross-head-results} extends the comparison to NLL, categorical Brier scores, and the width and reliability of prediction intervals. Low-confidence estimates lower NLL and Brier scores and narrow numerical prediction intervals, but the intervals are still conservative, containing true values more often than their declared coverage implies. Backbone-specific comparisons find weaker relative performance with TabNet and GRAPE, where sparse feature selection and message averaging may reduce the contribution of weak evidence.

\subsubsection{Cross-confidence Comparisons}
\label{sec:cross-confidence}

\begin{figure*}[t]
\centering
\includegraphics[width=\textwidth]{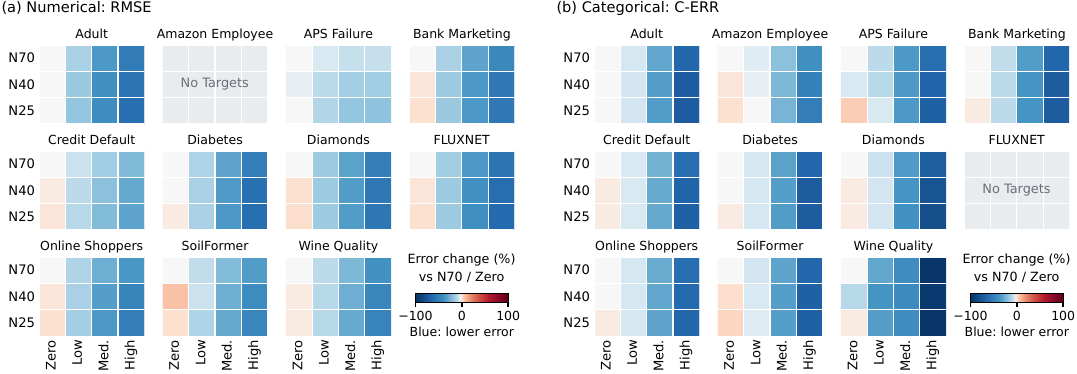}
\caption{Cross-confidence comparisons for (a) numerical RMSE and (b) categorical C-ERR. Within each pixel map, rows show the three training recipes and columns the four inference confidence levels. Colors indicate percentage changes from \nano{-N70} at \texttt{CL-Zero} (top-left pixel in each map), using means pooled across all nine backbones and five seeds after excluding failed models (Appendix~\ref{app:loss-metrics}).}
\label{fig:cross-confidence}
\end{figure*}

Using the same trained models, we next examine the impact of input confidence. Figure~\ref{fig:cross-confidence} compares the three training recipes across the four confidence levels at inference. At \texttt{CL-Zero}, \nano{-N70} usually has the lowest error, with the other two recipes often shown in red in the leftmost column of each map. At \texttt{CL-Medium} and \texttt{CL-High}, \nano{-N25} and \nano{-N40} generally achieve lower errors than \nano{-N70}. Across all three recipes, errors decrease substantially and monotonically from \texttt{CL-Zero} to \texttt{CL-High}, visible as progressively deeper blue from left to right within each row.

These observations suggest two implications. First, \emph{the training mixture can be chosen to better reflect the expected availability of evidence at deployment}: \nano{-N70} can be a reasonable choice when estimates are often absent, while more frequent availability may favor greater exposure to estimates during training. Second, \emph{all three recipes support the full range of inference confidence}: a single trained model can benefit from more reliable estimates supplied with higher confidence.

Appendix~\ref{app:cross-confidence-results} further compares predictions with supplied estimates, supporting the shift from missing-value prediction to estimate refinement. The gains are strongest for weak estimates, while further refinement of already accurate estimates can introduce errors. Appendix~\ref{app:cross-confidence-results} also attempts to confuse the models by either mixing confidence levels or deliberately assigning miscalibrated confidence. The models still benefit from mixed-confidence inputs, but severe overconfidence and underconfidence can reduce performance, showing that confidence distortion is a separate source of error even when the supplied estimates remain unchanged.

\subsection{N.A. Results}
\label{sec:na-results}

Verified native N.A. cases are limited in the benchmarks, so we augment Online Shoppers, SoilFormer, and Diabetes with synthetic numerical and categorical attributes whose applicability depends on existing fields, bringing the share of N.A. cells to approximately 14--17\%. For example, the added SoilFormer field ``woodland type'' is inapplicable to a site without woodland cover. Details are given in Appendix~\ref{app:na-results}. We compare MsEmb, which merges N.O. and N.A., with \nano{} using a separate N.A. embedding, first without and then with schema-defined reasons. We use \texttt{CL-Zero} to isolate the effects of N.A. treatment.

\begin{table}[t]
\caption{Point prediction with three embedding treatments on N.A.-augmented datasets. }
\label{tab:na-input-heads}
\centering
\footnotesize
\newcommand{\naTableNumberFont}{\fontsize{8pt}{9pt}\selectfont}
\setlength{\tabcolsep}{3pt}
% Generated by results_new/na/src/02_table.py.
% Collected scores SHA256: e9ef9c8acefe993163c3e063e850022bdc886eb69835b45453cc7d569941cad7
% Metric-wise Drop worst: four seeds per backbone; mean and sample SD.
\begin{tabular*}{\columnwidth}{@{\extracolsep{\fill}}lccc@{}}
\toprule
Benchmark & MsEmb & \shortstack{\nano{}\\w/o reasons} & \shortstack{\nano{}\\with reasons}\\
\midrule
\multicolumn{4}{l}{\textit{(a) Numerical RMSE}}\\
Online Shoppers & {\naTableNumberFont $0.682 \pm 0.081$} & {\naTableNumberFont $0.671 \pm 0.079$} & {\naTableNumberFont $\mathbf{0.670} \pm 0.078$}\\
SoilFormer & {\naTableNumberFont $0.442 \pm 0.068$} & {\naTableNumberFont $0.440 \pm 0.073$} & {\naTableNumberFont $\mathbf{0.438} \pm 0.069$}\\
Diabetes & {\naTableNumberFont $0.434 \pm 0.042$} & {\naTableNumberFont $\mathbf{0.419} \pm 0.043$} & {\naTableNumberFont $0.421 \pm 0.048$}\\
\midrule
\multicolumn{4}{l}{\textit{(b) Categorical C-ERR (\%)}}\\
Online Shoppers & {\naTableNumberFont $25.77 \pm 4.27$} & {\naTableNumberFont $25.82 \pm 4.84$} & {\naTableNumberFont $\mathbf{25.74} \pm 4.73$}\\
SoilFormer & {\naTableNumberFont $19.91 \pm 1.90$} & {\naTableNumberFont $19.03 \pm 2.01$} & {\naTableNumberFont $\mathbf{18.97} \pm 2.16$}\\
Diabetes & {\naTableNumberFont $15.00 \pm 2.09$} & {\naTableNumberFont $14.64 \pm 1.83$} & {\naTableNumberFont $\mathbf{14.59} \pm 1.77$}\\
\bottomrule
\end{tabular*}

\end{table}

Table~\ref{tab:na-input-heads} shows that distinguishing N.A. from N.O. generally improves prediction, and adding schema-defined reasons yields further gains in most comparisons. The other metrics show the same trends. The gains on these augmented datasets are consistent but modest, possibly because N.A. cells make up only a small share of the data. Nevertheless, explicitly representing N.A. preserves a semantic distinction needed for a complete input interface.

\subsection{Ablation studies}
In Appendix~\ref{app:mask-ratio-ablation}, we compare training mask ratios of 15\%, 30\%, 60\%, and 100\% to examine how reducing clean context affects learning from missing and uncertain inputs. \nano{} generally achieves better point accuracy at 30\% or 60\% than at either extreme, with modest gains from increasing masking from 30\% to 60\%. Models trained at every ratio continue to benefit from more reliable estimates supplied with higher confidence, even when training retains no clean context. These results favor moderate masking as a practical default for \nano{}, preserving reconstruction accuracy while supporting inference across varying input confidence.

In Appendix~\ref{app:heterogeneous-confidence-ablation}, we test three training profiles with confidence bands fixed by column, reflecting systematic differences in estimate availability and reliability. Training with a profile that matches inference usually yields modest gains over \nano{-N70}, which samples all four confidence levels within every column. However, changing the confidence pattern at deployment can reduce accuracy, especially when columns trained only with zero-confidence targets receive high-confidence estimates (as expected, since these columns were not exposed to estimates during training). These results favor retaining multiple confidence levels within each column during training unless its confidence range at deployment is known to remain stable.

\section{Conclusion}
\label{sec:conclusion}

Many applications operate between perfectly reliable observations and complete ignorance: measurements may be uncertain, while unobserved values may still admit informed estimates. A binary view of observed and missing inputs leaves these intermediate states difficult to express. Treating missingness as the limit of vanishing confidence is a natural alternative, but requires different estimates to converge to a common representation of the \emph{Not-Observed} (N.O.) state. \nano{} establishes this common limit within a continuous input encoding. The numerical weighting rule in Eq.~\eqref{eq:precision} illustrates \nano{}'s core idea: it determines confidence from estimation error and prior uncertainty, giving a continuous blend between value and N.O. representations. As estimation uncertainty grows without bound, the encoding approaches the same N.O. representation regardless of the supplied value. In addition, \nano{}'s separate representation for \emph{Not-Applicable} (N.A.) entries preserves inapplicability and its reasons, making the input interface semantically complete. Experiments show that \nano{} achieves competitive prediction accuracy compared with existing missing-input treatments and that the same trained model improves steadily as more reliable estimates are supplied with higher confidence. These results support the generalisation from missing-value prediction to confidence-conditioned refinement of estimates, with a single trained model operating across input confidence levels without retraining.

The main practical limitation concerns calibration between declared confidence and estimation error. This relationship can be specified during training using data or domain knowledge at the table or column level, but may differ at deployment, especially when confidence is assigned by end users. Application-specific pairs of declared confidence and estimation error offer a direct way to address this mismatch, although collecting them can be costly. Our ongoing work explores whether schema-level annotations shared across records can support this alignment within the existing framework. Such annotations would summarize each column's confidence--error relationship under deployment conditions and account for the predictive context provided by other columns. They could guide both the design of training conditions and users' quantitative confidence assignments.
\ificmlshowauthors
\section*{Acknowledgments}

This work was supported by the Earth Rover Program, a not-for-profit organization funded by the Bezos Earth Fund and UBS Optimus Foundation. The funders had no role in study design, data collection and analysis, the decision to publish, or preparation of the manuscript.
\fi

\bibliographystyle{icml2026}
\bibliography{references}

\clearpage
\appendix
\section{Theory for Not-Observed Inputs}
\label{app:method}

This appendix supplies the proofs and modeling details behind the Not-Observed (N.O.) encoding in Section~\ref{sec:no}. We derive the confidence weight from second moments, specialize it to numerical and categorical evidence, and establish when the resulting blend is an exact Bayesian posterior mean. Each proof restates the assumptions it needs.

For cell $(i,j)$, let $x_{ij}$ be its true value, $v_{ij}$ an observation or estimate, and $e_j$ the column's value encoder into $\R^D$. Write $u_{ij}=e_j(x_{ij})$ for the true token and $z_{ij}=e_j(v_{ij})$ for the evidence token. The confidence $\alpha_{ij}\in[0,1]$ controls the input encoding
\begin{equation*}
 h_j(v_{ij},\alpha_{ij})
 =\alpha_{ij}z_{ij}+(1-\alpha_{ij})b_j^{\no},
 \tag{\ref*{eq:gate}}
 \label{eq:app-input-blend}
\end{equation*}
This is Eq.~\eqref{eq:gate} in Method, with $b_j^{\no}\in\R^D$ the column's N.O. representation. The proofs first set this reference to the prior mean $\mu_j=\E[u_{ij}]$; we later distinguish that theoretical choice from a learned reference. All expectations hold the encoder parameters fixed.

\subsection{A Common Scale for Evidence}
\label{app:second-moment}

This subsection proves Theorem~\ref{thm:evidence-blending} and justifies the prior reference used in Eq.~\eqref{eq:gate}. The derivation concerns abstract random vectors and does not require a tabular setting or Gaussian distributions.

\subsubsection{Proof of Theorem~\ref{thm:evidence-blending}}

Let $u,z\in\R^D$ denote the true quantity and its evidence, with $\mu=\E[u]$. Assume
\begin{equation*}
 u=\mu+\theta+\xi,\qquad z=\mu+\theta+\varepsilon.
 \tag{\ref*{eq:shared-signal}}
 \label{eq:app-shared-signal}
\end{equation*}
Here $\theta$ is the variation shared by the true quantity and its evidence, while $\xi$ and $\varepsilon$ are their respective residuals. All three components have zero means and finite second moments, and are pairwise uncorrelated. The evidence $z$ is nonconstant, so $\E\|z-\mu\|^2>0$.

We hold this joint distribution fixed and estimate $u$ using the scalar blend
\begin{equation*}
 \hat u(\lambda)=\lambda z+(1-\lambda)\mu,
 \qquad \lambda\in\R.
 \tag{\ref*{eq:reference-blend}}
 \label{eq:app-reference-blend}
\end{equation*}
\textbf{Theorem~\ref{thm:evidence-blending}, restated.} The risk $R(\lambda)=\E\|u-\hat u(\lambda)\|^2$ has the unique minimizer
\begin{equation*}
 \lambda^*=\frac{\E\|\theta\|^2}
 {\E\|\theta\|^2+\E\|\varepsilon\|^2}\in[0,1].
 \tag{\ref*{eq:general-weight}}
 \label{eq:app-general-weight}
\end{equation*}

\begin{proof}
The estimation error is $(1-\lambda)\theta+\xi-\lambda\varepsilon$. Pairwise uncorrelatedness eliminates the cross terms, giving
\begin{equation}
 \begin{aligned}
 R(\lambda)&=\E\|u-\hat u(\lambda)\|^2\\
 &=(1-\lambda)^2\E\|\theta\|^2+\E\|\xi\|^2
   +\lambda^2\E\|\varepsilon\|^2.
 \end{aligned}
 \label{eq:general-risk}
\end{equation}
The coefficient of $\lambda^2$ in Eq.~\eqref{eq:general-risk} is positive because $z$ is nonconstant. The risk is therefore strictly convex, and setting its derivative to zero gives Eq.~\eqref{eq:app-general-weight}. Both second moments are nonnegative, so $\lambda^*\in[0,1]$.
\end{proof}

This proves the weight in Eq.~\eqref{eq:general-weight} of Method. For a cell, setting $b_j^{\no}=\mu_j$ and $\alpha_{ij}=\lambda_{ij}^*$ transfers this scalar optimality to the input encoding in Eq.~\eqref{eq:gate}.

\subsubsection{The Prior Reference and the No-Information Limit}
\label{app:encoding-endpoints}

Without evidence beyond the prior, any constant estimate $a\in\R^D$ satisfies
\begin{equation}
 \E\|u-a\|^2=\E\|u-\mu\|^2+\|a-\mu\|^2.
 \label{eq:prior-only-risk}
\end{equation}
The cross term vanishes because $\E[u-\mu]=0$. Equation~\eqref{eq:prior-only-risk} therefore makes $\mu$ the unique optimal constant estimate under squared error, justifying the prior-mean reference in Eq.~\eqref{eq:gate}.

The same reference is approached when evidence becomes dominated by noise. Holding $\E\|\theta\|^2$ fixed while $\E\|\varepsilon\|^2$ grows, the optimal coefficient tends to zero and
\begin{equation}
 \E\|\hat u(\lambda^*)-\mu\|^2
 =\frac{\bigl(\E\|\theta\|^2\bigr)^2}
 {\E\|\theta\|^2+\E\|\varepsilon\|^2}
 \longrightarrow0.
 \label{eq:app-noise-limit}
\end{equation}
Equation~\eqref{eq:app-noise-limit} establishes mean-square convergence to the prior reference even as the evidence variance grows. It complements the continuity of Eq.~\eqref{eq:gate}: the fixed-candidate encoding has a common zero-confidence endpoint, and optimally weighted noisy evidence also converges to that endpoint. Neither statement requires Gaussian distributions.

\subsection{Numerical Inputs}
\label{app:num-specialization}

This subsection derives the numerical confidence in Eq.~\eqref{eq:precision} and proves Theorem~\ref{thm:gaussian-blending}. We then clarify how error in numerical values differs from the token-space error assumed in these results.

\subsubsection{Specialization of Theorem~\ref{thm:evidence-blending}}

For cell $(i,j)$, suppose that the true token $u_{ij}$ has mean $\mu_j$ and covariance $\tau_j^2I_D$, where $I_D$ is the $D$-dimensional identity matrix. Let the estimation error $\varepsilon_{ij}$ be centered, uncorrelated with $u_{ij}$, and have covariance $\sigma_{ij}^2I_D$. Here $\tau_j$ is the column's prior scale and $\sigma_{ij}$ is the cell's estimation-error scale, both measured in token space, with $\tau_j,\sigma_{ij}\ge0$ and $\tau_j^2+\sigma_{ij}^2>0$. We model the evidence as
\begin{equation}
 z_{ij}=u_{ij}+\varepsilon_{ij}.
 \label{eq:app-numerical-additive}
\end{equation}
The additive model in Eq.~\eqref{eq:app-numerical-additive} is the shared-signal model of Eq.~\eqref{eq:app-shared-signal} with $\theta_{ij}=u_{ij}-\mu_j$ and $\xi_{ij}=0$. Thus removing the estimation error recovers the entire true token, rather than only its shared component. The relevant second moments are
\begin{equation}
 \E\|\theta_{ij}\|^2=D\tau_j^2,\qquad
 \E\|\varepsilon_{ij}\|^2=D\sigma_{ij}^2.
 \label{eq:app-numerical-moments}
\end{equation}
Substituting these moments into Eq.~\eqref{eq:app-general-weight} gives
\begin{equation*}
 \alpha_{ij}=\lambda_{ij}^*
 =\frac{\tau_j^2}{\tau_j^2+\sigma_{ij}^2}.
 \tag{\ref*{eq:precision}}
 \label{eq:app-numerical-confidence}
\end{equation*}
The dimension cancels, recovering Eq.~\eqref{eq:precision} without assuming Gaussian distributions. For fixed $\tau_j>0$, zero error gives $\alpha_{ij}=1$, while unbounded error gives the N.O. limit $\alpha_{ij}\to0$.

\subsubsection{Proof of Theorem~\ref{thm:gaussian-blending}}

We now strengthen the distributional assumptions while retaining the same variance-based weight. Suppose
\begin{equation}
 \begin{aligned}
 u_{ij}&\sim\normal(\mu_j,\tau_j^2I_D),\\
 \varepsilon_{ij}&\sim\normal(0,\sigma_{ij}^2I_D),\\
 z_{ij}&=u_{ij}+\varepsilon_{ij},
 \end{aligned}
 \label{eq:app-gaussian-assumptions}
\end{equation}
with $u_{ij}$ and $\varepsilon_{ij}$ independent, $\tau_j,\sigma_{ij}\ge0$, and $\tau_j^2+\sigma_{ij}^2>0$.

\textbf{Theorem~\ref{thm:gaussian-blending}, restated.} The blend with the confidence in Eq.~\eqref{eq:app-numerical-confidence} is the exact posterior mean:
\begin{equation*}
 \E[u_{ij}\mid z_{ij}]
 =\alpha_{ij}z_{ij}+(1-\alpha_{ij})\mu_j.
 \tag{\ref*{eq:gaussian-mean}}
 \label{eq:app-gaussian-mean}
\end{equation*}
We also establish its optimality among all estimators based on $z_{ij}$, extending the scalar optimality of Theorem~\ref{thm:evidence-blending}.

\begin{proof}
First take $\tau_j,\sigma_{ij}>0$. Independence of the additive error gives the Gaussian likelihood $z_{ij}\mid u_{ij}\sim\normal(u_{ij},\sigma_{ij}^2I_D)$. Bayes' rule combines this likelihood with the prior:
\begin{equation}
 \begin{aligned}
 p(u_{ij}\mid z_{ij})
 &\propto p(z_{ij}\mid u_{ij})p(u_{ij})\\
 &\propto\exp\left[
 -\frac{\|z_{ij}-u_{ij}\|^2}{2\sigma_{ij}^2}
 -\frac{\|u_{ij}-\mu_j\|^2}{2\tau_j^2}
 \right].
 \end{aligned}
 \label{eq:app-gaussian-kernel}
\end{equation}
Completing the square in Eq.~\eqref{eq:app-gaussian-kernel} gives a Gaussian with parameters
\begin{equation}
 \begin{aligned}
 \omega_{ij}^2&=(\tau_j^{-2}+\sigma_{ij}^{-2})^{-1},\\
 m_{ij}&=\omega_{ij}^2
          (\tau_j^{-2}\mu_j+\sigma_{ij}^{-2}z_{ij}).
 \end{aligned}
 \label{eq:app-gaussian-parameters}
\end{equation}
The posterior is therefore
\begin{equation}
 \begin{aligned}
 u_{ij}\mid z_{ij}&\sim\normal(m_{ij},\omega_{ij}^2I_D),\\
 \omega_{ij}^2&=\frac{\tau_j^2\sigma_{ij}^2}
                      {\tau_j^2+\sigma_{ij}^2}.
 \end{aligned}
 \label{eq:posterior}
\end{equation}
Here $m_{ij}$ is the posterior mean and $\omega_{ij}^2$ the variance of each coordinate. In Eq.~\eqref{eq:app-gaussian-parameters}, the coefficient of $z_{ij}$ is the error precision $\sigma_{ij}^{-2}$ divided by the sum of the prior and error precisions:
\begin{equation}
 \frac{\sigma_{ij}^{-2}}{\tau_j^{-2}+\sigma_{ij}^{-2}}
 =\frac{\tau_j^2}{\tau_j^2+\sigma_{ij}^2}
 =\alpha_{ij}.
 \label{eq:app-gaussian-precision}
\end{equation}
The coefficient of $\mu_j$ is the remaining weight $1-\alpha_{ij}$. This proves the posterior mean in Eq.~\eqref{eq:gaussian-mean} of Method, using exactly the second-moment confidence in Eq.~\eqref{eq:precision}.

If $\tau_j=0$ and $\sigma_{ij}>0$, the prior already fixes $u_{ij}=\mu_j$ almost surely, giving posterior mean $\mu_j$ and weight zero. If $\sigma_{ij}=0$ and $\tau_j>0$, the evidence equals $u_{ij}$ almost surely, giving posterior mean $z_{ij}$ and weight one. These cover the permitted zero-variance cases.
\end{proof}

To verify the stronger squared-error optimality, take any square-integrable estimator $g(z_{ij})$. Since $\E[u_{ij}-m_{ij}\mid z_{ij}]=0$, the cross term in the squared-error expansion vanishes:
\begin{equation}
 \begin{aligned}
 \E\|u_{ij}-g(z_{ij})\|^2
 &=\E\|u_{ij}-m_{ij}\|^2\\
 &\quad+\E\|m_{ij}-g(z_{ij})\|^2.
 \end{aligned}
 \label{eq:conditional-risk}
\end{equation}
The second term in Eq.~\eqref{eq:conditional-risk} is nonnegative and vanishes only when $g(z_{ij})=m_{ij}$ almost surely. The posterior mean therefore minimizes expected squared error over all such estimators, including nonlinear ones.

\subsubsection{From the Reference Model to Numerical Encoding}
\label{app:nonlinear-encoder}

For a numerical column, $e_j=f_j$ maps values in $\R^{D_j}$ into $\R^D$, with $D_j=1$ for a scalar value. The preceding results describe error after this encoding. In an application, however, the estimate and its error are specified in data space. For a data-space perturbation $\eta_{ij}^{\mathrm{data}}$ with $v_{ij}=x_{ij}+\eta_{ij}^{\mathrm{data}}$, the induced token error is
\begin{equation}
 \varepsilon_{ij}=f_j(x_{ij}+\eta_{ij}^{\mathrm{data}})-f_j(x_{ij}).
 \label{eq:app-encoded-error}
\end{equation}
A nonlinear $f_j$ need not preserve centered, independent Gaussian errors: the token error in Eq.~\eqref{eq:app-encoded-error} can be biased, depend on $x_{ij}$, or have nonisotropic covariance. A Gaussian prior over values likewise need not induce a Gaussian token prior. A data-space error scale $\sigma_{ij}^{\mathrm{data}}$ therefore cannot generally be identified with the token-space scale $\sigma_{ij}$ in Eq.~\eqref{eq:app-numerical-confidence}.

At inference, an application supplies a numerical estimate in data space together with confidence. Using Eq.~\eqref{eq:precision} to calibrate that confidence from a data-space error scale is therefore an approximation, rather than an exact token-space posterior calculation. Appendix~\ref{app:input-heads} specifies the error scales used in our experiments. Learning $b_j^{\no}$ independently also relaxes the prior-mean assumption, as discussed in Section~\ref{sec:no-num}, while preserving the continuous encoding and common zero-confidence endpoint of Eq.~\eqref{eq:gate}.

\subsection{Categorical Inputs}
\label{app:categorical}

This subsection proves the categorical weight in Eq.~\eqref{eq:cat-confidence} and expands it in terms of the class prior and observation model. We then show why the prior-resampling distribution in Eq.~\eqref{eq:cat-candidate-distribution} gives both this optimal weight and an exact posterior mean.

\subsubsection{Proof of Theorem~\ref{thm:categorical-blending}}

Let column $j$ have $K_j$ categories with fixed embeddings $e_j(k)\in\R^D$. The true category $x_{ij}$ has prior $\rho_j(k)=\Pr(x_{ij}=k)$. Its candidate $v_{ij}$ follows the observation model $Q_{ij}(\ell\mid k)=\Pr(v_{ij}=\ell\mid x_{ij}=k)$. Together they specify
\begin{equation}
 \Pr(x_{ij}=k,v_{ij}=\ell)=\rho_j(k)Q_{ij}(\ell\mid k).
 \label{eq:app-cat-joint}
\end{equation}
Write $u_{ij}=e_j(x_{ij})$, $z_{ij}=e_j(v_{ij})$, and define the prior mean
\begin{equation*}
 \mu_j=\E[u_{ij}]=\sum_{k=1}^{K_j}\rho_j(k)e_j(k).
 \tag{\ref*{eq:cat-token}}
 \label{eq:app-cat-prior-mean}
\end{equation*}
The theoretical encoding uses $b_j^{\no}=\mu_j$. Assume that $u_{ij}$ and the nonconstant $z_{ij}$ satisfy the shared-signal assumptions of Section~\ref{app:second-moment}.

\textbf{Theorem~\ref{thm:categorical-blending}, restated.} With the joint distribution fixed, the unique optimal scalar weight for the blend $\hat u_{ij}(\lambda)=\lambda z_{ij}+(1-\lambda)\mu_j$ is
\begin{equation*}
 \alpha_{ij}=\lambda_{ij}^*
 =\frac{\E\langle u_{ij}-\mu_j,z_{ij}-\mu_j\rangle}
 {\E\|z_{ij}-\mu_j\|^2}\in[0,1].
 \tag{\ref*{eq:cat-confidence}}
 \label{eq:app-cat-confidence}
\end{equation*}

\begin{proof}
Expanding the squared error gives
\begin{equation}
 \begin{aligned}
 R_{ij}(\lambda)
 &=\E\|u_{ij}-\mu_j\|^2\\
 &\quad-2\lambda\E\langle u_{ij}-\mu_j,z_{ij}-\mu_j\rangle\\
 &\quad+\lambda^2\E\|z_{ij}-\mu_j\|^2.
 \end{aligned}
 \label{eq:cat-general-risk}
\end{equation}
The last coefficient is positive, so setting the derivative to zero yields the ratio in Eq.~\eqref{eq:app-cat-confidence}. Under the shared-signal assumptions, its numerator is $\E\|\theta_{ij}\|^2$ and its denominator is $\E\|\theta_{ij}\|^2+\E\|\varepsilon_{ij}\|^2$. The ratio is therefore the general weight in Eq.~\eqref{eq:app-general-weight} and lies in $[0,1]$.
\end{proof}

Writing $\widetilde e_j(k)=e_j(k)-\mu_j$ gives the complete class-wise form
\begin{equation}
 \lambda_{ij}^*=
 \frac{\displaystyle\sum_{k,\ell}\rho_j(k)Q_{ij}(\ell\mid k)
 \langle\widetilde e_j(k),\widetilde e_j(\ell)\rangle}
 {\displaystyle\sum_{k,\ell}\rho_j(k)Q_{ij}(\ell\mid k)
 \|\widetilde e_j(\ell)\|^2},
 \label{eq:cat-class-weight}
\end{equation}
where both indices range from $1$ to $K_j$. Equation~\eqref{eq:cat-class-weight} expands Eq.~\eqref{eq:cat-confidence} of Method: its numerator measures alignment between true and candidate embeddings around the prior mean, and its denominator measures candidate variation. The shared-signal assumptions ensure a weight in $[0,1]$. For an arbitrary observation model, the optimal weight may fall outside this interval, and the resulting scalar blend need not equal the posterior mean. The channel below establishes both properties explicitly.

\subsubsection{The Prior-Resampling Channel}

We now verify the reference channel underlying Eq.~\eqref{eq:cat-candidate-distribution} in Method. Suppose the true category follows $\rho_j$ and fix $\alpha_{ij}\in[0,1]$ independently of that category. Retain the true category with probability $\alpha_{ij}$; otherwise draw a replacement independently from $\rho_j$. This defines
\begin{equation}
 Q_{ij}(\ell\mid k)
 =\alpha_{ij}\ind[\ell=k]+(1-\alpha_{ij})\rho_j(\ell).
 \label{eq:cat-channel}
\end{equation}
The true category and replacement follow the same prior. For a fixed $x_{ij}$, the candidate distribution is $q_{ij}(\ell)=Q_{ij}(\ell\mid x_{ij})$, which gives Eq.~\eqref{eq:cat-candidate-distribution}. This forward distribution generates estimates conditional on the true category; the posterior used for prediction reverses that conditioning.

\textbf{Optimal blending weight.} We hold the channel in Eq.~\eqref{eq:cat-channel} fixed and optimize the blend coefficient $\lambda$. This distinguishes the prescribed confidence $\alpha_{ij}$ from the weight to be derived.

To verify the shared-signal assumptions, draw $x'_{ij}\sim\rho_j$ independently of $x_{ij}$, and let $\beta_{ij}\sim\operatorname{Bernoulli}(\alpha_{ij})$ be independent of both. With $u'_{ij}=e_j(x'_{ij})$, the candidate token is $z_{ij}=\beta_{ij}u_{ij}+(1-\beta_{ij})u'_{ij}$. When $\beta_{ij}=1$, the true token is retained and its variation is shared; when $\beta_{ij}=0$, the two tokens come from independent prior draws and contribute separate residuals. These cases give the decomposition
\begin{equation}
 \begin{aligned}
 \theta_{ij}&=\beta_{ij}(u_{ij}-\mu_j),\\
 \xi_{ij}&=(1-\beta_{ij})(u_{ij}-\mu_j),\\
 \varepsilon_{ij}&=(1-\beta_{ij})(u'_{ij}-\mu_j).
 \end{aligned}
 \label{eq:cat-shared-signal}
\end{equation}
All components are centered. Since $\beta_{ij}(1-\beta_{ij})=0$ and the two prior draws are independent, they are pairwise uncorrelated. Unlike the numerical specialization, this construction allows a nonzero residual $\xi_{ij}$. Let $\nu_j=\E\|u_{ij}-\mu_j\|^2$ denote the total prior embedding variance. Then
\begin{equation}
 \begin{aligned}
 \E\|\theta_{ij}\|^2&=\alpha_{ij}\nu_j,\\
 \E\|\varepsilon_{ij}\|^2&=(1-\alpha_{ij})\nu_j.
 \end{aligned}
 \label{eq:cat-channel-moments}
\end{equation}
For $\nu_j>0$, Theorem~\ref{thm:evidence-blending} therefore gives
\begin{equation}
 \lambda_{ij}^*
 =\frac{\alpha_{ij}\nu_j}
 {\alpha_{ij}\nu_j+(1-\alpha_{ij})\nu_j}
 =\alpha_{ij}.
 \label{eq:cat-channel-weight}
\end{equation}
Thus the chosen channel parameter agrees with the optimal weight in the input blend. If $\nu_j=0$, all prior-supported categories have the same embedding and the weight is immaterial.

\textbf{Exact posterior mean.} Marginalizing Eq.~\eqref{eq:cat-channel} over the true category gives
\begin{equation*}
 \Pr(v_{ij}=\ell\mid\alpha_{ij})
 =\sum_{k=1}^{K_j}\rho_j(k)Q_{ij}(\ell\mid k)
 =\rho_j(\ell).
\end{equation*}
Bayes' rule therefore gives, for $\rho_j(\ell)>0$,
\begin{equation}
 \begin{aligned}
 &\Pr(x_{ij}=k\mid v_{ij}=\ell,\alpha_{ij})\\
 &=\frac{\rho_j(k)Q_{ij}(\ell\mid k)}{\rho_j(\ell)}\\
 &=\alpha_{ij}\ind[k=\ell]+(1-\alpha_{ij})\rho_j(k).
 \end{aligned}
 \label{eq:app-cat-posterior}
\end{equation}
Taking the mean embedding yields
\begin{equation}
 \begin{aligned}
 &\E[u_{ij}\mid v_{ij}=\ell,\alpha_{ij}]\\
 &\qquad=\alpha_{ij}e_j(\ell)+(1-\alpha_{ij})\mu_j.
 \end{aligned}
 \label{eq:app-cat-posterior-mean}
\end{equation}
With $b_j^{\no}=\mu_j$, Eq.~\eqref{eq:app-cat-posterior-mean} is the input encoding in Eq.~\eqref{eq:gate}. Together with Eq.~\eqref{eq:cat-channel-weight}, it proves the two properties claimed for Eq.~\eqref{eq:cat-candidate-distribution}: the prescribed confidence is the optimal scalar weight, and the blend is the exact posterior mean. At zero confidence, Eq.~\eqref{eq:app-cat-posterior} reduces to the prior independently of the candidate.

An independent replacement can still be correct by chance, so confidence is distinct from categorical accuracy:
\begin{equation}
 \begin{aligned}
 &\Pr(v_{ij}=x_{ij}\mid x_{ij}=k,\alpha_{ij})\\
 &\qquad=\alpha_{ij}+(1-\alpha_{ij})\rho_j(k).
 \end{aligned}
 \label{eq:cat-forward-correct}
\end{equation}

The posterior identity in Eq.~\eqref{eq:app-cat-posterior-mean} holds for any fixed category embeddings. Unlike numerical error passed through a nonlinear encoder, embedding lookup preserves this categorical channel exactly. Its probabilistic interpretation instead depends on a matched prior, confidence independent of the true category, and the reference $b_j^{\no}=\mu_j$. An estimated prior or freely learned reference need not retain this exact interpretation, but the sampling identity in Eq.~\eqref{eq:cat-distribution-blending} and common zero-confidence endpoint in Eq.~\eqref{eq:gate} still hold.

\section{Experimental Setup Details}
\label{app:protocol}

This appendix expands Section~\ref{sec:experimental-setup} in the same order: datasets, input heads, backbones, losses and metrics, and training and evaluation.

\subsection{Datasets}
\label{app:datasets}

This subsection expands the dataset overview in Section~\ref{sec:experimental-setup}: the contents and characteristics of the source tables, followed by preprocessing and partitioning.

\subsubsection{Contents and Characteristics}

\textbf{Wine Quality.}\quad
The table combines red and white wines described by eleven physicochemical measurements and one wine-color category \citep{cortez2009modeling}. Its small size and complete observations provide a compact, predominantly numerical case.

\textbf{Online Shoppers.}\quad
Browsing sessions combine page counts, durations, and behavioral rates with calendar, visitor, device, and traffic categories \citep{sakar2019prediction}. The table is complete and mixes numerical and categorical attributes within a relatively small collection of records.

\textbf{Credit Default.}\quad
Client attributes accompany six months of billing and repayment history \citep{yeh2009comparisons}. Numerical amounts and categorical repayment states form repeated monthly attributes, providing structured dependencies within complete records.

\textbf{Amazon Employee.}\quad
Resource-access records describe resources, managers, and employee roles through eight categorical attributes \citep{hamner2013amazon}. This purely categorical table contains 15,283 categories in total, including more than 7,500 in the resource field alone. Large vocabularies increase the cost of joint prior estimation and conditional inference, and require prediction layers with one output per category.

\textbf{Bank Marketing.}\quad
Customer attributes, campaign information, and previous-contact history combine numerical and categorical fields \citep{moro2014telemarketing}. Native missingness is uneven, particularly in credit-default status and previous-contact information.

\textbf{Adult.}\quad
Census records describe demographic, employment, and income attributes \citep{becker1996adult}. The retained table is predominantly categorical, with native unknowns in employment and country-of-origin fields.

\textbf{Diamonds.}\quad
Physical measurements and prices are accompanied by categorical cut, color, and clarity grades \citep{wickham2016ggplot2}. The table is nearly complete and offers numerical relationships alongside a small set of grading categories.

\textbf{SoilFormer.}\quad
The LUCAS-based table combines soil measurements, mapped properties, and environmental covariates with uneven coverage across records \citep{leng2026lucas}. Four numerical fields are vectors: field-capacity and wilting-point profiles each contain seven depths, while precipitation and temperature each contain twelve monthly values.

\textbf{APS Failure.}\quad
Anonymized truck measurements accompany a failure category \citep{scania2016aps}. The numerical attributes comprise 100 scalars and seven ten-bin histograms, giving 107 numerical fields with 170 components. Missingness varies substantially across fields, exceeding 80\% for one measurement.

\textbf{Diabetes.}\quad
Hospital encounters contain demographic, diagnostic, medication, and utilization attributes \citep{strack2014impact}. Most retained fields are categorical, including age and weight bands, diagnosis groups, and medication states. Availability is highly uneven: weight is missing in approximately 97\% of encounters.

\textbf{FLUXNET.}\quad
Half-hourly meteorological and ecosystem-flux records come from three Australian tower sites: Alpine Peatland, Collie, and Emerald \citep{grover2026alpine,beringer2026collie,feitz2026emerald}. This numerical-only table covers wetland, forest, and grassland conditions. Observation gaps vary across variables; net ecosystem carbon exchange is unavailable in approximately 62\% of retained records.

\subsubsection{Preprocessing and Partitioning}

\textbf{Feature Selection.}\quad All source records are retained apart from the FLUXNET filtering described below. We exclude row identifiers, Adult's survey weight and duplicate education encoding, Amazon's redundant role code, and Diabetes' two constant medication columns and deterministic medication-presence summary. The expert quality score in Wine Quality and approval outcome in Amazon Employee are omitted; other source outcome fields remain table attributes. SoilFormer's selected schema excludes suitability scores and saturated hydraulic conductivity, whose supplied values are affected by an upstream encoding error.

\textbf{Numerical Scaling.}\quad Each numerical component is standardized using the mean and population standard deviation of its observed training values. The same parameters are applied to validation and test values; constant training components use unit scale. Vector components are standardized separately while retaining their depth, month, or histogram-bin order. No additional logarithmic transformation or clipping is applied.

\textbf{Categorical Harmonization.}\quad Categorical codes denote labels rather than numerical distances. We merge synonymous WRB soil-class labels in SoilFormer and map each Diabetes diagnosis field to the nine diagnostic groups of \citet{strack2014impact}. Source vocabularies otherwise retain their categories, including levels absent from training. Diabetes' weight, payer, and medical-specialty fields use training vocabularies with a reserved unseen category, which remains distinct from a missing value.

\textbf{Native Missing Values.}\quad Source missing-state indicators become N.O.: \texttt{?} in Adult and Diabetes, \texttt{unknown} in Bank Marketing, \texttt{na} in APS Failure, and $-9999$ in FLUXNET. Bank Marketing's elapsed-time sentinel $999$, unavailable administrative codes in Diabetes, and laboratory entries indicating no measurement are handled similarly. Zero diamond dimensions and all-zero sand--silt--clay compositions in SoilFormer also become N.O.; valid zeros remain observed. For FLUXNET fields with quality flags, only direct observations (\texttt{QC=0}) are retained as observed values; supplied gap-filled values become N.O. Records with fewer than two observed selected fields are removed.

\textbf{Source Inapplicability.}\quad We interpret a few relationships in the original records as N.A.: previous-contact time and outcome in Bank Marketing when there has been no previous contact; occupation in Adult for never-worked records with unknown occupation; and secondary-diagnosis slots beyond the recorded diagnosis count in Diabetes. Other unavailable entries remain N.O.

\textbf{Data Partitions.}\quad All tables except FLUXNET allocate groups to training, validation, and test in proportions of 70\%/15\%/15\%. Grouping prevents repeated or linked records from crossing partitions. Records with identical predictors and missing states stay together; source outcomes and Diamonds' reported depth are excluded from this comparison. Groups also connect encounters from the same Diabetes patient and SoilFormer records with repeated sampling points or matching coordinates. Repeated rows are retained, and the original Adult and APS Failure partitions are pooled before this allocation. FLUXNET instead uses chronological partitions within each site: the last available year is test, the preceding year is validation, and earlier years are training. The first seven days of each validation and test year are removed to separate adjacent partitions. Site identities and timestamps are metadata rather than features. Table~\ref{tab:datasets} gives the resulting record counts.

\begin{table}[tb]
\centering
\caption{Records after preprocessing and partitioning. Group proportions need not equal row proportions.}
\label{tab:datasets}
\small
\setlength{\tabcolsep}{3pt}
\begin{tabular*}{\linewidth}{@{\extracolsep{\fill}}lrrr@{}}
\toprule
Dataset & Train & Validation & Test\\
\midrule
Wine Quality & 4,553 & 965 & 979\\
Online Shoppers & 8,626 & 1,859 & 1,845\\
Credit Default & 20,991 & 4,503 & 4,506\\
Amazon Employee & 22,938 & 4,915 & 4,916\\
Bank Marketing & 28,830 & 6,180 & 6,178\\
Adult & 34,208 & 7,335 & 7,299\\
Diamonds & 37,744 & 8,103 & 8,093\\
SoilFormer & 45,267 & 9,660 & 9,777\\
APS Failure & 53,200 & 11,400 & 11,400\\
Diabetes & 71,265 & 15,308 & 15,193\\
FLUXNET & 70,162 & 48,692 & 38,406\\
\bottomrule
\end{tabular*}
\end{table}

\subsection{Input Heads}
\label{app:input-heads}

The input heads differ in what an incomplete cell contributes to prediction, while sharing the backbones and reconstruction task of Section~\ref{sec:experimental-setup}. All heads except EAct use the same form of column-wise value encoder: a two-layer ReLU MLP for numerical values and an embedding lookup for categories. EAct uses a joint expected-activation layer. The four baselines merge N.O. and N.A. into one missing state.

\subsubsection{Context Exclusion (CExcl)}

CExcl prevents missing cells from contributing context at any backbone layer, following the exclusion principle used in attention masking \citep[e.g.,][]{caruso2026naim}. Available values retain their ordinary encodings. Missing values supply no value content, while their column positions remain available for prediction.

\subsubsection{Missing-state Embedding (MsEmb)}

MsEmb represents a missing value in column $j$ by an independently learnable vector $b_j^{\mathrm{miss}}\in\R^D$, shared across records \citep[e.g.,][]{somepalli2021saint}. Available cells retain $e_j(x_{ij})$. Column identities $c_j$ are added to both, and all positions contribute context, allowing the backbone to learn from missingness itself.

\subsubsection{Expected Activation (EAct)}

EAct \citep{smieja2018processing} integrates uncertainty about missing inputs into the first neural layer. A record is represented by $y\in\R^L$, comprising numerical components and one-hot categorical coordinates. An eight-component diagonal Gaussian mixture $\kappa$ is fitted by expectation--maximization on the training data, integrating missing coordinates out of the likelihood. The fitted distribution is shared across backbones and seeds. Its Gaussian description of one-hot categories is a continuous approximation.

Conditioning $\kappa$ on record $i$'s available input values gives $\kappa_i$, with those values held fixed. EAct analytically averages the first-layer response over this conditional distribution:
\begin{equation}
 r_i^{\mathrm{EAct}}=\E_{y\sim\kappa_i}\!\left[\operatorname{ReLU}\!\left(W^{\mathrm{EAct}}y+b^{\mathrm{EAct}}\right)\right]\in\R^{MD},
 \label{eq:ea-head-activation}
\end{equation}
where $W^{\mathrm{EAct}}\in\R^{MD\times L}$ and $b^{\mathrm{EAct}}\in\R^{MD}$ are trainable. The activations are reshaped into $M$ tokens and combined with column identities.

EAct's joint layer has $LMD$ trainable weights and propagates the conditional moments of every mixture component. Large categorical vocabularies expand $L$ through one-hot encoding, increasing both parameter count and computation relative to column-wise encoding.

\subsubsection{Extended Gaussian Copula (EGC)}

EGC \citep{zhao2022probabilistic} represents numerical and categorical dependencies through a joint latent Gaussian model. Each numerical component is linked to one Gaussian coordinate through an empirical marginal transformation; a categorical column with $K_j$ classes is linked to $K_j$ Gaussian utilities whose maximum determines the category. The marginal transformations are fitted to observed training values, with smoothed categorical frequencies. The joint latent correlation matrix is estimated on 4,096 training records using Gibbs Monte Carlo expectation--maximization. These fitted quantities are then frozen across backbones and seeds.

For each record, the frozen model is conditioned on the available fields. The estimated conditional latent means are mapped through empirical quantile functions for numerical values and an argmax mapping for categories, producing a single imputation. The resulting values are encoded with a missing-state indicator:
\begin{equation}
 h_{ij}^{\mathrm{EGC}}=e_j(\widetilde x_{ij})+\delta_{ij}^{\mathrm{miss}}b_j^{\mathrm{miss}},
 \label{eq:egc-head-encoding}
\end{equation}
Here $\widetilde x_{ij}$ is the observed or imputed value, $\delta_{ij}^{\mathrm{miss}}$ equals one for a missing input and zero otherwise, and $b_j^{\mathrm{miss}}$ is a learned column-specific vector. The backbone receives $c_j+h_{ij}^{\mathrm{EGC}}$ at every position. The missing-state embedding distinguishes reconstructed values from recorded ones.

Conditional imputation must be repeated as the observed context changes. Its latent dimension $L$ is the number of numerical components plus the sum of categorical cardinalities, so large vocabularies increase both the $O(L^2)$ correlation storage and inference cost. We limit conditional iterations using a budget calibrated to the CPU time of one neural training epoch. The evaluated EGC head is therefore a computationally bounded approximation, whose conditional inference need not reach convergence.

\subsubsection{\nano{}}

Given an estimate $v_{ij}$ and confidence $\alpha_{ij}$, \nano{} applies Eq.~\eqref{eq:gate} with independently learnable N.O. embeddings for both feature types. Absent and zero-confidence estimates have the same N.O. representation. A numerical vector shares one confidence across components. N.A. has a separate learned representation, optionally conditioned on available reasons through Eq.~\eqref{eq:reason}.

To test confidence under controlled conditions, we simulate estimates with known error relative to training-derived references. Numerical reference scales $\tau_j^{\mathrm{ref}}$ and categorical priors $\rho_j^{\mathrm{ref}}$ are fixed across backbones and training seeds, without domain-knowledge adjustments. They calibrate the simulated evidence without constraining the learned N.O. embeddings. Let $\mathcal O_j$ index the training records in which column $j$ is observed and applicable. The numerical formulas below use scalar cells; for vectors, they apply to each component.

For numerical columns, three-fold cross-fitting estimates how accurately each field can be predicted from the remaining available fields. An eight-component mixed numerical/categorical reference model is fitted on two folds and evaluated on the held-out fold. Write $\widehat x_{ij}^{\mathrm{OOF}}$ for its out-of-fold prediction, with both prediction and recorded value in training-standardized coordinates. The reference scale is
\begin{equation}
 \tau_j^{\mathrm{ref}}
 =\sqrt{\frac{1}{|\mathcal O_j|}
 \sum_{i\in\mathcal O_j}
 \left(x_{ij}-\widehat x_{ij}^{\mathrm{OOF}}\right)^2}.
 \label{eq:empirical-reference-scale}
\end{equation}
This is a data-space reference error, distinct from the ideal token-space prior spread $\tau_j$. It captures conditional prediction error, including reference-model error, rather than measurement noise alone. Duplicate or entity groups are kept within folds; FLUXNET uses temporal blocks with separation around held-out intervals.

For the positive-confidence levels in Section~\ref{sec:no-results}, numerical estimates are generated from actively masked values in training-standardized coordinates. For $\alpha_{ij}\in(0,1]$, the data-space approximation to Eq.~\eqref{eq:precision} gives
\begin{equation}
 \begin{aligned}
 \sigma_{ij}^{\mathrm{sim}}
 &=\tau_j^{\mathrm{ref}}\sqrt{\frac{1-\alpha_{ij}}{\alpha_{ij}}},\\
 v_{ij}&=x_{ij}+\sigma_{ij}^{\mathrm{sim}}\eta_{ij},
 \quad \eta_{ij}\sim\normal(0,1),
 \end{aligned}
 \label{eq:simulated-numerical-estimate}
\end{equation}
with independent standard-normal draws across components. Thus confidence controls the error relative to the empirical reference scale: lower confidence produces more dispersed estimates. A vector shares one confidence while retaining component-specific noise scales.

For categorical columns, the prior is the empirical frequency
\begin{equation}
 \rho_j^{\mathrm{ref}}(k)
 =\frac{1}{|\mathcal O_j|}
 \sum_{i\in\mathcal O_j}\ind[x_{ij}=k],
 \quad k=1,\ldots,K_j.
 \label{eq:empirical-categorical-prior}
\end{equation}
No smoothing is applied, so categories absent from training retain probability zero.

For a masked categorical value $x_{ij}$, we instantiate the reference channel in Eq.~\eqref{eq:cat-candidate-distribution} with $\rho_j^{\mathrm{ref}}$ and draw $v_{ij}\sim q_{ij}^{\mathrm{sim}}$ from
\begin{equation}
 q_{ij}^{\mathrm{sim}}(k)=\alpha_{ij}\ind[k=x_{ij}]
 +(1-\alpha_{ij})\rho_j^{\mathrm{ref}}(k).
 \label{eq:simulated-categorical-estimate}
\end{equation}
The true category is retained with probability $\alpha_{ij}$; otherwise, a category is sampled independently from the empirical prior. This draw may coincide with the true category, so the probability of a correct estimate is $\alpha_{ij}+(1-\alpha_{ij})\rho_j^{\mathrm{ref}}(x_{ij})$. In the four main confidence conditions of Section~\ref{sec:no-results}, declared confidence equals the generating value of $\alpha_{ij}$ for both feature types. Higher levels therefore pair more reliable estimates with greater input weight; \texttt{CL-Zero} uses the learned N.O. reference alone.

\subsection{Backbones}
\label{app:backbones}

The nine backbones listed in Section~\ref{sec:experimental-setup} are adapted to cell-wise completion through a common input-head interface and numerical/categorical prediction readouts.

\subsubsection{Architecture Overview}

\textbf{FT-Transformer.}\quad Feature self-attention and gated feed-forward layers learn dependencies between columns \citep{gorishniy2021revisiting}. Separate readouts at each column replace the original record-level prediction head.

\textbf{NPT.}\quad Attention alternates between records and between features within a record \citep{kossen2021self}. Each target record attends to itself and a fixed reference panel of 32 training records. The panel is not updated by target records, preventing information exchange between them. Panel records and their duplicate groups are not used as training targets. The panel contains only observed, applicable values and is shared with SAINT and GRAPE across input heads and seeds.

\textbf{NAIM.}\quad The original model uses masked attention to suppress contributions from missing inputs \citep{caruso2026naim}. We retain its bias-free attention and flattened row readout, while allowing the input head to determine which cells contribute context.

\textbf{SAINT.}\quad Column attention and intersample attention combine within-record and cross-record information \citep{somepalli2021saint}. The fixed reference-panel protocol described for NPT is used for intersample attention. The completion model is trained without SAINT's optional contrastive pretraining.

\textbf{ReMasker.}\quad A Transformer encoder processes context tokens, and a decoder reconstructs all column positions \citep{du2024remasker}. The input head determines the encoder context, while every column retains a decoder query, supporting both excluded inputs and missing-state embeddings. The common completion task replaces the original remasking routine.

\textbf{PMAE-derived Mixer.}\quad The encoder and decoder retain residual mixing across token and channel dimensions from PMAE-mix \citep{kim2025pmae,tolstikhin2021mixer}. The input head determines which cells contribute context throughout mixing, while decoder queries retain all column positions. The common completion task replaces proportional remasking.

\textbf{TabNet.}\quad Sequential sparse selection focuses successive decision steps on different features \citep{arik2021tabnet}. Selection is adapted to logical cells: one selection weight applies to all dimensions of a cell embedding. The encoder--decoder structure and sparse-selection penalty are retained.

\textbf{VAEAC.}\quad A conditional latent-variable model combines a context-dependent prior, a training-only approximate posterior, and a decoder \citep{ivanov2019vaeac}. Input-head embeddings supply the conditioning context. Only the approximate posterior receives reconstruction targets; prediction uses the conditional prior and decoder. At inference, 16 latent samples define an equally weighted mixture of numerical distributions and an average of categorical probabilities.

\textbf{GRAPE.}\quad Records and columns form a bipartite graph, with cells represented on edges \citep{you2020handling}. The input head determines edge representations and which edges contribute messages. Each query uses the fixed training panel independently, preventing information exchange between held-out records and allowing inductive completion.

\subsubsection{Model Sizes}

Backbone depth is fixed across datasets, while the base width $D_{\mathrm{base}}\in\{32,48,64\}$ increases with record count (Table~\ref{tab:backbones}). Most backbones use token width $D_{\mathrm{base}}$; NPT and SAINT use $D_{\mathrm{base}}/2$, with NPT's row representation capped at 1,024 dimensions. Within each dataset, each backbone retains the same dimensions across input heads and seeds.

Table~\ref{tab:backbones} separates representation and prediction-readout parameters, excluding input heads and counting VAEAC's approximate posterior with its representation layers. Each cell's readout predicts numerical means or class logits together with one uncertainty scalar. Large categorical vocabularies therefore increase prediction-layer size even when the representation layers remain small.

\begin{table*}[t]
\centering
\caption{Trainable parameters in thousands, shown as representation layers $+$ prediction readouts, rounded separately to integers. A displayed zero denotes fewer than 500 parameters. $D_{\mathrm{base}}$ is the base width. Highlighted readout counts show the large prediction layers needed for Amazon Employee's 15,283 categories.}
\label{tab:backbones}
\fontsize{8pt}{10pt}\selectfont
\setlength{\tabcolsep}{0pt}
\begin{tabular*}{\textwidth}{@{\extracolsep{\fill}}lrrrrrrrrrr@{}}
\toprule
Dataset & $D_{\mathrm{base}}$ & \shortstack{FT-\\Transformer} & NPT & NAIM & SAINT & ReMasker & \shortstack{PMAE\\Mixer} & TabNet & VAEAC & GRAPE\\
\midrule
Wine Quality & 32 & 21+1 & 239+0 & 17+10 & 195+0 & 27+1 & 16+1 & 70+1 & 280+13 & 22+1\\
Online Shoppers & 32 & 21+3 & 495+2 & 17+56 & 282+2 & 27+3 & 19+3 & 92+3 & 340+70 & 22+3\\
Credit Default & 48 & 47+6 & 1,894+3 & 38+144 & 827+3 & 59+6 & 39+6 & 253+6 & 869+171 & 49+6\\
Amazon Employee & 48 & 47+\textbf{749} & 268+\textbf{382} & 38+\textbf{5,887} & 309+\textbf{382} & 59+\textbf{749} & 32+\textbf{749} & 121+\textbf{749} & 3,480+\textbf{8,945} & 47+\textbf{749}\\
Bank Marketing & 48 & 47+4 & 1,477+2 & 38+83 & 730+2 & 59+4 & 37+4 & 228+4 & 805+100 & 49+4\\
Adult & 48 & 47+6 & 618+3 & 38+74 & 471+3 & 59+6 & 34+6 & 162+6 & 661+98 & 48+6\\
Diamonds & 48 & 47+2 & 391+1 & 38+18 & 374+1 & 59+2 & 33+2 & 137+2 & 589+25 & 48+2\\
SoilFormer & 64 & 84+17 & 5,041+8 & 67+566 & 2,045+9 & 105+17 & 70+17 & 591+17 & 1,897+641 & 87+17\\
APS Failure & 64 & 84+18 & 4,187+3 & 67+1,936 & 6,305+9 & 105+18 & 199+18 & 1,669+18 & 4,365+2,038 & 96+18\\
Diabetes & 64 & 84+22 & 4,528+7 & 67+988 & 2,678+11 & 105+22 & 81+22 & 751+22 & 2,284+1,091 & 88+22\\
FLUXNET & 64 & 84+2 & 952+1 & 67+18 & 778+1 & 105+2 & 57+2 & 271+2 & 1,105+25 & 84+2\\
\bottomrule
\end{tabular*}
\end{table*}

\subsection{Loss and Metrics}
\label{app:loss-metrics}

This subsection expands the loss and evaluation criteria in Section~\ref{sec:experimental-setup}, including the categorical objective, uncertainty bounds, and metrics beyond point accuracy.

\subsubsection{Loss Functions}

\textbf{Cell-Specific Uncertainty.}\quad Reconstruction difficulty depends on both the feature and the context available in a record. Alongside a numerical mean $\hat\mu_{ij}$ or categorical probabilities $\hat p_{ij}$, the readout predicts an uncertainty scalar $\hat\zeta_{ij}$ for each cell. This allows heteroscedastic loss attenuation \citep{kendall2017uncertainties}: reconstructions can receive different weights even within the same column. Output uncertainty describes the model's prediction, whereas input confidence $\alpha_{ij}$ describes the reliability of supplied evidence.

\textbf{Reconstruction Losses.}\quad The full numerical and categorical objectives corresponding to Eq.~\eqref{eq:reconstruction-loss} are
\begin{equation}
 \ell_{ij}=
 \begin{cases}
 e^{-\hat\zeta_{ij}}\dfrac{\|x_{ij}-\hat\mu_{ij}\|_2^2}{D_j}+\hat\zeta_{ij},&\text{numerical},\\
 -e^{-\hat\zeta_{ij}}\log\hat p_{ij}(x_{ij})+\hat\zeta_{ij},&\text{categorical}.
 \end{cases}
 \label{eq:full-reconstruction-loss}
\end{equation}
For a numerical cell, $\hat\zeta_{ij}$ is a log variance shared across its components. The numerical loss is twice the dimension-normalized Gaussian NLL up to an additive constant. Dividing squared error by $D_j$ prevents longer vectors from receiving greater weight solely because of their dimension. The factor $e^{-\hat\zeta_{ij}}$ attenuates uncertain reconstructions, while the additive term penalizes uncertainty inflation. For a categorical cell, the same tradeoff is applied heuristically to cross-entropy. Its attenuation scalar is not a categorical variance and does not alter the predicted probabilities, so this objective differs from categorical predictive NLL.

Nonvariational backbones average target-cell losses within each feature type and sum the two means. VAEAC instead sums reconstruction terms within each record and averages over latent samples, retaining its variational regularization. Its decoder applies Eq.~\eqref{eq:full-reconstruction-loss} to each sampled latent state. TabNet also retains its sparsity penalty.

\textbf{Soft Bounds.}\quad To prevent extreme variances and attenuation weights, the unconstrained readout $\widetilde\zeta_{ij}$ is transformed as
\begin{equation}
 \hat\zeta_{ij}
 =B_j\tanh\!\left(\frac{\widetilde\zeta_{ij}}{B_j}\right),
 \label{eq:uncertainty-soft-bound}
\end{equation}
with $B_j=4$ for numerical cells and $B_j=2$ for categorical cells, shared across all compared heads. The transformation is approximately linear near zero and smoothly saturates toward the bounds, preserving differentiability throughout optimization.

\subsubsection{Metrics}

\textbf{Point Accuracy.}\quad RMSE measures numerical error in training-standardized units, while C-ERR measures the fraction of categorical target cells predicted incorrectly. Both are computed per column and then averaged equally across columns of the corresponding feature type, preventing fields with more targets from dominating the comparison. For a numerical vector, squared errors are averaged across components before computing the column RMSE. We also report pooled RMSE and categorical accuracy, which weight target cells equally rather than columns. Balanced accuracy first averages recall across categories present among a column's targets and then averages columns, so frequent categories do not dominate the score. All of these metrics assess point predictions independently of predicted uncertainty.

\textbf{Predictive Uncertainty.}\quad NLL evaluates the predictive density at a recorded numerical value or the probability assigned to a recorded category, penalizing confident errors. Let $\hat f_{ij}$ be the numerical predictive density and $\hat p_{ij}$ the categorical predictive probabilities. We use
\begin{equation}
 \ell_{ij}^{\mathrm{NLL}}=
 \begin{cases}
 -D_j^{-1}\log\hat f_{ij}(x_{ij}),&\text{numerical},\\
 -\log\hat p_{ij}(x_{ij}),&\text{categorical}.
 \end{cases}
 \label{eq:predictive-nll}
\end{equation}
Numerical NLL uses the predicted Gaussian density or, for VAEAC, the predictive mixture density. Categorical NLL uses the probability of the recorded class without the training loss's attenuation factor. The Brier score complements categorical NLL by comparing the full predicted probability vector with a vector that places probability one on the recorded class \citep{gneiting2007proper}:
\begin{equation}
 \mathrm{Brier}_{ij}
 =\sum_{k=1}^{K_j}\left(\hat p_{ij}(k)-\ind[k=x_{ij}]\right)^2.
 \label{eq:categorical-brier}
\end{equation}
Lower scores indicate better probability predictions. The sum is not divided by $K_j$, which would make the score shrink automatically as vocabulary size increases.

A numerical prediction interval is a range derived from the predictive distribution, with nominal coverage $\gamma$ stating how often it should contain the true value. We use $\gamma\in\{.5,.8,.95\}$ and equal-tail intervals, bounded by the $(1-\gamma)/2$ and $(1+\gamma)/2$ quantiles of the predictive marginal, including the mixture marginal for VAEAC. Observed coverage is the fraction of target values inside their intervals; width is the upper bound minus the lower bound, in training-standardized units. Coverage close to $\gamma$ indicates calibration at that level. A narrower interval is more precise, but may miss more true values, so width and coverage must be read together. For vector-valued cells, both are averaged across components before averaging over target cells.

\textbf{Estimate Refinement.}\quad Completion accuracy alone does not show whether a model improves the supplied estimates. We therefore compare prediction and estimate errors on the same target cells. Numerical refinement is estimate MSE minus prediction MSE, with positive values indicating improvement. Relative comparisons divide prediction error by estimate error, so a ratio below one indicates refinement. For categorical estimates, correction is the fraction of incorrect estimates made correct, and corruption is the fraction of correct estimates made incorrect. These rates distinguish correcting errors from introducing new ones.

\textbf{Aggregation.}\quad Distributional scores average target cells within each feature type. Aggregate tables give equal weight to each model across the nine backbones and five seeds, reporting a mean and sample SD that reflect both architecture and seed variation. Backbone-specific summaries average seeds within each architecture. Relative point-error changes compare conditions on the same dataset; NLL uses absolute differences because numerical density NLL can be negative.

In the N.O. study, we exclude extreme prediction failures when \texttt{CL-Zero} test pooled numerical RMSE exceeds both 10 training-standardized units and ten times the corresponding validation RMSE at the selected checkpoint. The entire model is excluded from every metric and inference condition; all other seeds are retained. Across all heads and recipes, twelve models are excluded, all using VAEAC on SoilFormer and none using \nano{-N70}. The separate N.A. trimming rule is given in Appendix~\ref{app:na-results}.

\subsection{Training and Evaluation}
\label{app:training}

This subsection expands the shared masked feature modeling protocol in Section~\ref{sec:experimental-setup}. The confidence mixtures and test levels for the N.O. study are defined in Section~\ref{sec:no-results}.

\textbf{Native Missingness.}\quad Native N.O. and N.A. cells retain their states in the input table but never provide reconstruction targets: the former lack recorded references, and the latter have no applicable values. Their input representations follow Appendix~\ref{app:input-heads}; applicability itself is not predicted. In the N.O. study, \nano{} uses separate N.A. embeddings without reason conditioning.

\textbf{Active Feature Masking.}\quad We independently hide each applicable observed cell with probability $0.3$, selecting all components of a numerical vector together. Recorded values from these cells provide reconstruction targets. Hiding all selected values simultaneously prevents one target from becoming clean context for another. Baseline heads treat the hidden cells as missing inputs. \nano{} can also receive simulated estimates with confidence through Eq.~\eqref{eq:gate}, calibrated by Eqs.~\eqref{eq:simulated-numerical-estimate} and~\eqref{eq:simulated-categorical-estimate}. This separates learning from recorded targets from the treatment of native missingness.

\textbf{Training.}\quad Masks are redrawn each epoch so that different observed subsets serve as context and targets. For a given dataset, backbone, and seed, the input heads share these masks. The common optimization budget is 100 epochs with AdamW, learning rate $10^{-3}$, weight decay $10^{-5}$, and batch size 128. Thus the comparison holds the reconstruction task and training budget fixed while varying input treatment. \nano{} also resamples candidates and confidence each epoch from the mixture in Section~\ref{sec:no-results}.

\textbf{Validation and Test.}\quad Fixed evaluation masks and candidate draws are shared across heads and seeds, and only actively masked cells are scored. Checkpoints are selected by the sum of validation numerical macro RMSE and pooled categorical error expressed as a fraction, omitting an absent feature type. Baselines use \texttt{CL-Zero}; \nano{} averages the selection score equally over the four confidence levels. Each trained model then uses its selected checkpoint for every test condition. Changes across confidence levels therefore measure the response to different evidence without retraining or condition-specific checkpoint selection.

\raggedbottom
\section{Details of Experimental Results}
\label{app:experimental-results}

This appendix extends Section~\ref{sec:experiments} by examining predictive uncertainty, backbone dependence, and refinement of supplied estimates, then describing the N.A. augmentation. The N.O. analyses use the same trained models and masked test cells across inference conditions, following the aggregation and exclusion rules in Appendix~\ref{app:loss-metrics}.

\subsection{Cross-head Comparisons}
\label{app:cross-head-results}

We ask whether the point-accuracy gains in Section~\ref{sec:no-results} extend to predictive distributions and whether they depend on the backbone.

\subsubsection{Comparisons across Metrics}
\label{app:aggregate-nll}

\textbf{NLL.}\quad NLL evaluates prediction accuracy together with output uncertainty, testing whether weak estimates also improve probabilistic prediction. Moving \nano{-N70} from \texttt{CL-Zero} to \texttt{CL-Low} lowers numerical and categorical NLL wherever the corresponding features are present (Table~\ref{tab:aggregate-nll}). With low-confidence estimates, it achieves the lowest numerical NLL on all ten datasets with numerical features and the lowest categorical NLL on seven of the ten datasets with categorical features. The gains from weak estimates therefore extend to probabilistic prediction.

\begin{table*}[p]
\vbox to \textheight{%
\caption{Cross-head NLL, averaged over surviving results from nine backbones and five seeds. Numerical NLL uses training-standardized coordinates and can be negative. Input conditions, exclusions, and highlighting follow Table~\ref{tab:table_abs_failure_filtered}.}
\label{tab:aggregate-nll}
\centering
\begingroup
\small
\newcommand{\noTableDataFont}{\fontsize{8pt}{9pt}\selectfont}
\renewcommand{\arraystretch}{0.88}
% Generated by results_new/no/src/11_manuscript_table.py.
% Regenerate numeric cells; edit captions and spacing in the parent table.
% Data: results_new/no/metrics/table_prob_failure_filtered/main.csv
% Data SHA256: 876028d2049f9bf001dc0704f1ffd953c6fdcc74ceb309c640a07ef33984199e
\begin{tabular*}{\textwidth}{@{\extracolsep{\fill}}lccccc|c@{}}
\toprule
 & \multicolumn{5}{c}{\texttt{CL-Zero} (No Evidence)} & \texttt{CL-Low} \\
\cmidrule(lr){2-6}\cmidrule(l){7-7}
Benchmark & CExcl & MsEmb & EAct & EGC & \nano{-N70} & \nano{-N70} \\
\midrule
\multicolumn{7}{@{}l}{\textit{(a) Numerical features: NLL (nats) $\downarrow$}} \\
\addlinespace[1pt]
\noTableDataFont Adult & \noTableDataFont $1.27\,{\scriptstyle\pm\,0.99}$ & \noTableDataFont $1.31\,{\scriptstyle\pm\,1.05}$ & \noTableDataFont $1.68\,{\scriptstyle\pm\,1.71}$ & \noTableDataFont $1.42\,{\scriptstyle\pm\,1.18}$ & \noTableDataFont $\boldsymbol{1.06}\,{\scriptstyle\pm\,0.10}$ & \noTableDataFont $\boldsymbol{\mathit{0.78}}\,{\scriptstyle\pm\,0.07}$ \\
\noTableDataFont Amazon Employee & \noTableDataFont --- & \noTableDataFont --- & \noTableDataFont --- & \noTableDataFont --- & \noTableDataFont --- & \noTableDataFont --- \\
\noTableDataFont APS Failure & \noTableDataFont $0.45\,{\scriptstyle\pm\,2.36}$ & \noTableDataFont $2.89\,{\scriptstyle\pm\,8.89}$ & \noTableDataFont $\boldsymbol{-0.05}\,{\scriptstyle\pm\,0.33}$ & \noTableDataFont $0.72\,{\scriptstyle\pm\,3.61}$ & \noTableDataFont $0.53\,{\scriptstyle\pm\,1.66}$ & \noTableDataFont $\boldsymbol{\mathit{-0.36}}\,{\scriptstyle\pm\,0.61}$ \\
\noTableDataFont Bank Marketing & \noTableDataFont $-0.08\,{\scriptstyle\pm\,0.24}$ & \noTableDataFont $-0.08\,{\scriptstyle\pm\,0.31}$ & \noTableDataFont $-0.04\,{\scriptstyle\pm\,0.11}$ & \noTableDataFont $-0.10\,{\scriptstyle\pm\,0.14}$ & \noTableDataFont $\boldsymbol{-0.16}\,{\scriptstyle\pm\,0.10}$ & \noTableDataFont $\boldsymbol{\mathit{-0.24}}\,{\scriptstyle\pm\,0.10}$ \\
\noTableDataFont Credit Default & \noTableDataFont $\boldsymbol{0.11}\,{\scriptstyle\pm\,0.11}$ & \noTableDataFont $0.12\,{\scriptstyle\pm\,0.12}$ & \noTableDataFont $0.25\,{\scriptstyle\pm\,0.21}$ & \noTableDataFont $0.15\,{\scriptstyle\pm\,0.09}$ & \noTableDataFont $0.14\,{\scriptstyle\pm\,0.12}$ & \noTableDataFont $\boldsymbol{\mathit{-0.02}}\,{\scriptstyle\pm\,0.11}$ \\
\noTableDataFont Diabetes & \noTableDataFont $1.14\,{\scriptstyle\pm\,0.35}$ & \noTableDataFont $\boldsymbol{1.07}\,{\scriptstyle\pm\,0.09}$ & \noTableDataFont $1.21\,{\scriptstyle\pm\,0.41}$ & \noTableDataFont $\dagger$ & \noTableDataFont $\boldsymbol{1.07}\,{\scriptstyle\pm\,0.07}$ & \noTableDataFont $\boldsymbol{\mathit{0.91}}\,{\scriptstyle\pm\,0.07}$ \\
\noTableDataFont Diamonds & \noTableDataFont $\boldsymbol{-0.46}\,{\scriptstyle\pm\,0.07}$ & \noTableDataFont $\boldsymbol{-0.46}\,{\scriptstyle\pm\,0.06}$ & \noTableDataFont $-0.37\,{\scriptstyle\pm\,0.09}$ & \noTableDataFont $-0.45\,{\scriptstyle\pm\,0.07}$ & \noTableDataFont $-0.44\,{\scriptstyle\pm\,0.09}$ & \noTableDataFont $\boldsymbol{\mathit{-0.47}}\,{\scriptstyle\pm\,0.07}$ \\
\noTableDataFont FLUXNET & \noTableDataFont $0.18\,{\scriptstyle\pm\,0.11}$ & \noTableDataFont $0.19\,{\scriptstyle\pm\,0.11}$ & \noTableDataFont $1.12\,{\scriptstyle\pm\,5.87}$ & \noTableDataFont $\boldsymbol{0.17}\,{\scriptstyle\pm\,0.09}$ & \noTableDataFont $0.18\,{\scriptstyle\pm\,0.08}$ & \noTableDataFont $\boldsymbol{\mathit{0.09}}\,{\scriptstyle\pm\,0.07}$ \\
\noTableDataFont Online Shoppers & \noTableDataFont $\boldsymbol{0.35}\,{\scriptstyle\pm\,0.17}$ & \noTableDataFont $0.37\,{\scriptstyle\pm\,0.19}$ & \noTableDataFont $0.52\,{\scriptstyle\pm\,0.13}$ & \noTableDataFont $\boldsymbol{0.35}\,{\scriptstyle\pm\,0.15}$ & \noTableDataFont $0.37\,{\scriptstyle\pm\,0.19}$ & \noTableDataFont $\boldsymbol{\mathit{0.23}}\,{\scriptstyle\pm\,0.18}$ \\
\noTableDataFont SoilFormer & \noTableDataFont $0.17\,{\scriptstyle\pm\,0.53}$ & \noTableDataFont $0.94\,{\scriptstyle\pm\,5.68}$ & \noTableDataFont $10.01\,{\scriptstyle\pm\,63.87}$ & \noTableDataFont $\boldsymbol{0.09}\,{\scriptstyle\pm\,0.12}$ & \noTableDataFont $0.12\,{\scriptstyle\pm\,0.15}$ & \noTableDataFont $\boldsymbol{\mathit{0.06}}\,{\scriptstyle\pm\,0.13}$ \\
\noTableDataFont Wine Quality & \noTableDataFont $0.88\,{\scriptstyle\pm\,0.10}$ & \noTableDataFont $0.89\,{\scriptstyle\pm\,0.13}$ & \noTableDataFont $0.87\,{\scriptstyle\pm\,0.11}$ & \noTableDataFont $\boldsymbol{0.86}\,{\scriptstyle\pm\,0.08}$ & \noTableDataFont $0.88\,{\scriptstyle\pm\,0.11}$ & \noTableDataFont $\boldsymbol{\mathit{0.80}}\,{\scriptstyle\pm\,0.12}$ \\
\midrule
\multicolumn{7}{@{}l}{\textit{(b) Categorical features: NLL (nats) $\downarrow$}} \\
\addlinespace[1pt]
\noTableDataFont Adult & \noTableDataFont $\boldsymbol{0.867}\,{\scriptstyle\pm\,0.018}$ & \noTableDataFont $0.869\,{\scriptstyle\pm\,0.025}$ & \noTableDataFont $0.877\,{\scriptstyle\pm\,0.029}$ & \noTableDataFont $0.871\,{\scriptstyle\pm\,0.025}$ & \noTableDataFont $0.875\,{\scriptstyle\pm\,0.028}$ & \noTableDataFont $\boldsymbol{\mathit{0.852}}\,{\scriptstyle\pm\,0.034}$ \\
\noTableDataFont Amazon Employee & \noTableDataFont $2.452\,{\scriptstyle\pm\,0.412}$ & \noTableDataFont $2.459\,{\scriptstyle\pm\,0.436}$ & \noTableDataFont $\boldsymbol{2.311}\,{\scriptstyle\pm\,0.392}$ & \noTableDataFont $\dagger$ & \noTableDataFont $2.477\,{\scriptstyle\pm\,0.438}$ & \noTableDataFont $2.442\,{\scriptstyle\pm\,0.451}$ \\
\noTableDataFont APS Failure & \noTableDataFont $0.026\,{\scriptstyle\pm\,0.003}$ & \noTableDataFont $0.026\,{\scriptstyle\pm\,0.003}$ & \noTableDataFont $0.027\,{\scriptstyle\pm\,0.007}$ & \noTableDataFont $\boldsymbol{0.025}\,{\scriptstyle\pm\,0.004}$ & \noTableDataFont $0.026\,{\scriptstyle\pm\,0.003}$ & \noTableDataFont $\boldsymbol{\mathit{0.024}}\,{\scriptstyle\pm\,0.003}$ \\
\noTableDataFont Bank Marketing & \noTableDataFont $\boldsymbol{0.726}\,{\scriptstyle\pm\,0.036}$ & \noTableDataFont $0.730\,{\scriptstyle\pm\,0.047}$ & \noTableDataFont $0.733\,{\scriptstyle\pm\,0.031}$ & \noTableDataFont $\boldsymbol{0.726}\,{\scriptstyle\pm\,0.033}$ & \noTableDataFont $0.728\,{\scriptstyle\pm\,0.036}$ & \noTableDataFont $\boldsymbol{\mathit{0.710}}\,{\scriptstyle\pm\,0.043}$ \\
\noTableDataFont Credit Default & \noTableDataFont $0.519\,{\scriptstyle\pm\,0.047}$ & \noTableDataFont $0.522\,{\scriptstyle\pm\,0.050}$ & \noTableDataFont $0.540\,{\scriptstyle\pm\,0.028}$ & \noTableDataFont $\boldsymbol{0.517}\,{\scriptstyle\pm\,0.038}$ & \noTableDataFont $0.527\,{\scriptstyle\pm\,0.046}$ & \noTableDataFont $\boldsymbol{\mathit{0.517}}\,{\scriptstyle\pm\,0.050}$ \\
\noTableDataFont Diabetes & \noTableDataFont $0.493\,{\scriptstyle\pm\,0.016}$ & \noTableDataFont $\boldsymbol{0.491}\,{\scriptstyle\pm\,0.016}$ & \noTableDataFont $0.494\,{\scriptstyle\pm\,0.020}$ & \noTableDataFont $\dagger$ & \noTableDataFont $0.493\,{\scriptstyle\pm\,0.016}$ & \noTableDataFont $\boldsymbol{\mathit{0.481}}\,{\scriptstyle\pm\,0.019}$ \\
\noTableDataFont Diamonds & \noTableDataFont $1.241\,{\scriptstyle\pm\,0.086}$ & \noTableDataFont $1.246\,{\scriptstyle\pm\,0.079}$ & \noTableDataFont $1.277\,{\scriptstyle\pm\,0.079}$ & \noTableDataFont $\boldsymbol{1.232}\,{\scriptstyle\pm\,0.058}$ & \noTableDataFont $1.274\,{\scriptstyle\pm\,0.092}$ & \noTableDataFont $1.243\,{\scriptstyle\pm\,0.101}$ \\
\noTableDataFont FLUXNET & \noTableDataFont --- & \noTableDataFont --- & \noTableDataFont --- & \noTableDataFont --- & \noTableDataFont --- & \noTableDataFont --- \\
\noTableDataFont Online Shoppers & \noTableDataFont $1.022\,{\scriptstyle\pm\,0.061}$ & \noTableDataFont $1.022\,{\scriptstyle\pm\,0.064}$ & \noTableDataFont $\boldsymbol{1.016}\,{\scriptstyle\pm\,0.050}$ & \noTableDataFont $1.019\,{\scriptstyle\pm\,0.056}$ & \noTableDataFont $1.026\,{\scriptstyle\pm\,0.059}$ & \noTableDataFont $\boldsymbol{\mathit{1.001}}\,{\scriptstyle\pm\,0.071}$ \\
\noTableDataFont SoilFormer & \noTableDataFont $\boldsymbol{0.896}\,{\scriptstyle\pm\,0.061}$ & \noTableDataFont $0.900\,{\scriptstyle\pm\,0.067}$ & \noTableDataFont $1.091\,{\scriptstyle\pm\,0.551}$ & \noTableDataFont $0.915\,{\scriptstyle\pm\,0.074}$ & \noTableDataFont $0.908\,{\scriptstyle\pm\,0.058}$ & \noTableDataFont $\boldsymbol{\mathit{0.874}}\,{\scriptstyle\pm\,0.069}$ \\
\noTableDataFont Wine Quality & \noTableDataFont $0.040\,{\scriptstyle\pm\,0.012}$ & \noTableDataFont $0.043\,{\scriptstyle\pm\,0.018}$ & \noTableDataFont $0.040\,{\scriptstyle\pm\,0.009}$ & \noTableDataFont $\boldsymbol{0.028}\,{\scriptstyle\pm\,0.006}$ & \noTableDataFont $0.043\,{\scriptstyle\pm\,0.013}$ & \noTableDataFont $0.034\,{\scriptstyle\pm\,0.011}$ \\
\bottomrule
\end{tabular*}

\par
\endgroup

\vfill
\begin{minipage}{\textwidth}
\makeatletter
\def\@captype{figure}
\makeatother
\setcaptiontype{figure}
\centering
\includegraphics[width=\textwidth]{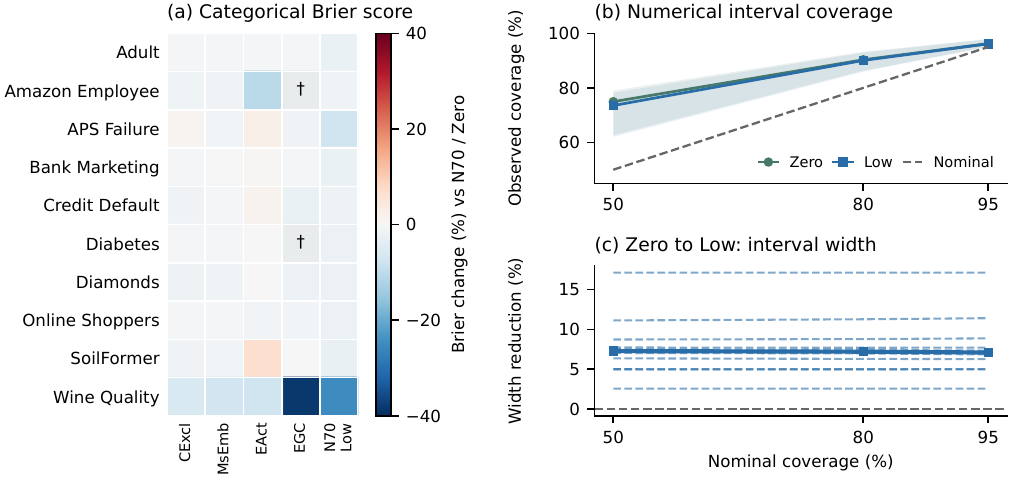}
\caption{Additional cross-head uncertainty metrics. (a) Brier-score changes relative to \nano{-N70} at \texttt{CL-Zero}; blue indicates lower scores. The four baselines use \texttt{CL-Zero}, and N70/Low uses the same \nano{-N70} models with low-confidence estimates. $\dagger$ marks unsuccessful prior fitting. (b) Numerical coverage for \nano{-N70}: lines show medians across datasets, shaded bands show interquartile ranges, and the dashed line shows nominal coverage. (c) Interval-width reductions from Zero to Low: dashed lines show individual datasets and the dark line their median. Model-run scores are averaged within each dataset before these comparisons.}
\label{fig:cross-head-uncertainty}
\end{minipage}
\par\vskip0pt
}
\end{table*}

\textbf{Categorical Brier Scores.}\quad The Brier score evaluates the full class-probability vector, providing a complementary check to NLL. Low-confidence estimates lower \nano{-N70}'s Brier score on all ten datasets with categorical features, giving the lowest score among the compared heads on six (Figure~\ref{fig:cross-head-uncertainty}(a)). The categorical gains therefore persist under two different probability scores rather than depending on NLL alone.

\textbf{Numerical Intervals.}\quad A prediction interval gives a range intended to contain the true numerical value at a stated frequency. Its width measures how tightly the prediction locates that value, while coverage measures how often the interval actually contains it. Low-confidence estimates narrow \nano{-N70}'s intervals on all ten datasets with numerical features at the 50\%, 80\%, and 95\% stated coverage levels (Figure~\ref{fig:cross-head-uncertainty}(b,c)). Median width reductions are about 7\% at all three levels, while coverage changes little. The intervals remain conservative: intervals intended to cover 50\% and 80\% of true values actually cover approximately 74\% and 90\%, respectively, using medians across datasets. Better NLL and narrower intervals therefore do not establish that the learned predictive distributions are fully calibrated.

\subsubsection{Backbone Dependence}
\label{app:backbone-dependence}

The shared input-head comparison tests compatibility across architectures, but it can conceal differences in how backbones use weak evidence. At \texttt{CL-Low}, \nano{-N70} achieves the lowest point error in most comparisons using the five Transformer backbones, the Mixer, and VAEAC. Its relative performance is weaker with TabNet and GRAPE, so this limitation is not shared by non-Transformer architectures generally.

Tables~\ref{tab:remasker-heads} and~\ref{tab:tabnet-heads} examine ReMasker and TabNet, with which \nano{-N70} has its strongest and weakest average ranks among input heads. These ranks give equal weight to datasets, feature types, and the Zero and Low inference conditions, using mean errors over seeds within each comparison. The two examples show how the same input interface can yield different gains from weak evidence.

\textbf{ReMasker.}\quad Without estimates, \nano{-N70} is competitive with the strongest baselines on most datasets. Low-confidence estimates reduce its error in all 20 dataset--feature-type comparisons, with median reductions of 10.4\% in RMSE and 3.6\% in C-ERR. It then attains the lowest RMSE on all ten datasets with numerical features and the lowest C-ERR on nine of the ten datasets with categorical features. ReMasker's reconstruction mechanism can therefore use the additional evidence conveyed by the blended input embeddings.

\textbf{TabNet and GRAPE.}\quad With TabNet, low-confidence estimates reduce error on every dataset for each feature type present, but the median reductions are smaller: 3.7\% for RMSE and 1.1\% for C-ERR. Its weaker relative performance reflects both a less competitive starting point without estimates and smaller gains from weak evidence. GRAPE likewise has a small median C-ERR reduction of 0.9\%, although RMSE improves on all ten datasets with numerical features. Sparse feature selection in TabNet and message averaging in GRAPE may attenuate weak evidence. These interpretations remain tentative because backbone capacity and optimization also differ.

\begin{table*}[p]
\vbox to \textheight{%
\begin{minipage}{\textwidth}
\setcaptiontype{table}
\caption{Cross-head point prediction with ReMasker, averaged over five seeds. Input conditions and highlighting follow Table~\ref{tab:table_abs_failure_filtered}; $\dagger$ indicates unsuccessful EGC prior fitting.}
\label{tab:remasker-heads}
\centering
\small
\newcommand{\noTableDataFont}{\fontsize{8pt}{9pt}\selectfont}
\renewcommand{\arraystretch}{0.88}
% Generated by results_new/no/src/12_backbone_tables.py.
% Backbone: remasker
% Data: results_new/no/metrics/table_abs_failure_filtered/by_torso.csv
% Data SHA256: cb050f9e8cabe51200cb49d307f34be127f1bfa955df342636a4f95da739b77f
\begin{tabular*}{\textwidth}{@{\extracolsep{\fill}}lccccc|c@{}}
\toprule
 & \multicolumn{5}{c}{\texttt{CL-Zero} (No Evidence)} & \texttt{CL-Low} \\
\cmidrule(lr){2-6}\cmidrule(l){7-7}
Benchmark & CExcl & MsEmb & EAct & EGC & \nano{-N70} & \nano{-N70} \\
\midrule
\multicolumn{7}{@{}l}{\textit{(a) Numerical features: macro RMSE in training z-score units $\downarrow$}} \\
\addlinespace[1pt]
\noTableDataFont Adult & \noTableDataFont $0.936\,{\scriptstyle\pm\,0.002}$ & \noTableDataFont $0.934\,{\scriptstyle\pm\,0.001}$ & \noTableDataFont $0.936\,{\scriptstyle\pm\,0.001}$ & \noTableDataFont $0.935\,{\scriptstyle\pm\,0.001}$ & \noTableDataFont $\boldsymbol{0.933}\,{\scriptstyle\pm\,0.001}$ & \noTableDataFont $\boldsymbol{\mathit{0.829}}\,{\scriptstyle\pm\,0.004}$ \\
\noTableDataFont Amazon Employee & \noTableDataFont --- & \noTableDataFont --- & \noTableDataFont --- & \noTableDataFont --- & \noTableDataFont --- & \noTableDataFont --- \\
\noTableDataFont APS Failure & \noTableDataFont $0.662\,{\scriptstyle\pm\,0.004}$ & \noTableDataFont $0.622\,{\scriptstyle\pm\,0.005}$ & \noTableDataFont $0.646\,{\scriptstyle\pm\,0.002}$ & \noTableDataFont $0.624\,{\scriptstyle\pm\,0.003}$ & \noTableDataFont $\boldsymbol{0.615}\,{\scriptstyle\pm\,0.004}$ & \noTableDataFont $\boldsymbol{\mathit{0.590}}\,{\scriptstyle\pm\,0.006}$ \\
\noTableDataFont Bank Marketing & \noTableDataFont $0.481\,{\scriptstyle\pm\,0.005}$ & \noTableDataFont $0.484\,{\scriptstyle\pm\,0.005}$ & \noTableDataFont $0.511\,{\scriptstyle\pm\,0.004}$ & \noTableDataFont $0.488\,{\scriptstyle\pm\,0.003}$ & \noTableDataFont $\boldsymbol{0.471}\,{\scriptstyle\pm\,0.007}$ & \noTableDataFont $\boldsymbol{\mathit{0.415}}\,{\scriptstyle\pm\,0.008}$ \\
\noTableDataFont Credit Default & \noTableDataFont $0.615\,{\scriptstyle\pm\,0.005}$ & \noTableDataFont $\boldsymbol{0.605}\,{\scriptstyle\pm\,0.005}$ & \noTableDataFont $0.635\,{\scriptstyle\pm\,0.004}$ & \noTableDataFont $0.617\,{\scriptstyle\pm\,0.002}$ & \noTableDataFont $0.608\,{\scriptstyle\pm\,0.006}$ & \noTableDataFont $\boldsymbol{\mathit{0.561}}\,{\scriptstyle\pm\,0.005}$ \\
\noTableDataFont Diabetes & \noTableDataFont $0.836\,{\scriptstyle\pm\,0.002}$ & \noTableDataFont $0.835\,{\scriptstyle\pm\,0.001}$ & \noTableDataFont $0.840\,{\scriptstyle\pm\,0.001}$ & \noTableDataFont $\dagger$ & \noTableDataFont $\boldsymbol{0.829}\,{\scriptstyle\pm\,0.001}$ & \noTableDataFont $\boldsymbol{\mathit{0.749}}\,{\scriptstyle\pm\,0.002}$ \\
\noTableDataFont Diamonds & \noTableDataFont $0.255\,{\scriptstyle\pm\,0.001}$ & \noTableDataFont $\boldsymbol{0.253}\,{\scriptstyle\pm\,0.001}$ & \noTableDataFont $0.275\,{\scriptstyle\pm\,0.003}$ & \noTableDataFont $0.260\,{\scriptstyle\pm\,0.003}$ & \noTableDataFont $0.258\,{\scriptstyle\pm\,0.002}$ & \noTableDataFont $\boldsymbol{\mathit{0.224}}\,{\scriptstyle\pm\,0.002}$ \\
\noTableDataFont FLUXNET & \noTableDataFont $0.466\,{\scriptstyle\pm\,0.004}$ & \noTableDataFont $\boldsymbol{0.464}\,{\scriptstyle\pm\,0.004}$ & \noTableDataFont $0.488\,{\scriptstyle\pm\,0.003}$ & \noTableDataFont $0.467\,{\scriptstyle\pm\,0.007}$ & \noTableDataFont $0.472\,{\scriptstyle\pm\,0.002}$ & \noTableDataFont $\boldsymbol{\mathit{0.413}}\,{\scriptstyle\pm\,0.000}$ \\
\noTableDataFont Online Shoppers & \noTableDataFont $0.674\,{\scriptstyle\pm\,0.003}$ & \noTableDataFont $0.672\,{\scriptstyle\pm\,0.005}$ & \noTableDataFont $0.690\,{\scriptstyle\pm\,0.003}$ & \noTableDataFont $\boldsymbol{0.665}\,{\scriptstyle\pm\,0.002}$ & \noTableDataFont $0.675\,{\scriptstyle\pm\,0.003}$ & \noTableDataFont $\boldsymbol{\mathit{0.586}}\,{\scriptstyle\pm\,0.003}$ \\
\noTableDataFont SoilFormer & \noTableDataFont $0.461\,{\scriptstyle\pm\,0.007}$ & \noTableDataFont $\boldsymbol{0.449}\,{\scriptstyle\pm\,0.002}$ & \noTableDataFont $0.472\,{\scriptstyle\pm\,0.002}$ & \noTableDataFont $0.450\,{\scriptstyle\pm\,0.001}$ & \noTableDataFont $0.453\,{\scriptstyle\pm\,0.004}$ & \noTableDataFont $\boldsymbol{\mathit{0.416}}\,{\scriptstyle\pm\,0.003}$ \\
\noTableDataFont Wine Quality & \noTableDataFont $0.666\,{\scriptstyle\pm\,0.003}$ & \noTableDataFont $0.665\,{\scriptstyle\pm\,0.004}$ & \noTableDataFont $0.655\,{\scriptstyle\pm\,0.001}$ & \noTableDataFont $\boldsymbol{0.654}\,{\scriptstyle\pm\,0.001}$ & \noTableDataFont $0.667\,{\scriptstyle\pm\,0.004}$ & \noTableDataFont $\boldsymbol{\mathit{0.615}}\,{\scriptstyle\pm\,0.002}$ \\
\midrule
\multicolumn{7}{@{}l}{\textit{(b) Categorical features: macro C-ERR (\%) $\downarrow$}} \\
\addlinespace[1pt]
\noTableDataFont Adult & \noTableDataFont $30.11\,{\scriptstyle\pm\,0.08}$ & \noTableDataFont $29.87\,{\scriptstyle\pm\,0.08}$ & \noTableDataFont $\boldsymbol{29.72}\,{\scriptstyle\pm\,0.05}$ & \noTableDataFont $\boldsymbol{29.72}\,{\scriptstyle\pm\,0.18}$ & \noTableDataFont $29.74\,{\scriptstyle\pm\,0.10}$ & \noTableDataFont $\boldsymbol{\mathit{28.67}}\,{\scriptstyle\pm\,0.12}$ \\
\noTableDataFont Amazon Employee & \noTableDataFont $37.83\,{\scriptstyle\pm\,0.39}$ & \noTableDataFont $36.53\,{\scriptstyle\pm\,0.34}$ & \noTableDataFont $\boldsymbol{31.89}\,{\scriptstyle\pm\,0.10}$ & \noTableDataFont $\dagger$ & \noTableDataFont $37.54\,{\scriptstyle\pm\,0.32}$ & \noTableDataFont $36.62\,{\scriptstyle\pm\,0.38}$ \\
\noTableDataFont APS Failure & \noTableDataFont $0.77\,{\scriptstyle\pm\,0.06}$ & \noTableDataFont $0.75\,{\scriptstyle\pm\,0.08}$ & \noTableDataFont $\boldsymbol{0.70}\,{\scriptstyle\pm\,0.08}$ & \noTableDataFont $0.79\,{\scriptstyle\pm\,0.13}$ & \noTableDataFont $0.73\,{\scriptstyle\pm\,0.08}$ & \noTableDataFont $\boldsymbol{\mathit{0.70}}\,{\scriptstyle\pm\,0.03}$ \\
\noTableDataFont Bank Marketing & \noTableDataFont $28.09\,{\scriptstyle\pm\,0.07}$ & \noTableDataFont $\boldsymbol{27.95}\,{\scriptstyle\pm\,0.19}$ & \noTableDataFont $28.54\,{\scriptstyle\pm\,0.31}$ & \noTableDataFont $28.20\,{\scriptstyle\pm\,0.21}$ & \noTableDataFont $28.11\,{\scriptstyle\pm\,0.23}$ & \noTableDataFont $\boldsymbol{\mathit{25.81}}\,{\scriptstyle\pm\,0.15}$ \\
\noTableDataFont Credit Default & \noTableDataFont $21.05\,{\scriptstyle\pm\,0.31}$ & \noTableDataFont $\boldsymbol{20.56}\,{\scriptstyle\pm\,0.26}$ & \noTableDataFont $21.75\,{\scriptstyle\pm\,0.15}$ & \noTableDataFont $20.83\,{\scriptstyle\pm\,0.15}$ & \noTableDataFont $20.80\,{\scriptstyle\pm\,0.11}$ & \noTableDataFont $\boldsymbol{\mathit{20.15}}\,{\scriptstyle\pm\,0.06}$ \\
\noTableDataFont Diabetes & \noTableDataFont $22.11\,{\scriptstyle\pm\,0.05}$ & \noTableDataFont $21.97\,{\scriptstyle\pm\,0.07}$ & \noTableDataFont $\boldsymbol{21.77}\,{\scriptstyle\pm\,0.08}$ & \noTableDataFont $\dagger$ & \noTableDataFont $21.92\,{\scriptstyle\pm\,0.14}$ & \noTableDataFont $\boldsymbol{\mathit{21.12}}\,{\scriptstyle\pm\,0.18}$ \\
\noTableDataFont Diamonds & \noTableDataFont $49.80\,{\scriptstyle\pm\,0.48}$ & \noTableDataFont $\boldsymbol{49.32}\,{\scriptstyle\pm\,1.02}$ & \noTableDataFont $52.20\,{\scriptstyle\pm\,0.38}$ & \noTableDataFont $49.75\,{\scriptstyle\pm\,0.36}$ & \noTableDataFont $50.18\,{\scriptstyle\pm\,0.68}$ & \noTableDataFont $\boldsymbol{\mathit{47.61}}\,{\scriptstyle\pm\,0.51}$ \\
\noTableDataFont FLUXNET & \noTableDataFont --- & \noTableDataFont --- & \noTableDataFont --- & \noTableDataFont --- & \noTableDataFont --- & \noTableDataFont --- \\
\noTableDataFont Online Shoppers & \noTableDataFont $35.65\,{\scriptstyle\pm\,0.49}$ & \noTableDataFont $35.35\,{\scriptstyle\pm\,0.21}$ & \noTableDataFont $35.49\,{\scriptstyle\pm\,0.18}$ & \noTableDataFont $\boldsymbol{35.19}\,{\scriptstyle\pm\,0.12}$ & \noTableDataFont $35.65\,{\scriptstyle\pm\,0.40}$ & \noTableDataFont $\boldsymbol{\mathit{34.36}}\,{\scriptstyle\pm\,0.36}$ \\
\noTableDataFont SoilFormer & \noTableDataFont $28.36\,{\scriptstyle\pm\,0.22}$ & \noTableDataFont $\boldsymbol{27.96}\,{\scriptstyle\pm\,0.25}$ & \noTableDataFont $30.78\,{\scriptstyle\pm\,0.17}$ & \noTableDataFont $28.31\,{\scriptstyle\pm\,0.16}$ & \noTableDataFont $28.42\,{\scriptstyle\pm\,0.24}$ & \noTableDataFont $\boldsymbol{\mathit{27.08}}\,{\scriptstyle\pm\,0.06}$ \\
\noTableDataFont Wine Quality & \noTableDataFont $1.22\,{\scriptstyle\pm\,0.32}$ & \noTableDataFont $1.22\,{\scriptstyle\pm\,0.32}$ & \noTableDataFont $1.22\,{\scriptstyle\pm\,0.20}$ & \noTableDataFont $\boldsymbol{0.93}\,{\scriptstyle\pm\,0.32}$ & \noTableDataFont $1.15\,{\scriptstyle\pm\,0.39}$ & \noTableDataFont $\boldsymbol{\mathit{0.79}}\,{\scriptstyle\pm\,0.39}$ \\
\bottomrule
\end{tabular*}

\end{minipage}

\par\vfill
\begin{minipage}{\textwidth}
\setcaptiontype{table}
\caption{Cross-head point prediction with TabNet, averaged over five seeds. Input conditions, highlighting, and symbols follow Table~\ref{tab:remasker-heads}.}
\label{tab:tabnet-heads}
\centering
\small
\newcommand{\noTableDataFont}{\fontsize{8pt}{9pt}\selectfont}
\renewcommand{\arraystretch}{0.88}
% Generated by results_new/no/src/12_backbone_tables.py.
% Backbone: tabnet
% Data: results_new/no/metrics/table_abs_failure_filtered/by_torso.csv
% Data SHA256: cb050f9e8cabe51200cb49d307f34be127f1bfa955df342636a4f95da739b77f
\begin{tabular*}{\textwidth}{@{\extracolsep{\fill}}lccccc|c@{}}
\toprule
 & \multicolumn{5}{c}{\texttt{CL-Zero} (No Evidence)} & \texttt{CL-Low} \\
\cmidrule(lr){2-6}\cmidrule(l){7-7}
Benchmark & CExcl & MsEmb & EAct & EGC & \nano{-N70} & \nano{-N70} \\
\midrule
\multicolumn{7}{@{}l}{\textit{(a) Numerical features: macro RMSE in training z-score units $\downarrow$}} \\
\addlinespace[1pt]
\noTableDataFont Adult & \noTableDataFont $0.945\,{\scriptstyle\pm\,0.002}$ & \noTableDataFont $0.945\,{\scriptstyle\pm\,0.002}$ & \noTableDataFont $\boldsymbol{0.944}\,{\scriptstyle\pm\,0.002}$ & \noTableDataFont $0.945\,{\scriptstyle\pm\,0.003}$ & \noTableDataFont $0.952\,{\scriptstyle\pm\,0.005}$ & \noTableDataFont $\boldsymbol{\mathit{0.914}}\,{\scriptstyle\pm\,0.012}$ \\
\noTableDataFont Amazon Employee & \noTableDataFont --- & \noTableDataFont --- & \noTableDataFont --- & \noTableDataFont --- & \noTableDataFont --- & \noTableDataFont --- \\
\noTableDataFont APS Failure & \noTableDataFont $0.770\,{\scriptstyle\pm\,0.015}$ & \noTableDataFont $0.764\,{\scriptstyle\pm\,0.004}$ & \noTableDataFont $0.762\,{\scriptstyle\pm\,0.006}$ & \noTableDataFont $\boldsymbol{0.749}\,{\scriptstyle\pm\,0.003}$ & \noTableDataFont $0.770\,{\scriptstyle\pm\,0.010}$ & \noTableDataFont $0.766\,{\scriptstyle\pm\,0.007}$ \\
\noTableDataFont Bank Marketing & \noTableDataFont $\boldsymbol{0.526}\,{\scriptstyle\pm\,0.009}$ & \noTableDataFont $0.548\,{\scriptstyle\pm\,0.006}$ & \noTableDataFont $0.527\,{\scriptstyle\pm\,0.005}$ & \noTableDataFont $0.566\,{\scriptstyle\pm\,0.018}$ & \noTableDataFont $0.536\,{\scriptstyle\pm\,0.013}$ & \noTableDataFont $\boldsymbol{\mathit{0.508}}\,{\scriptstyle\pm\,0.017}$ \\
\noTableDataFont Credit Default & \noTableDataFont $0.705\,{\scriptstyle\pm\,0.005}$ & \noTableDataFont $0.707\,{\scriptstyle\pm\,0.012}$ & \noTableDataFont $\boldsymbol{0.692}\,{\scriptstyle\pm\,0.005}$ & \noTableDataFont $0.702\,{\scriptstyle\pm\,0.004}$ & \noTableDataFont $0.709\,{\scriptstyle\pm\,0.009}$ & \noTableDataFont $0.701\,{\scriptstyle\pm\,0.007}$ \\
\noTableDataFont Diabetes & \noTableDataFont $0.854\,{\scriptstyle\pm\,0.003}$ & \noTableDataFont $0.851\,{\scriptstyle\pm\,0.003}$ & \noTableDataFont $\boldsymbol{0.849}\,{\scriptstyle\pm\,0.002}$ & \noTableDataFont $\dagger$ & \noTableDataFont $0.851\,{\scriptstyle\pm\,0.004}$ & \noTableDataFont $\boldsymbol{\mathit{0.795}}\,{\scriptstyle\pm\,0.005}$ \\
\noTableDataFont Diamonds & \noTableDataFont $0.293\,{\scriptstyle\pm\,0.004}$ & \noTableDataFont $0.287\,{\scriptstyle\pm\,0.003}$ & \noTableDataFont $0.289\,{\scriptstyle\pm\,0.004}$ & \noTableDataFont $\boldsymbol{0.279}\,{\scriptstyle\pm\,0.005}$ & \noTableDataFont $0.292\,{\scriptstyle\pm\,0.006}$ & \noTableDataFont $\boldsymbol{\mathit{0.262}}\,{\scriptstyle\pm\,0.005}$ \\
\noTableDataFont FLUXNET & \noTableDataFont $0.525\,{\scriptstyle\pm\,0.026}$ & \noTableDataFont $\boldsymbol{0.507}\,{\scriptstyle\pm\,0.006}$ & \noTableDataFont $0.517\,{\scriptstyle\pm\,0.011}$ & \noTableDataFont $0.526\,{\scriptstyle\pm\,0.010}$ & \noTableDataFont $0.526\,{\scriptstyle\pm\,0.017}$ & \noTableDataFont $\boldsymbol{\mathit{0.474}}\,{\scriptstyle\pm\,0.016}$ \\
\noTableDataFont Online Shoppers & \noTableDataFont $0.748\,{\scriptstyle\pm\,0.015}$ & \noTableDataFont $0.761\,{\scriptstyle\pm\,0.019}$ & \noTableDataFont $0.756\,{\scriptstyle\pm\,0.008}$ & \noTableDataFont $\boldsymbol{0.715}\,{\scriptstyle\pm\,0.004}$ & \noTableDataFont $0.771\,{\scriptstyle\pm\,0.017}$ & \noTableDataFont $0.747\,{\scriptstyle\pm\,0.020}$ \\
\noTableDataFont SoilFormer & \noTableDataFont $0.543\,{\scriptstyle\pm\,0.008}$ & \noTableDataFont $0.554\,{\scriptstyle\pm\,0.004}$ & \noTableDataFont $0.546\,{\scriptstyle\pm\,0.008}$ & \noTableDataFont $\boldsymbol{0.535}\,{\scriptstyle\pm\,0.005}$ & \noTableDataFont $0.559\,{\scriptstyle\pm\,0.005}$ & \noTableDataFont $0.545\,{\scriptstyle\pm\,0.005}$ \\
\noTableDataFont Wine Quality & \noTableDataFont $0.727\,{\scriptstyle\pm\,0.012}$ & \noTableDataFont $0.735\,{\scriptstyle\pm\,0.012}$ & \noTableDataFont $\boldsymbol{0.675}\,{\scriptstyle\pm\,0.005}$ & \noTableDataFont $0.689\,{\scriptstyle\pm\,0.010}$ & \noTableDataFont $0.718\,{\scriptstyle\pm\,0.006}$ & \noTableDataFont $0.694\,{\scriptstyle\pm\,0.007}$ \\
\midrule
\multicolumn{7}{@{}l}{\textit{(b) Categorical features: macro C-ERR (\%) $\downarrow$}} \\
\addlinespace[1pt]
\noTableDataFont Adult & \noTableDataFont $30.38\,{\scriptstyle\pm\,0.17}$ & \noTableDataFont $30.83\,{\scriptstyle\pm\,0.22}$ & \noTableDataFont $\boldsymbol{30.31}\,{\scriptstyle\pm\,0.20}$ & \noTableDataFont $30.69\,{\scriptstyle\pm\,0.31}$ & \noTableDataFont $31.10\,{\scriptstyle\pm\,0.07}$ & \noTableDataFont $30.78\,{\scriptstyle\pm\,0.18}$ \\
\noTableDataFont Amazon Employee & \noTableDataFont $42.50\,{\scriptstyle\pm\,0.70}$ & \noTableDataFont $46.10\,{\scriptstyle\pm\,1.74}$ & \noTableDataFont $\boldsymbol{34.70}\,{\scriptstyle\pm\,0.23}$ & \noTableDataFont $\dagger$ & \noTableDataFont $46.69\,{\scriptstyle\pm\,1.00}$ & \noTableDataFont $46.30\,{\scriptstyle\pm\,1.03}$ \\
\noTableDataFont APS Failure & \noTableDataFont $0.76\,{\scriptstyle\pm\,0.09}$ & \noTableDataFont $0.77\,{\scriptstyle\pm\,0.12}$ & \noTableDataFont $0.83\,{\scriptstyle\pm\,0.12}$ & \noTableDataFont $\boldsymbol{0.68}\,{\scriptstyle\pm\,0.08}$ & \noTableDataFont $0.72\,{\scriptstyle\pm\,0.08}$ & \noTableDataFont $0.69\,{\scriptstyle\pm\,0.09}$ \\
\noTableDataFont Bank Marketing & \noTableDataFont $31.52\,{\scriptstyle\pm\,0.24}$ & \noTableDataFont $31.95\,{\scriptstyle\pm\,0.09}$ & \noTableDataFont $\boldsymbol{30.30}\,{\scriptstyle\pm\,0.20}$ & \noTableDataFont $32.07\,{\scriptstyle\pm\,0.36}$ & \noTableDataFont $31.57\,{\scriptstyle\pm\,0.79}$ & \noTableDataFont $30.94\,{\scriptstyle\pm\,0.95}$ \\
\noTableDataFont Credit Default & \noTableDataFont $25.26\,{\scriptstyle\pm\,0.29}$ & \noTableDataFont $26.07\,{\scriptstyle\pm\,0.20}$ & \noTableDataFont $\boldsymbol{24.04}\,{\scriptstyle\pm\,0.36}$ & \noTableDataFont $24.50\,{\scriptstyle\pm\,0.34}$ & \noTableDataFont $25.83\,{\scriptstyle\pm\,0.39}$ & \noTableDataFont $25.81\,{\scriptstyle\pm\,0.52}$ \\
\noTableDataFont Diabetes & \noTableDataFont $22.70\,{\scriptstyle\pm\,0.11}$ & \noTableDataFont $23.19\,{\scriptstyle\pm\,0.18}$ & \noTableDataFont $\boldsymbol{22.34}\,{\scriptstyle\pm\,0.05}$ & \noTableDataFont $\dagger$ & \noTableDataFont $23.03\,{\scriptstyle\pm\,0.24}$ & \noTableDataFont $22.97\,{\scriptstyle\pm\,0.31}$ \\
\noTableDataFont Diamonds & \noTableDataFont $55.27\,{\scriptstyle\pm\,1.22}$ & \noTableDataFont $57.31\,{\scriptstyle\pm\,3.05}$ & \noTableDataFont $\boldsymbol{53.99}\,{\scriptstyle\pm\,0.68}$ & \noTableDataFont $54.30\,{\scriptstyle\pm\,0.99}$ & \noTableDataFont $57.11\,{\scriptstyle\pm\,2.52}$ & \noTableDataFont $56.31\,{\scriptstyle\pm\,2.58}$ \\
\noTableDataFont FLUXNET & \noTableDataFont --- & \noTableDataFont --- & \noTableDataFont --- & \noTableDataFont --- & \noTableDataFont --- & \noTableDataFont --- \\
\noTableDataFont Online Shoppers & \noTableDataFont $40.46\,{\scriptstyle\pm\,0.49}$ & \noTableDataFont $40.88\,{\scriptstyle\pm\,0.19}$ & \noTableDataFont $\boldsymbol{36.82}\,{\scriptstyle\pm\,0.43}$ & \noTableDataFont $39.34\,{\scriptstyle\pm\,0.68}$ & \noTableDataFont $40.65\,{\scriptstyle\pm\,0.20}$ & \noTableDataFont $40.52\,{\scriptstyle\pm\,0.12}$ \\
\noTableDataFont SoilFormer & \noTableDataFont $\boldsymbol{31.22}\,{\scriptstyle\pm\,0.64}$ & \noTableDataFont $32.56\,{\scriptstyle\pm\,0.37}$ & \noTableDataFont $32.41\,{\scriptstyle\pm\,0.19}$ & \noTableDataFont $31.68\,{\scriptstyle\pm\,0.66}$ & \noTableDataFont $31.42\,{\scriptstyle\pm\,0.61}$ & \noTableDataFont $\boldsymbol{\mathit{31.03}}\,{\scriptstyle\pm\,0.65}$ \\
\noTableDataFont Wine Quality & \noTableDataFont $1.29\,{\scriptstyle\pm\,0.32}$ & \noTableDataFont $1.72\,{\scriptstyle\pm\,0.39}$ & \noTableDataFont $1.65\,{\scriptstyle\pm\,0.20}$ & \noTableDataFont $\boldsymbol{0.86}\,{\scriptstyle\pm\,0.20}$ & \noTableDataFont $1.36\,{\scriptstyle\pm\,0.30}$ & \noTableDataFont $\boldsymbol{\mathit{0.86}}\,{\scriptstyle\pm\,0.54}$ \\
\bottomrule
\end{tabular*}

\end{minipage}
\par\vskip0pt
}
\end{table*}

\subsection{Cross-confidence Analysis}
\label{app:cross-confidence-results}

This subsection extends Section~\ref{sec:cross-confidence} in two directions. We first ask when predictions improve on the supplied estimates themselves, directly testing the shift from missing-value prediction to estimate refinement. We then test mixed-quality inputs and vary declared confidence independently of estimate quality. Both analyses use the three training recipes and summarize performance across datasets containing the relevant feature type.

\subsubsection{Prediction Gains and Estimate Refinement}
\label{app:estimate-refinement}

\begin{figure*}[t]
\begin{minipage}{\textwidth}
\centering
\includegraphics[width=\textwidth]{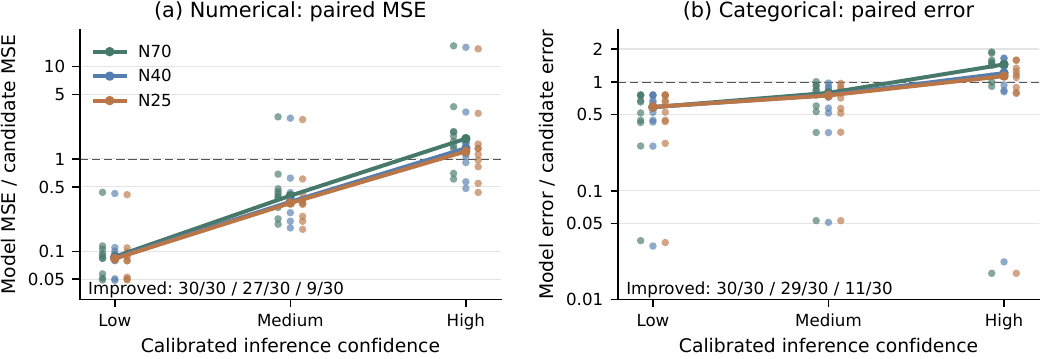}
\caption{Prediction error relative to the supplied estimate; ratios below one indicate refinement. Paired candidate and model errors are averaged over surviving backbones and seeds within each dataset and recipe before division. Numerical errors are MSE in training-standardized units; categorical errors are pooled misclassification rates. Colored points represent datasets, and solid lines show dataset medians for each recipe. Both vertical axes are logarithmic. Counts give improved comparisons out of 30 at Low, Medium, and High.}
\label{fig:estimate-refinement}
\end{minipage}

\par\medskip
\begin{minipage}{\textwidth}
\centering
\includegraphics[width=\textwidth]{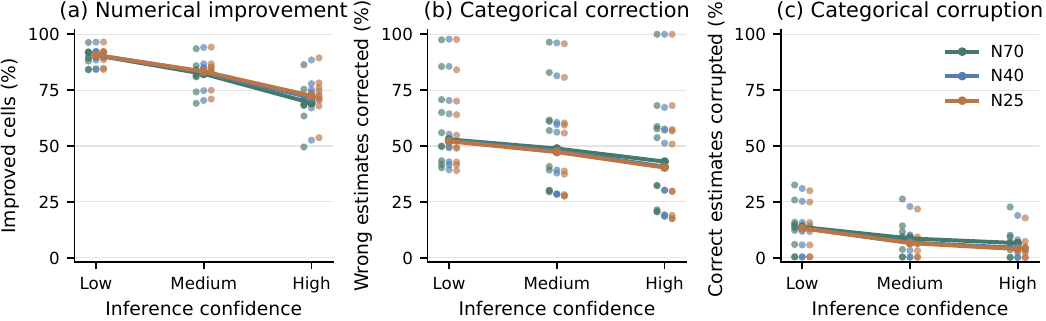}
\caption{Cell-level refinement. (a) Fraction of numerical targets with smaller squared error after prediction; a vector-valued cell uses its mean squared component error. (b) Fraction of incorrect categorical estimates corrected. (c) Fraction of correct categorical estimates corrupted. Numerical fractions average surviving model scores; categorical rates pool the corresponding model--target pairs within each dataset and recipe. Colored points represent datasets, and solid lines show their medians. The two categorical rates use different denominators and must be weighted by the prevalence of incorrect and correct estimates to obtain the net accuracy change.}
\label{fig:refinement-mechanisms}
\end{minipage}
\end{figure*}

Improvement over prediction without estimates does not necessarily imply improvement over the estimates themselves. Figure~\ref{fig:estimate-refinement} compares predictions with their supplied estimates on the same cells, using numerical MSE and categorical error with equal weight per target. This paired comparison isolates refinement from the benefit of merely receiving evidence. For each feature type, medians summarize the 30 comparisons formed by ten datasets and three recipes. Figure~\ref{fig:refinement-mechanisms} further distinguishes how often estimates improve from how much the aggregate error changes.

\textbf{Numerical Refinement.}\quad At \texttt{CL-Low}, model MSE is lower than estimate MSE across all datasets and recipes, with a median reduction of 91.4\%. Aggregate refinement remains common at \texttt{CL-Medium}, but occurs in only nine of 30 comparisons at \texttt{CL-High}. Nevertheless, the median fraction of cells improved remains 71.2\% at High. Many comparisons therefore improve most estimates while increasing aggregate MSE: a smaller number of large errors can outweigh widespread local gains. Weak estimates leave substantial room for refinement, whereas already accurate estimates require the model to preserve their quality as well as correct their errors.

\textbf{Categorical Correction and Corruption.}\quad The model lowers estimate error across all datasets and recipes at Low and Medium, but in only eleven of 30 comparisons at High. The median relative reduction at Low is 41.3\%. To explain the difference, Figure~\ref{fig:refinement-mechanisms}(b,c) separates correcting an incorrect estimate from changing a correct one into an error. At High, the model corrects a median 41.6\% of incorrect estimates and corrupts 4.6\% of correct estimates. However, incorrect estimates account for only 8.3\% of targets, compared with 46.1\% at Low. Because most High estimates are already correct, even a small corruption rate can outweigh the corrections. The median net accuracy gain consequently changes from $+16.3$ percentage points at Low to $-1.5$ at High, identifying preservation of accurate evidence as an important limit of the learned refinement.

\subsubsection{Mixed and Miscalibrated Confidence at Inference}
\label{app:miscalibrated-confidence}

\begin{figure*}[p]
\vbox to \textheight{%
\begin{minipage}{\textwidth}
\setcaptiontype{figure}
\centering
\includegraphics[width=\textwidth]{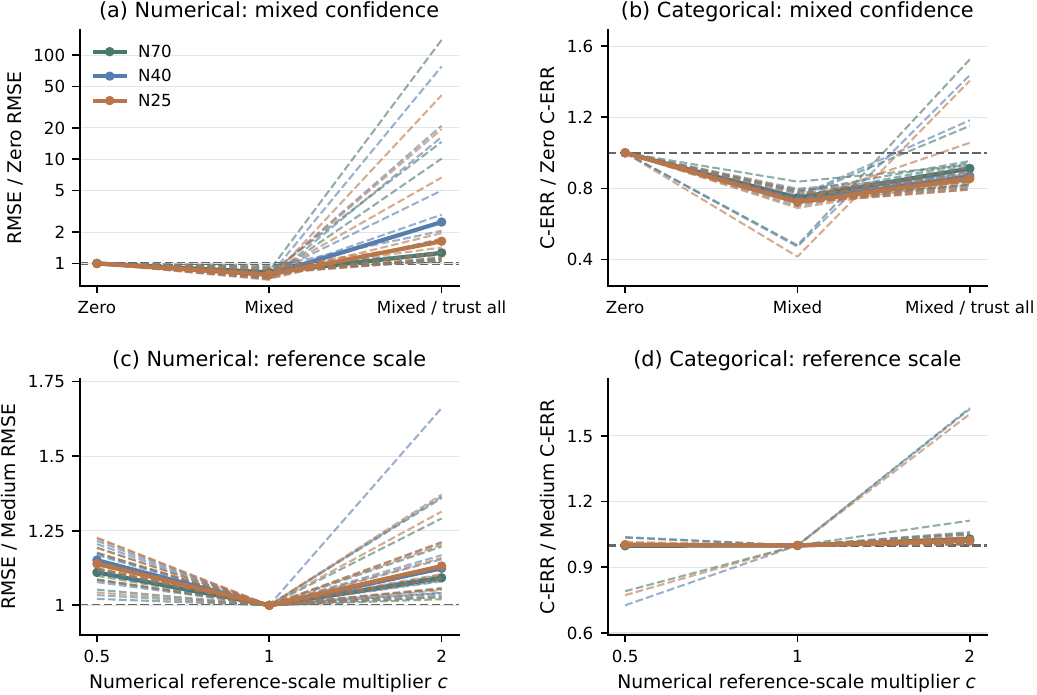}
\caption{Mixed confidence and reference-scale misspecification. (a,b) Errors relative to \texttt{CL-Zero} for a calibrated mixture and the same candidates treated as fully trusted; cells without estimates remain at zero confidence. (c,d) Errors relative to calibrated Medium when the numerical reference scale is multiplied by $c$, retaining the candidate values. Categorical inputs are unchanged in this intervention; their prediction changes reflect altered numerical context. Dashed lines represent datasets, and solid lines show their medians for each recipe. Ratios use mean errors over surviving backbones and seeds within each dataset. Only panel (a) uses a logarithmic vertical axis.}
\label{fig:mixed-scale-confidence}
\end{minipage}

\par\vfill
\begin{minipage}{\textwidth}
\setcaptiontype{figure}
\centering
\includegraphics[width=\textwidth]{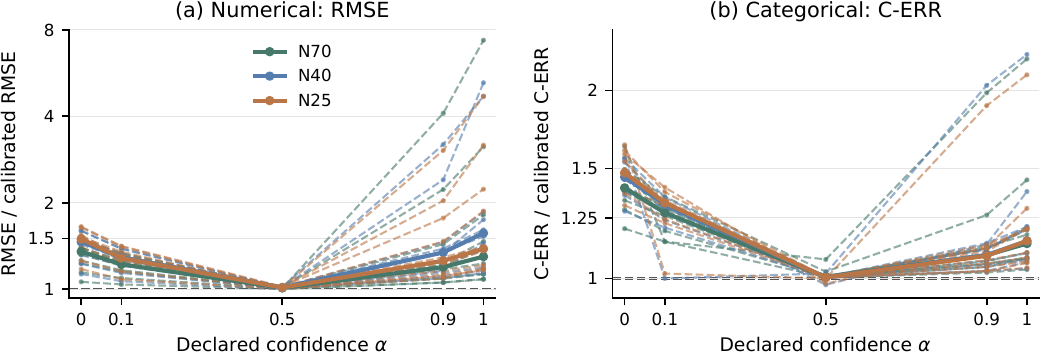}
\caption{Confidence interventions with unchanged medium-quality estimates. Each error is divided by the error obtained using the same candidates with their original calibrated confidence scores. Dashed lines represent datasets; solid lines show dataset medians for each recipe. Numerical RMSE and categorical C-ERR follow the column-macro definitions used in the main text. Both vertical axes are logarithmic. At $\alpha=0$, candidate values have no effect on the input.}
\label{fig:fixed-candidate-confidence}
\end{minipage}
\par\vskip0pt
}
\end{figure*}

\textbf{Mixed Confidence.}\quad To test whether one record can combine different qualities of evidence, we independently assign each target cell to Zero, Low, Medium, or High with equal probability. This calibrated mixture lowers RMSE and C-ERR relative to \texttt{CL-Zero} across all datasets and recipes, with median reductions of 20.0\% and 26.5\% (Figure~\ref{fig:mixed-scale-confidence}(a,b)). Declaring the same available estimates fully trusted worsens both metrics relative to the calibrated mixture, by a median 124.0\% in RMSE and 18.0\% in C-ERR; cells without estimates remain at zero confidence. The model can therefore use mixed-quality evidence, but gains depend on expressing how reliable each estimate is rather than simply marking it as available.

\textbf{Confidence at Fixed Estimate Quality.}\quad To separate the effect of declared confidence from that of better evidence, we retain the \texttt{CL-Medium} estimates and replace their confidence scores with a shared $\alpha\in\{0,0.1,0.5,0.9,1\}$. Increasing confidence alone does not produce the monotonic gains observed in the calibrated sweep (Figure~\ref{fig:fixed-candidate-confidence}). The intermediate choice $\alpha=0.5$ gives the lowest error across all datasets and recipes. These controls support interpreting the improvement in Section~\ref{sec:cross-confidence} as the benefit of more reliable estimates receiving appropriate weight, rather than greater input weight alone.

\textbf{Overconfidence and Underconfidence.}\quad We also test severe distortions at both ends of the reliability range. Assigning $\alpha=0.9$ to \texttt{CL-Low} estimates increases RMSE and C-ERR across all datasets and recipes relative to calibrated \texttt{CL-Low}, by a median 186.6\% and 25.2\%; prediction also becomes worse than using no estimates. Conversely, assigning $\alpha=0.1$ to \texttt{CL-High} estimates worsens both metrics relative to calibrated \texttt{CL-High}, by a median 64.6\% and 147.8\%. These interventions retain the estimates themselves: overweighting unreliable evidence or suppressing useful evidence introduces prediction error independently of estimate quality.

\textbf{Misspecified Reference Scales.}\quad Errors in the prior reference can distort confidence even when estimate quality is unchanged. We multiply the numerical reference scale $\tau_j^{\mathrm{ref}}$ by $c\in\{0.5,1,2\}$ when computing confidence from Eq.~\eqref{eq:precision}, while keeping the Medium estimates and their error scales fixed. A smaller reference lowers confidence; a larger one raises it. Both departures worsen RMSE across all datasets and recipes, with median increases of 12.9\% for $c=0.5$ and 10.3\% for $c=2$ (Figure~\ref{fig:mixed-scale-confidence}(c,d)). Categorical estimates and confidence remain unchanged, but doubling the numerical reference also increases categorical error wherever numerical context is available. Misweighting numerical evidence can therefore impair categorical predictions as well, supporting calibration of the reference scale used to assign confidence.

\subsection{N.A. Data Augmentation}
\label{app:na-results}

This subsection details the controlled augmentation in Section~\ref{sec:na-results}, which tests whether preserving inapplicability and its reasons helps predict applicable values. We retain the original records, source features, and data partitions of Online Shoppers, SoilFormer, and Diabetes, and add deterministic attributes whose applicability depends on existing columns. Table~\ref{tab:na-augmentation-overview} summarizes the resulting tables.

\subsubsection{Augmented Fields and Applicability Rules}
\label{app:na-augmentation}

The added fields use two kinds of applicability conditions. Necessary prerequisites include a positive page count for a per-page duration. Reporting scopes restrict a soil summary to a land-use group or a medical summary to a treatment history. Scope thresholds are determined from training covariates before examining held-out records. These constructions introduce known dependencies between applicability and existing fields; they test the value of preserving this information rather than establishing the prevalence or effects of native N.A. in other datasets.

A false applicability condition gives N.A.; a true condition gives an observed value when the required source values are available and valid. An unresolved condition or an unavailable applicable value gives N.O., while valid zero values remain observed. This preserves the distinction between an undefined value and an unknown one. Table~\ref{tab:na-aug-fields} lists the added attributes, their conditions, and their value definitions; fields in the conditions serve as reason inputs.

\begin{table}[t]
\caption{Dimensions after N.A. augmentation. Added counts refer to new logical columns. N.A. rates cover all verified N.A. cells across the three data partitions before artificial masking. Each vector-valued numerical cell counts once.}
\label{tab:na-augmentation-overview}
\centering
\footnotesize
\setlength{\tabcolsep}{3pt}
% BEGIN GENERATED: na_augmentation_overview_body.tex
\begin{tabular*}{\columnwidth}{@{\extracolsep{\fill}}lrrrr@{}}
\toprule
Dataset & \shortstack{Added\\NUM/CAT} & NUM & CAT & N.A. (\%)\\
\midrule
Online Shoppers & 5/5 & 15 & 13 & 16.6\\
SoilFormer & 15/3 & 44 & 8 & 14.5\\
Diabetes & 4/20 & 12 & 57 & 14.0\\
\bottomrule
\end{tabular*}
% END GENERATED: na_augmentation_overview_body.tex
\end{table}

\subsubsection{Input Treatments and Aggregation}

To isolate the contribution of N.A. semantics, MsEmb merges N.O. and N.A. into one learned reference, while the two \nano{} treatments use a separate N.A. embedding, without or with the reason projection in Section~\ref{sec:na}. All treatments use \texttt{CL-Zero}. Only applicable observed cells are masked and scored, so the comparison evaluates how N.A. context helps predict meaningful values rather than predicting N.A. itself. Reasons contribute only when their fields remain applicable and observed after masking.

No N.A. model meets the failure criterion in Appendix~\ref{app:loss-metrics}. For each dataset, input head, backbone, and metric, we omit the largest of the five test scores and pool the remaining four per backbone, giving 36 scores for the mean and sample SD. This trimming is applied separately to each test metric, so the omitted seed can differ; it does not change validation-based checkpoint selection.

\begin{table*}[p]
\caption{Added attributes and availability conditions in the three N.A.-augmented benchmarks. NUM and CAT identify numerical and categorical attributes. Fields referenced in Available When are reason inputs; symbols and qualifications are defined in the benchmark-specific notes below.}
\label{tab:na-aug-fields}
\label{tab:na-aug-shoppers}
\label{tab:na-aug-soilformer}
\label{tab:na-aug-diabetes}
\centering
\fontsize{7.5pt}{8.5pt}\selectfont
\setlength{\tabcolsep}{4pt}
\renewcommand{\arraystretch}{1.0}
% BEGIN GENERATED: na_aug_combined_body.tex
\begin{tabular*}{\textwidth}{@{\extracolsep{\fill}}lll@{}}
\toprule
Added Attribute & Available When & Value Definition\\
\midrule
\multicolumn{3}{@{}l@{}}{\textbf{(a) Online Shoppers}}\\[1pt]
Administrative Mean Page Duration (NUM) & A $>$ 0 & DA/A\\
Administrative Visit Pattern (CAT) & A $>$ 0 & One page / multiple pages, from A\\
Informational Mean Page Duration (NUM) & I $>$ 0 & DI/I\\
Informational Visit Pattern (CAT) & I $>$ 0 & One page / multiple pages, from I\\
Mean Seconds Per Page (NUM) & P $\geq$ 12 $\lor$ (A $>$ 0 $\land$ I $>$ 0) & (DA+DI+DP)/(A+I+P)\\
Section Mix (CAT) & P $\geq$ 12 $\lor$ (A $>$ 0 $\land$ I $>$ 0) & Presence pattern of A, I and P\\
Product Page Share (NUM) & (V = Returning $\land$ P $\geq$ 10) $\lor$ I $>$ 0 & P/(A+I+P)\\
Rate Profile (CAT) & (V = Returning $\land$ P $\geq$ 10) $\lor$ I $>$ 0 & No bounce (B=0); exit-dominant (X$\geq$2*B); other\\
Support Seconds Per Page (NUM) & DP $\geq$ 900 $\lor$ A $>$ 0 & (DA+DI)/(A+I+P)\\
Visitor Calendar (CAT) & DP $\geq$ 900 $\lor$ A $>$ 0 & Joint V and W category\\
\midrule
\multicolumn{3}{@{}l@{}}{\textbf{(b) SoilFormer}}\\[1pt]
Cropland Cover Crop Median (NUM) & $\mathrm{LC}_1=\mathrm{B}\land\mathrm{LU}=111$ & Mapped cover-crop fraction CC\\
Dry Months Mean Precipitation (NUM) & $\exists m:\ T_m>0\land P_m<2T_m$ & Mean P over months with T$>$0 and P$<$2*T\\
Forestry Woodland Type (CAT) & $\mathrm{LC}_1=\mathrm{C}\land\mathrm{LU}=120$ & Broadleaved / coniferous / mixed, from LC\\
Subzero Month Pattern (CAT) & $\exists m:\ T_m<0$ & Calendar-ordered set of months with T$<$0\\
Top Capacity Gap (NUM) & $\mathrm{LC}_1\in\{\mathrm{B,D,E}\}\land\mathrm{LU}\in\{111,120\}$ & Mean (FC-WP) at 0, 5 and 15 cm\\
Top Field Capacity Mean (NUM) & $\mathrm{LC}_1\in\{\mathrm{B,D,E}\}\land\mathrm{LU}\in\{111,120\}$ & Mean FC at 0, 5 and 15 cm\\
Monthly Precip Range (NUM) & mean(T) $\geq$ 9 $\lor$ sum(P) $\geq$ 800 & max(P)-min(P)\\
Monthly Temperature Range (NUM) & mean(T) $\geq$ 9 $\lor$ sum(P) $\geq$ 800 & max(T)-min(T)\\
Depth Density Contrast (NUM) & PHW-PHC $\geq$ 0.6 $\lor$ SOC $\geq$ 20 & BD0-BD1\\
Cover Family (CAT) & PHW-PHC $\geq$ 0.6 $\lor$ SOC $\geq$ 20 & First letter of LC\\
Deep Shallow Contrast (NUM) & $G_s\ge20\lor Z\ge250$ & Deep minus shallow mean (FC-WP)\\
Field Capacity Depth Contrast (NUM) & $G_s\ge20\lor Z\ge250$ & Deep minus shallow mean FC\\
Carbon Nitrogen Ratio (NUM) & PHW $<$ 6.2 $\lor$ sum(P) $\geq$ 700 & SOC/N, with N$>$0\\
Carbon Volume Proxy (NUM) & PHW $<$ 6.2 $\lor$ sum(P) $\geq$ 700 & SOC*BD\\
pH Method Gap (NUM) & BD $\geq$ 1.0 $\land$ PHW $\geq$ 5.0 & PHW-PHC\\
Summer--Winter Temperature Gap (NUM) & BD $\geq$ 1.0 $\land$ PHW $\geq$ 5.0 & Mean T in JJA minus mean T in DJF\\
Summer--Winter Precipitation Gap (NUM) & LAT $\geq$ 53 $\lor$ Z $\geq$ 300 & Mean P in JJA minus mean P in DJF\\
Summer Precip Share (NUM) & LAT $\geq$ 53 $\lor$ Z $\geq$ 300 & JJA precipitation / annual precipitation\\
\midrule
\multicolumn{3}{@{}l@{}}{\textbf{(c) Diabetes}}\\[1pt]
Prior Care Emergency Share (NUM) & O+E+I $>$ 0 & E/(O+E+I)\\
Adjusted Drug Up Share (NUM) & Any Up or Down status in DRUGS & Share of Up among Up/Down drug fields\\
Insulin Dose Status (CAT) & INS is Steady, Up, or Down & INS: Steady / Up / Down\\
Metformin Dose Status (CAT) & MET is Steady, Up, or Down & MET: Steady / Up / Down\\
Lab Rate (NUM) & INS $\ne$ No $\lor$ O $>$ 0 & L/H\\
Therapy Mix (CAT) & INS $\ne$ No $\lor$ O $>$ 0 & Prescription pattern: INS, MET, GLI-or-GLY\\
Medication Per Day Band (CAT) & O+E+I $>$ 0 $\lor$ (C=Ch $\land$ M$\geq$12) & M/H: below 3 / [3,6) / at least 6\\
Prior Care Pattern (CAT) & O+E+I $>$ 0 $\lor$ (C=Ch $\land$ M$\geq$12) & Presence pattern of O, E and I\\
Procedure Per Day Band (CAT) & Oral drug prescribed $\lor M\ge16$ & R/H: zero / (0,0.5] / above 0.5\\
Common Oral Adjustment (CAT) & Oral drug prescribed $\lor M\ge16$ & Adjustment profile of MET, GLI and GLY\\
Diagnosis Per Day Band (CAT) & H$\geq$6 $\lor$ D$\geq$8 & D/H: below 1 / [1,2) / at least 2\\
Primary Diagnosis Group (CAT) & H$\geq$6 $\lor$ D$\geq$8 & Existing diagnosis group DX\\
Lab Medication Band (CAT) & M$\geq$18 $\lor$ O+E+I$>$0 & L/M: below 2 / [2,4) / at least 4\\
Dominant Prior Care (CAT) & M$\geq$18 $\lor$ O+E+I$>$0 & Maximal O/E/I; none / positive tie\\
Procedure Medication Ratio (NUM) & L$\geq$50 $\lor$ C=Ch & R/M\\
Insulin Metformin Adjustment (CAT) & L$\geq$50 $\lor$ C=Ch & Adjustment profile of INS and MET\\
Prior Care Count Band (CAT) & R$\geq$3 $\lor$ INS$\ne$No & O+E+I: zero / one / at least two\\
Procedure Band (CAT) & R$\geq$3 $\lor$ INS$\ne$No & R: zero / 1--2 / at least 3\\
Prior Emergency Band (CAT) & D$\geq$8 $\lor$ O+E$>$0 & E: zero / one / at least two\\
Medication Band (CAT) & D$\geq$8 $\lor$ O+E$>$0 & M: below 10 / 10--19 / at least 20\\
Prior Inpatient Band (CAT) & C=Ch $\lor$ H$\geq$5 & I: zero / one / at least two\\
Insulin Metformin State (CAT) & C=Ch $\lor$ H$\geq$5 & Joint INS and MET state\\
Diagnosis Medication Band (CAT) & M$\geq$20 $\lor$ D$\geq$8 & D/M: below 0.3 / [0.3,0.6) / at least 0.6\\
Any Drug Adjustment (CAT) & M$\geq$20 $\lor$ D$\geq$8 & Adjustment profile across DRUGS\\
\bottomrule
\end{tabular*}
\par\smallskip
\begin{minipage}{\textwidth}
\fontsize{7pt}{8pt}\selectfont\raggedright
\textbf{Online Shoppers.} \textit{Inputs:} A: administrative-page count; I: information-page count; P: product-page count; DA: administrative-page duration; DI: information-page duration; DP: product-page duration; V: visitor type; W: weekend indicator; B: bounce rate; X: exit rate. Durations are in seconds. Returning means Returning Visitor.
\par\smallskip
\textbf{SoilFormer.} \textit{Inputs:} LC: land-cover code; LU: land-use code; T: monthly temperatures; P: monthly precipitation; PHW: pH in water; PHC: pH in calcium chloride; SOC: organic carbon content; N: total nitrogen content; BD: bulk density; BD0: 0--10 cm bulk density; BD1: 10--20 cm bulk density; Z: elevation; LAT: latitude; FC: field-capacity water-content profile; WP: wilting-point water-content profile; CC: mapped cover-crop fraction. Shallow and deep profiles use 0, 5, 15 cm and 60, 100, 200 cm, respectively. JJA and DJF denote June--August and December--February. T is in degrees Celsius, P in millimeters, Z in meters, and LAT in degrees. Other thresholds retain source units. LC prefixes B/C/D/E denote cropland/woodland/shrubland/grassland; LU codes 111/120 denote agriculture/forestry. CC is a contextual prediction, and the dry-month rule is an empirical climate definition. LU uses the code before any colon. $\mathrm{LC}_1$ is the first letter of LC; $m$ indexes months; $G_s$ is mean FC$-$WP at 0, 5, and 15 cm.
\par\smallskip
\textbf{Diabetes.} \textit{Inputs:} O: previous outpatient visits; E: previous emergency visits; I: previous inpatient visits; H: hospital stay in days; L: laboratory count; R: procedure count; M: medication count; D: diagnosis count; C: treatment-change indicator; INS: insulin status; MET: metformin status; GLI: glipizide status; GLY: glyburide status; DX: primary diagnosis group; DRUGS: all retained medication-status fields. C=Ch denotes a treatment change. No means that a drug was not prescribed. Adjustment profiles distinguish none prescribed, steady only, upward only, downward only, and mixed upward/downward changes. DRUGS includes combination-drug fields separately. Counts divided by H are descriptive normalizations, not event-time measurements. An oral-drug condition holds when any of MET, GLI or GLY is prescribed (status other than No).
\end{minipage}
% END GENERATED: na_aug_combined_body.tex
\end{table*}

\section{N.O. Ablation Studies}
\label{app:no-ablations}

\subsection{Feature Mask Ratio}
\label{app:mask-ratio-ablation}

This subsection examines how the training mask ratio in Section~\ref{sec:experimental-setup} affects reconstruction. Masking reduces clean context while increasing the number of prediction targets. For \nano{}, these targets can carry uncertain estimates, so changing the ratio also changes how extensively the model learns from uncertain inputs.

\begin{figure*}[tp]
\centering
\includegraphics[width=\textwidth]{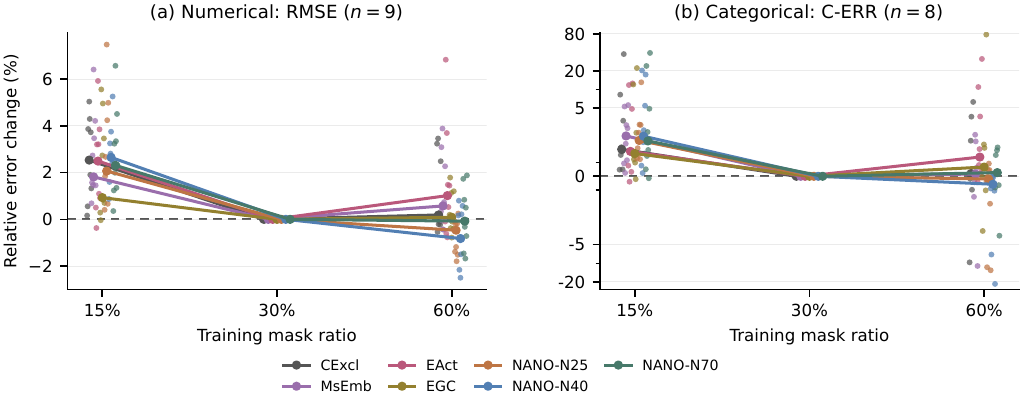}
\caption{Training-mask sensitivity at \texttt{CL-Zero}, relative to the same head trained with 30\% masking. Each dot is one dataset's relative error change after averaging results over five seeds; lines connect dataset medians. All seven heads have results for the included datasets: nine with numerical features in (a) and eight with categorical features in (b). Negative changes indicate lower error. The categorical axis is linear between $-5\%$ and $5\%$ and logarithmic outside this range.}
\label{fig:mask-ratio-heads}
\end{figure*}

\begin{figure*}[t]
\centering
\includegraphics[width=\textwidth]{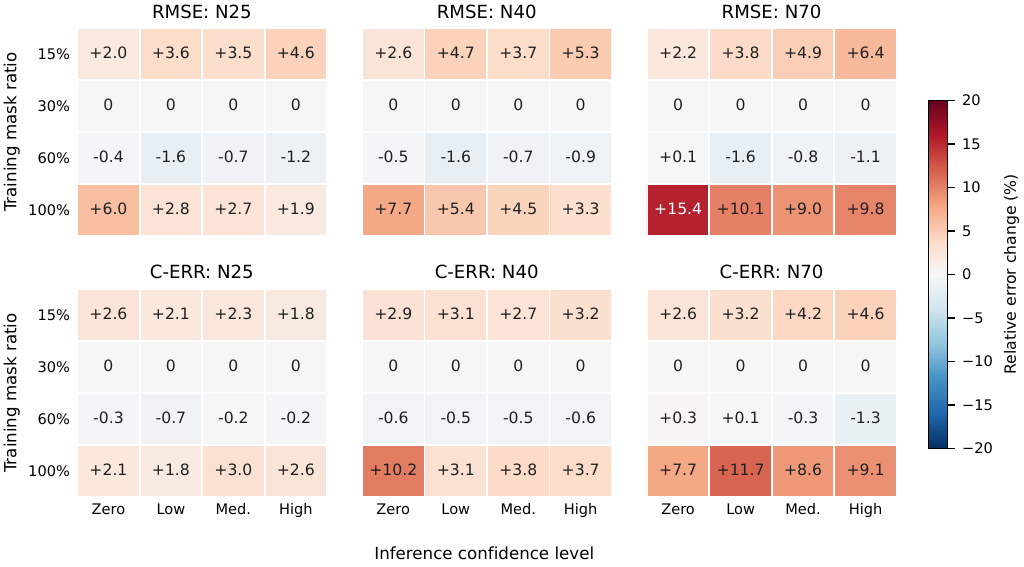}
\caption{Training-mask effects on \nano{} at each inference confidence level. Cells show median relative point-error changes across datasets after averaging results over five seeds. Each column uses the same recipe and confidence level trained at 30\% as its reference, so colors compare training masks rather than absolute performance across confidence levels. Blue indicates lower error. Ten datasets contribute to each cell, except categorical High, where the zero-error Wine Quality reference is omitted from relative changes. The 100\% arm includes only \nano{}; evaluation masking remains 30\% throughout.}
\label{fig:mask-ratio-confidence}
\end{figure*}

\paragraph{Experimental Setup.}
We compare training mask ratios of 15\%, 30\%, and 60\% for the four baseline heads and three \nano{} recipes, using all eleven datasets, FT-Transformer, and five seeds. We also test 100\% masking for \nano{} to examine learning without clean applicable observations. Baselines are omitted at this endpoint because they would have neither observed values nor informative estimates as context. Validation and test masking remain fixed at 30\%, with identical targets and estimates across training ratios. All other settings follow the main experiments.

\paragraph{Partial Masking.}
Figure~\ref{fig:mask-ratio-heads} compares each head with its own 30\% training reference when no estimates are supplied at inference. Reducing masking to 15\% generally increases error across heads. Increasing it to 60\% has mixed effects on the baselines, while \nano{-N25} and \nano{-N40} benefit more consistently. Across all inference confidence levels, the median reductions for \nano{} are modest: 0.88\% in RMSE and 0.36\% in C-ERR (Table~\ref{tab:mask-ratio-summary}). The improvement is more widespread for numerical prediction. Thus more extensive masking can help, but does not make 60\% uniformly preferable to 30\%.

\begin{table}[tb]
\caption{Median changes for \nano{} relative to the same dataset, recipe, and inference confidence at 30\% training masking. Scores are first averaged over five seeds, then paired changes are summarized across datasets, the three recipes, and four confidence levels. Point changes are relative percentages; NLL changes are signed differences in nats. The three zero-error C-ERR references are omitted only from percentage calculations. Negative values favor the new ratio.}
\label{tab:mask-ratio-summary}
\centering
\footnotesize
\setlength{\tabcolsep}{2pt}
\begin{tabular*}{\columnwidth}{@{\extracolsep{\fill}}rrrrr@{}}
\toprule
 & \multicolumn{2}{c}{Point change (\%)} & \multicolumn{2}{c}{NLL change}\\
\cmidrule(lr){2-3}\cmidrule(lr){4-5}
Mask & RMSE & C-ERR & NUM & CAT\\
\midrule
15\% & +3.88 & +2.98 & +0.0399 & +0.0134\\
30\% & 0 & 0 & 0 & 0\\
60\% & -0.88 & -0.36 & -0.0012 & -0.0018\\
100\% & +6.11 & +5.69 & +0.1153 & +0.0277\\
\bottomrule
\end{tabular*}

\end{table}

\paragraph{Full Masking.}
Figure~\ref{fig:mask-ratio-confidence} shows that removing all clean applicable observations during training usually reduces accuracy, with the largest penalties for \nano{-N70}. At this endpoint, \nano{-N25} still receives estimates on approximately 75\% of target cells, compared with 30\% for \nano{-N70}. Among the three recipes trained at 100\% masking, \nano{-N25} also gives the lowest error without inference estimates on almost all datasets for each feature type. The amount of evidence retained through estimates therefore matters more when clean context disappears. Full masking remains a mixture of denoising and missing-value reconstruction because every recipe includes zero-confidence targets.

\paragraph{Confidence Response and Predictive Uncertainty.}
The monotonic gains from increasing estimate reliability and confidence persist across training ratios, and \texttt{CL-High} improves on \texttt{CL-Zero} for every dataset, recipe, and available feature type. The ability to use better estimates therefore survives substantial changes in training corruption, including the loss of clean context at 100\% masking.

NLL also generally favors moderate masking over 15\% or 100\%. However, the numerical NLL gains at 60\% are less consistent than the RMSE gains. Increasing masking can therefore improve point predictions without equally consistent gains in the predicted distributions.

These results support moderate training ratios while showing that \nano{}'s confidence response persists across the tested masking choices. The comparison uses FT-Transformer and the same 30\% test task throughout. Since training masking changes both context availability and supervision density, the results guide the overall corruption level rather than isolate either effect.

\subsection{Heterogeneous Training Confidence}
\label{app:heterogeneous-confidence-ablation}

Estimate availability and reliability can differ systematically across columns. We test whether fixing a confidence band for each column improves prediction when that pattern is known at deployment. We then change the inference pattern to examine what is lost when training no longer covers multiple confidence levels within every column.

\paragraph{Experimental Setup.}
We compare \nano{-N70} with three column-specific profiles using FT-Transformer on all eleven datasets and five seeds. Columns are assigned to Low or High in equal proportions (LH); Zero or High in proportions 70\%/30\% (ZH); or Zero, Low, Medium, and High in proportions 70\%/10\%/10\%/10\% (4B). Each assigned band remains fixed across records and epochs, with confidence and simulated estimates redrawn within nonzero bands; Zero-assigned targets receive no estimate. One random assignment is shared across seeds for each dataset, with proportions rounded to whole columns. The 4B profile preserves \nano{-N70}'s intended overall mixture but fixes it by column; LH and ZH also change the overall availability or strength of evidence.

All models use 30\% masking and the same checkpoint-selection procedure. Confidence profiles apply only to masked targets; unmasked observations remain clean context. At inference, we either apply one of the four confidence bands to every column or use one of the three column profiles, holding targets and supplied evidence fixed across models.

\begin{figure}[t]
\centering
\includegraphics[width=\columnwidth]{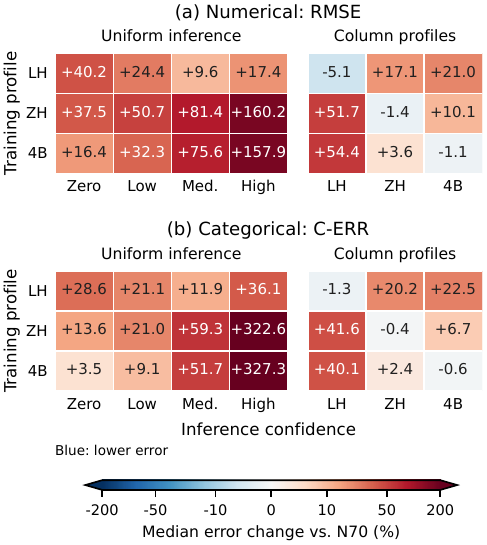}
\caption{Transfer between training and inference confidence profiles. Cells show median relative point-error changes from \nano{-N70} under the same inference view, after averaging results over five seeds within each dataset. Each column has its own reference, so colors compare models rather than absolute errors across confidence levels. The color scale uses $\operatorname{asinh}(\Delta/5)$ for percentage change $\Delta$, saturating at $\pm200\%$; annotations retain full values. Ten datasets contribute per feature type, except categorical High, where the zero-error Wine Quality reference is omitted only from relative changes.}
\label{fig:column-confidence-transfer}
\end{figure}

\begin{figure}[t]
\centering
\includegraphics[width=\columnwidth]{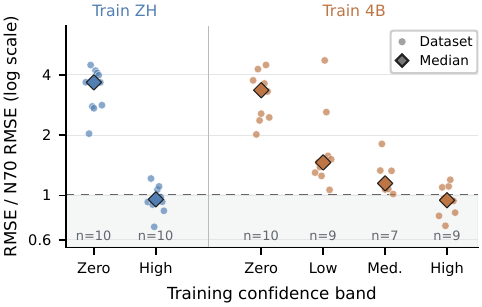}
\caption{Numerical RMSE at uniform High, grouped by each column's training band. Within each dataset and group, errors are averaged over the same columns and all five seeds in both models, then divided by the corresponding \nano{-N70} error. Dots show datasets and diamonds their medians; $n$ gives the number of datasets. The vertical axis is logarithmic.}
\label{fig:column-confidence-bands}
\end{figure}

\paragraph{Matching Deployment Patterns.}
Training with the matching column profile usually improves on \nano{-N70}, although the gains are modest outside LH (Table~\ref{tab:column-confidence-summary}). LH gives a median RMSE reduction of 5.10\%. For 4B, which most closely matches \nano{-N70}'s confidence proportions, the median reductions are 1.06\% in RMSE and 0.55\% in C-ERR. NLL generally supports these gains, although numerical improvements are less consistent for 4B. Matching a known deployment pattern can therefore help, but provides a limited advantage over training every column across all confidence levels.

\begin{table}[tb]
\caption{Matching column-confidence training and inference profiles, compared with \nano{-N70} on the same inputs. Point changes are median relative changes across datasets after averaging results over five seeds; negative values indicate improvement. NLL columns count datasets with lower NLL. Each metric is evaluated on the ten datasets containing its required feature type. LH, ZH, and 4B denote the profiles defined in the text.}
\label{tab:column-confidence-summary}
\centering
\footnotesize
\setlength{\tabcolsep}{2pt}
\begin{tabular*}{\columnwidth}{@{\extracolsep{\fill}}lrrrr@{}}
\toprule
 & \multicolumn{2}{c}{Point change (\%)} & \multicolumn{2}{c}{NLL wins}\\
\cmidrule(lr){2-3}\cmidrule(lr){4-5}
Profile & RMSE & C-ERR & NUM & CAT\\
\midrule
LH & -5.10 & -1.27 & 9/10 & 9/10\\
ZH & -1.44 & -0.43 & 8/10 & 8/10\\
4B & -1.06 & -0.55 & 6/10 & 8/10\\
\bottomrule
\end{tabular*}

\end{table}

\paragraph{Transfer across Confidence Profiles.}
Figure~\ref{fig:column-confidence-transfer} tests whether these gains transfer beyond the assigned profiles. When all columns use Zero or Low at inference, \nano{-N70} outperforms every fixed-profile model on both metrics wherever the required feature type is present. The ZH and 4B models also deteriorate sharply when all columns use High, despite receiving calibrated estimates. Their median C-ERR increases relative to \nano{-N70} are approximately 18 percentage points, so the large relative losses are not simply an artifact of small reference errors. Unlike \nano{-N70} and LH, these models frequently lose the monotonic improvement from Zero to High. Covering confidence levels across a table is therefore insufficient if individual columns never learn to use them.

\paragraph{Column-wise Diagnostic.}
To locate the transfer failure, Figure~\ref{fig:column-confidence-bands} separates uniform-High numerical errors by each column's training band. Error more than triples relative to \nano{-N70} at the median dataset for Zero-assigned columns in ZH and 4B. High-assigned columns remain close to the reference or improve slightly. A Zero-assigned column appears as clean context when unmasked, but never receives an estimate as a prediction target during training. The contrast points to the importance of learning how to use uncertain evidence for each column, rather than merely observing its values elsewhere in the training task.

\paragraph{Practical Implication and Scope.}
\nano{-N70} itself benefits from all three column-specific inference profiles relative to inference without estimates. Column-specific training is therefore unnecessary for using uneven evidence. When deployment conditions are uncertain, these results favor exposing every column to multiple confidence levels during training. Restricting a column to one band can help when that range remains stable, but reduces transfer to other ranges. This conclusion concerns training coverage: confidence at inference should still reflect each estimate's reliability. The comparison tests fixed band assignments, rather than column-specific mixtures that retain multiple levels.

\section{Data Acknowledgments}
\label{app:data-acknowledgments}

The FLUXNET products used in this study were produced and harmonized by the contributing site teams, regional eddy-covariance networks, and data processing centers. We acknowledge the Alpine Peatland, Collie, and Emerald site teams, the TERN data infrastructure, and the OzFlux network \citep{grover2026alpine,beringer2026collie,feitz2026emerald,beringer2016ozflux,isaac2017ozflux}. These products include a modified version of ERA5 hourly data supplied by the Copernicus Climate Change Service.

\section{Data and Code Availability}
\label{app:repository}

The data, source code, and experimental scripts for this study are available in the project repository at \url{https://github.com/earthroverprogram/nano_tabular}. The repository also provides instructions for reproducing the main experiments and ablation studies.

\section{AI Usage Declaration}
\label{app:ai-usage}

\textbf{Theory.}\quad The theoretical insights and mathematical derivations were developed by the authors. AI assisted with checking equation correctness and detecting errors or inconsistencies in notation.

\textbf{Code.}\quad The authors designed the experiments, including the selection of baselines, benchmarks, backbones, loss functions, and metrics. The design and development of \nano{} were carried out by the authors. AI assisted with data processing, adaptation of baselines and backbones, and auditing and testing the codebase.

\textbf{Writing.}\quad The authors designed and wrote the complete manuscript, including preparing figures and tables, choosing references, and designing the appendices. AI assisted with bibliography verification, word- and sentence-level language editing, and visual refinement of figures and tables. AI was not used for large-scale writing of the manuscript.

The authors retain full responsibility for the scientific claims, experimental results, references, code, and final manuscript.

% Top-align single-column floats left for the final page.
\makeatletter
\setlength{\@fptop}{0pt}
\makeatother
\end{document}